%% file: sample.tex
\documentclass[sigconf]{acmart}

\AtBeginDocument{%
  }

\setcopyright{acmlicensed}
\copyrightyear{2018}
\acmYear{2018}
\acmDOI{XXXXXXX.XXXXXXX}

\acmConference[Conference acronym 'XX]{Make sure to enter the correct
  conference title from your rights confirmation emai}{June 03--05,
  2018}{Woodstock, NY}
\acmISBN{978-1-4503-XXXX-X/18/06}

\usepackage{amsfonts}
\usepackage{amsmath}
\usepackage{pifont}
\graphicspath{{./img/}}
\usepackage[ruled,linesnumbered]{algorithm2e}
\usepackage{xspace}
\usepackage{multirow}
\usepackage{subfigure}
\usepackage{xcolor}
\usepackage{colortbl}
\usepackage{threeparttable}
\newcommand{\sysname}{\textsc{GraphNAD}\xspace}
\definecolor{lightgreen}{RGB}{218, 242, 208}

\begin{document}


\title{Fine-tuning is Not Fine: Mitigating Backdoor Attacks in GNNs with Limited Clean Data}

\author{Jiale Zhang}
\affiliation{%
  \institution{Yangzhou University}
  \city{Yangzhou}
  \state{Jiangsu}
  \country{China}}
\email{jialezhang@yzu.edu.cn}

\author{Bosen Rao}
\affiliation{%
  \institution{Yangzhou University}
  \city{Yangzhou}
  \state{Jiangsu}
  \country{China}}
\email{MX120230573@stu.yzu.edu.cn}

  \author{Chengcheng Zhu}
\affiliation{%
  \institution{Yangzhou University}
  \city{Yangzhou}
  \state{Jiangsu}
  \country{China}}
\email{MX120220554@stu.yzu.edu.cn}

  \author{Xiaobig Sun}
\affiliation{%
  \institution{Yangzhou University}
  \city{Yangzhou}
  \state{Jiangsu}
  \country{China}}
\email{xbsun@yzu.edu.cn}

  \author{Qingming Li}
\affiliation{%
  \institution{Zhejiang University}
  \city{Hangzhou}
  \country{China}}
\email{liqm@zju.edu.cn}

  \author{Haibo Hu}
\affiliation{%
  \institution{The Hong Kong Polytechnic University}
  \city{Hong Kong}
  \country{China}}
\email{haibo.hu@polyu.edu.hk}

  \author{Xiapu Luo}
\affiliation{%
  \institution{The Hong Kong Polytechnic University}
  \city{Hong Kong}
  \country{China}}
\email{csxluo@comp.polyu.edu.hk}

  \author{Qingqing Ye}
\affiliation{%
  \institution{The Hong Kong Polytechnic University}
  \city{Hong Kong}
  \country{China}}
\email{qqing.ye@polyu.edu.hk}

  \author{Shouling Ji}
\affiliation{%
  \institution{Zhejiang University}
  \city{Hangzhou}
  \country{China}}
\email{sji@zju.edu.cn}

\renewcommand{\shortauthors}{Jiale Zhang et al.}
\begin{abstract}
Graph Neural Networks (GNNs) have achieved remarkable performance through their message-passing mechanism, yielding promising results across various domains. However, recent studies have highlighted the vulnerability of GNNs to backdoor attacks, which can lead the model to misclassify graphs with attached triggers as the target class. 
The effectiveness of recent promising defense techniques, such as fine-tuning or distillation, is heavily contingent on having comprehensive knowledge of the sufficient training dataset. Empirical studies have shown that fine-tuning methods require a clean dataset of 20\% to reduce attack accuracy to below 25\%, while distillation methods require a clean dataset of 15\%. 
These clean datasets are used to train an advanced teacher model and purify backdoor neurons in the distillation process. However, obtaining such a large amount of clean data is commonly impractical. 

In this paper, we propose a practical backdoor mitigation framework, denoted as \sysname, which can capture high-quality intermediate-layer representations in GNNs to enhance the distillation process with limited clean data, achieving effective backdoor mitigation. 
To achieve this, we address the following key questions: \textit{How to identify the appropriate attention representations in graphs for distillation? How to enhance distillation with limited data?} By adopting the graph attention transfer method, \sysname can effectively align the intermediate-layer attention representations of the backdoored model with that of the teacher model, forcing the backdoor neurons to transform into benign ones. Besides, we extract the relation maps from intermediate-layer transformation and enforce the relation maps of the backdoored model to be consistent with that of the teacher model, thereby ensuring model accuracy while further reducing the influence of backdoors. 
Extensive experimental results show that by fine-tuning a teacher model with only 3\% of the clean data, \sysname can reduce the attack success rate to below 5\%, while the model performance degradation is almost negligible, far surpassing the state-of-the-art (SOTA) defense methods.

\end{abstract}



\maketitle
\input{data/1-introduction}
\input{data/2-background}
\input{data/3-threat_model}
\input{data/4-method}
\input{data/5-evaluation}

\input{data/6-related_work}
\input{data/7-conclusion}

\newpage

\bibliographystyle{ACM-Reference-Format}
\bibliography{sample}

\newpage
\appendix
\input{data/99-appendix}

\end{document}

%% file: data/1-introduction.tex
\section{Introduction}

Graph data structures are ubiquitous in the real world, encompassing molecular structures \cite{Mansimov2019MolecularGeometry}, social networks \cite{s1}, knowledge graphs \cite{Liu2022FederatedSocial}, and traffic prediction \cite{Shi_2020_CVPR}. For example, in social networks, individuals are represented as nodes, with their relationships forming the edges. While in molecular structures, the nodes and edges denote molecules and chemical bonds, respectively. Unlike structured data in Euclidean space, graph data structures are complex and rich in information \cite{xia2021graph,liang2024survey,besta2023demystifying}. To learn vectorized representations from these complex graphs that contain sufficient information, Graph Neural Networks (GNNs) have recently emerged as highly successful models for processing graph data \cite{GCN,GAT,GIN,GSA}, and they have been effectively applied to tasks such as node classification \cite{Liu2023GapformerGT}, link prediction \cite{Zhang2018LinkPrediction}, and graph classification \cite{Zhang2018AnED}. GNNs leverage message-passing and recursive propagation mechanisms to iteratively learn and aggregate node attributes and local graph structure for effective representation \cite{S11,wu2020comprehensive,khoshraftar2024survey}.

Nevertheless, with the continual growth in model size and the escalating costs of training, initializing model training from the ground up frequently leads to substantial expenses. Consequently, the approach of leveraging pre-trained models for the establishment of model frameworks has become a practical solution \cite{Ji2018ModelreuseAttacks,lu2021learning,cao2023pre}. While the incorporation of pre-trained models can markedly streamline the creation process of machine learning systems, it also raises considerable concerns when these models are released by untrustworthy third parties \cite{wu2022trustworthy,zhang2024survey,shen2022model}.
Specifically, these models are susceptible to a backdoor attack where they incorrectly classify test graphs with embedded triggers as belonging to a specific class, yet still achieve high accuracy on unaltered graphs \cite{GTA,contras}. Existing research on graph backdoor attacks primarily focuses on the design of backdoor triggers, which can be categorized into fixed triggers \cite{badsub,clean} and optimizable triggers \cite{transferable}. Moreover, studies are ongoing to explore the optimal placement of backdoor triggers to improve the effectiveness and stealthiness \cite{EXPBA,EXP2,motifBA}.

\textbf{Drawbacks of detection and fine-tuning methods.} Although extensive research has been conducted in the realm of defending against backdoor attacks within the domains of computer vision (CV) \cite{Gu2019BadNetsEvaluating,Li2021InvisibleBackdoor,Liu2019ABSScanning,Wang2019NeuralCleanse,Weber2023RABProvable,Kumari2023BayBFedBayesian,zhang2024flpurifier,Gong2023RedeemMyself,Li2021AntibackdoorLearning} and natural language processing (NLP) \cite{chen2021badnl,pan2022hidden,shen2022constrained,liu2022piccolo,zhang2024instruction,xi2024defending}, the direct application of these methods to the graph domain is limited by their neglect of the topological information inherent in graph data. Regrettably, the investigation into backdoor defense tailored specifically for GNNs remains in its nascent phase.
To the best of our knowledge, the existing backdoor defense methods for graphs primarily concentrate on backdoor detection \cite{Detect1,Detect2,Detect3}. While detection can identify potential risks, it is insufficient as the backdoored model still requires purification. A straightforward approach to mitigating backdoor attacks is directly fine-tuning the backdoored model on a small subset of clean data. Further refinement involves combining fine-tuning with pruning techniques, which aim to erase more stealthy backdoors. However, these fine-tuning methods not only require a large amount of clean data to achieve defense effectiveness but also result in the degradation of the main task accuracy (ACC) as shown in Table. \ref{T-fine-tining}.


\begin{table}
	\centering
	\caption{Empirical results of fine-tuning and distillation solutions on PROTEINS dataset under attack method GTA \cite{GTA}.}
	\label{T-fine-tining}
\begin{tabular}{c|c|cccc}
\hline
\multirow{2}{*}{\textbf{\begin{tabular}[c]{@{}c@{}}Clean Data\\ Ratio\end{tabular}}} & \multirow{2}{*}{\textbf{GNN Models}} & \multicolumn{2}{c}{\textbf{Fine-tuning}} & \multicolumn{2}{c}{\textbf{Distillation}} \\ \cline{3-6} 
                                                                                     &                                      & \textbf{ASR}        & \textbf{ACC}       & \textbf{ASR}        & \textbf{ACC}        \\ \hline
\multirow{3}{*}{\textbf{5\%}}                                                        & \textbf{GCN}                         & 84.37               & 71.84              & 72.47               & 68.44               \\
                                                                                     & \textbf{GAT}                         & 85.98               & 69.23              & 79.46               & 68.29               \\
                                                                                     & \textbf{GIN}                         & 42.45               & \cellcolor[HTML]{caedfb}\textbf{59.65}              & 41.08               & \cellcolor[HTML]{caedfb}\textbf{62.66}               \\ \hline
\multirow{3}{*}{\textbf{10\%}}                                                       & \textbf{GCN}                         & 79.72               & 72.45              & 69.58               & 73.16               \\
                                                                                     & \textbf{GAT}                         & 63.34               & 72.94              & 52.02               & 69.86               \\
                                                                                     & \textbf{GIN}                         & 35.82               & 63.61              & 30.87               & 63.28               \\ \hline
\multirow{3}{*}{\textbf{15\%}}                                                       & \textbf{GCN}                         & 72.09               & 73.80              & 47.70               & 69.74               \\
                                                                                     & \textbf{GAT}                         & 58.38               & 72.29              & 31.58               & 71.53               \\
                                                                                     & \textbf{GIN}                         & 27.34               & 64.87              & \cellcolor{lightgreen}\textbf{23.44}               & 72.76               \\ \hline
\multirow{3}{*}{\textbf{20\%}}                                                       & \textbf{GCN}                         & 63.42               & 74.48              & 59.21               & 71.84               \\
                                                                                     & \textbf{GAT}                         & 38.93               & 72.45              & 23.49               & 73.84               \\
                                                                                     & \textbf{GIN}                         & \cellcolor{lightgreen}\textbf{22.86}               & 69.88              & 20.95               & 69.06               \\ \hline

\multicolumn{4}{l}{\small $\bullet$ Before (in average): ASR (93.88\%), ACC (74.42\%).}\\
\end{tabular}
\vspace{-3mm}
\end{table}

\textbf{Shortcomings of distillation methods.} Another line of work involves using a small proportion of clean data to fine-tune a backdoored model as a teacher model. Then use the knowledge of the teacher model to purify the backdoor \cite{ge2021anti,zhu2023enhancing,zhu2023adfl}. Ideally, with sufficient clean data, fine-tuning and distillation are almost akin to retraining the model, making it relatively straightforward to eliminate potential backdoors. However, in real-world scenarios, we can not access to the original training set and can only access limited testing or validation data. Under such restrictions, directly using the original representation of the model as the knowledge is insufficient to completely erase the backdoor since the representation difference between the backdoored and the fine-tuned models on clean data is minimal. To tackle this problem, researchers have utilized attention maps to characterize intermediate layers and obtain higher-quality knowledge to distillate \cite{li2021neural,ge2021anti,Gong2023RedeemMyself,zhu2023adfl,zhang2024badcleaner}. 
However, such attention mechanisms are generally designed for Euclidean structured data processed by convolutional neural networks, which is not applicable to GNNs. Considering that GNNs have strong representation capabilities and naturally contain attention (such as GAT \cite{GAT}), we explored whether we can directly use the intermediate representation of GNNs. As illustrated in Table \ref{T-fine-tining}, empirical results show that no matter what type of GNN structure is used, at least 15$\%$ of clean data is required to reduce the attack success rate (ASR) to 25$\%$, which is impractical in real scenarios. \textit{Therefore, designing a new graph attention mechanism that can obtain higher-quality intermediate-layer representations for GNNs would be key to making distillation-based backdoor defense methods more effective.}

\textbf{Challenges.} In summary, there are following challenges in the application of distillation-based defense in GNNs: \ding{182} Traditional attention map methods are tailored for data exhibiting translation invariance, which cannot be directly applicable to graph-structured data. Consequently, \textit{how to design an attention representation that captures both feature and topological information becomes a critical question.} \ding{183} Since the unrealistic assumption that existing methods require a large amount of clean data, \textit{how to use the limited clean data to distill more knowledge to achieve effective backdoor erasing while ensuring the main task accuracy is another key issue.}

\textbf{Our work.} In this work, we introduce \sysname, a novel backdoor mitigation method, for defending backdoor attacks in GNNs. Drawing inspiration from knowledge distillation for fine-tuning and the concept of neural attention transfer in the CV domain, \sysname specifically fine-tunes the model using a small subset of clean data, treating it as the teacher model. Subsequently, we propose the notion of attention representation in GNNs, aligning the intermediate-layer attention representations of the backdoored model with that of the teacher model through a graph attention transfer method, inducing backdoor neurons to transform into benign neurons. Moreover, we consider not only the consistency between corresponding layers but also the congruence of the relationships between layers. By leveraging the attention relation congruence, \sysname effectively enhances the layer transformation of the backdoored model to be consistent with the clean model, ensuring the accuracy of the main task.

\textbf{Extensive evaluations.} We assessed \sysname against three advanced attack methods on four commonly used datasets. Specifically, we adapted some SOTA defense techniques from the CV domain and implemented tailored graph backdoor mitigation methods as our baseline approaches. The experimental results demonstrate that our method outperforms existing defense mechanisms, reducing the ASR to below 5$\%$ using only 3$\%$ of clean test data, while maintaining an accuracy decrease within 5$\%$. Remarkably, even using a publicly available dataset with a different distribution as auxiliary data, the ASR can be reduced to below 10$\%$. Furthermore, we investigated the robustness of our method across different backdoor settings such as trigger size, backdoor injection ratio, and varying model structures. Additionally, we conducted a parameter sensitivity analysis on the attention distillation mechanism, visually interpreting attention maps to gain a deeper understanding, and extensively carried out ablation experiments to analyze the role of each module. Finally, we validated the effectiveness of our method against adaptive attacks. We are committed to open-sourcing our code to foster research within the community.
        
\textbf{Contributions.} Our key contributions include:

\begin{itemize}
    \item We propose a novel defense framework named \sysname, which employs the guidance of teacher model to facilitate the distillation process aimed at eliminating backdoors from compromised model. Our framework introduces an attention representation that is suitable for graph-structured data.
    \item We present graph attention transfer to align layers and employ attention relation congruence to fully exploit the relationships between attention maps. It achieves effective backdoor mitigation (below to $5\%$) by leveraging only a small fraction of the clean testing dataset (less than $3\%$).
    \item We perform extensive experiments of \sysname on four general datasets. Evaluation results demonstrate that \sysname significantly surpasses the performance of SOTA methods and also appears effective under various settings. 
\end{itemize}

%% file: data/2-background.tex
\section{Background}

\subsection{Graph Neural Networks}

Given a graph $G=(V,E)$ with node feature vectors $X_v^d$ for $v\in V$, where $V=\{v_1,v_2,...,v_N\}$ signifies the set of nodes, $E$ denotes the set of edges, and $A \in \mathbb{R}^{N \times N} $ is the adjacency matrix. For any two nodes $v_i,v_j\in V$, if $(v_i, v_j) \in E$, it implies the presence of an edge between  $v_i$ with $v_j$ and $A_{ij} = 1$; otherwise, $A_{ij} = 0$. Here,  $N = |V|$ denotes the total number of nodes, and $d$ is the number of node feature dimensions.

The goal of graph learning is to acquire effective node representations by leveraging both structural information and node feature information through the mechanism of message propagation. Modern GNNs employ a strategy known as neighborhood aggregation. This approach entails the iterative update of a node's representation through the aggregate of its adjacent nodes' features. Upon completing $k$ iterations of this aggregation process, the node's representation captures the intricate structural details extending up to $k$ levels in its network vicinity. Formally, the $k$-th layer of a GNN can be defined as follows:
\begin{equation}
	\begin{split}
    a_v^{(k)}=\text{AGGREGATE}^{(k)}\left(\left\{h_u^{(k-1)}:u\in\mathcal{N}(v)\right\}\right),
	\end{split}
	\label{EAGGREGATE}
\end{equation}
where the $\text{AGGERGATE}$ function depends on the adjacency matrix $A$ and $\mathcal{N}(v)$ is a set of nodes adjacent to $v$.

One of the most prevalent models in graph learning is the Graph Convolutional Neural Network (GCN). Following the prior work \cite{GCN}, we define the graph convolutional layer to be 
\begin{equation}
	\begin{split}
     Z^{(l)}=\sigma(\tilde{D}^{-\frac{1}{2}}\tilde{A}\tilde{D}^{-\frac{1}{2}}Z^{(l-1)}W^{(l)}),
	\end{split}
	\label{EGCN}
\end{equation}
where $Z^{(l)}$ is the convolutional activations at the $l$-th layer, and initially, $Z^{(0)}=X$. $\tilde{A}=A+I$ is the adjacency matrix with added self-connections where $I\in\mathbb{R}^{n\times n}$ is the identity matrix. $\sigma(\cdot)$ is the element-wise nonlinear activation function such as $ReLU(\cdot)=\max(0,\cdot)$. $\tilde{D}_{ii}=\sum_j\tilde{A}_{ij}$ and $W^{(l)}$ is the trainable parameter for the $l$-th GCN layers.

\subsection{Backdoor Attacks on GNNs}
Whether in the fields of CV, NLP, or graph learning, backdoor attackers share a common goal: ensuring that a backdoored model outputs a target label for any input embedded with the trigger while maintaining normal behavior on benign inputs. Formally, let $F$ represent the clean model and $\tilde{F}$ represent the backdoored model. The attacker's objective can be defined as follows:


\begin{equation}
	\begin{split}
\left.\left\{\begin{array}{l}F(x)=\tilde{F}(x)\\\tilde{F}(x\oplus \delta)=y_t\end{array}\right.\right.,
	\end{split}
\end{equation}
where \( \delta \) denotes the backdoor trigger and $y_t$ denotes the backdoor target label. The training objective can then be expressed as:  
\begin{equation}
	\begin{split}
\min_{\mathbf{w}}&\underbrace{\mathbb{E}_{(x,y)\sim\mathcal{D}_{clean}}\ell(F_{w}(x), y)}_{\text{Average loss on clean data}}\\&+\lambda\cdot\underbrace{\mathbb{E}_{(x^{\prime}, y_{t})\sim\mathcal{D}_{poison}}\ell(F_w(x^{\prime}), y_{t})}_{\text{Average loss on poisoned data}}.
	\end{split}
\end{equation}

\begin{figure}
	\centering
	\includegraphics[width=1\linewidth]{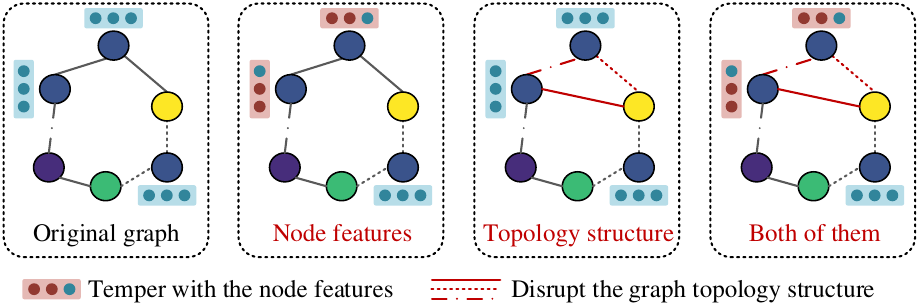}
	\caption{Different trigger injection types on graph.}
	\Description{Different trigger injection types on graph.}
	\label{FBA}
\end{figure}

Unlike image data, graph data includes both node features and topological structures, leading to distinct methods for constructing graph triggers. As shown in Fig. \ref{FBA}, these methods can be broadly categorized into three types:  

\begin{itemize}
    \item \textbf{Tampering with node features}: The attacker alters the embeddings of important node features within the graph, leading the model to learn incorrect representations of these nodes. 
    \item \textbf{Disrupting the topology structure}: The attacker modifies the graph's topology by adding or removing edges, thereby introducing structural perturbations that mislead the model.  
    \item \textbf{Modifying both node features and topology structure}: The attacker generates a composite trigger by simultaneously modifying node features and the topology structure. This trigger can be directly inserted into or replace corresponding components in the original graph.  
\end{itemize}

Such various backdoor attacks highlight the complex nature of backdoor attacks on graph data and emphasize the need for targeted defense mechanisms in GNNs.

\subsection{Knowledge Distillation}


Knowledge distillation was initially developed for model compression \cite{hinton2015distilling}, diverging from pruning and quantization methods. Utilizing the Teacher-Student architecture, it transfers knowledge from a large, complex model to a smaller student model, thereby enhancing the student’s generalization and achieving superior performance and precision. Generally, the distillation loss for knowledge distillation can be formulated as
\begin{equation}
	\begin{split}
        L_{KD}\big(f_t(x),f_s(x)\big)=\mathcal{L}_F\big(\Phi_t(f_t(x)),\Phi_s(f_s(x))\big) ,
	\end{split}
	\label{EKD}
\end{equation}
where $f_t(x)$ and $f_s(x)$ are the feature maps of the intermediate layers of teacher and student models, respectively. The transformation functions, $\Phi_t(f_t(x))$ and $\Phi_s(f_s(x))$, are usually applied when the feature maps of teacher and student models are not in the same shape. $\mathcal{L}(\cdot)$ indicates the similarity function used to match the feature maps of teacher and student models. Common loss functions are $L^2$-norm distance, cross-entropy loss and KL divergence.

Recently, researchers have proposed feature-based distillation \cite{komodakis2017paying,kim2018paraphrasing,passban2021alp} and relation-based distillation \cite{zhang2020distilling,minami2020knowledge,wang2019binarized} techniques. Among these, the attention mechanism \cite{komodakis2017paying}, re-characterizes intermediate layer knowledge, reflecting the neurons responsible for meaningful regions based on each neuron's contribution to the prediction outcomes. This mechanism is pivotal for acquiring high-quality knowledge from intermediate layers and has been widely used for distillation-based defense methods. \cite{li2021neural,Gong2023RedeemMyself}. However, 
existing research referred to graph distillation has primarily focused on designing various mechanisms to fully exploit the knowledge of different layers \cite{zhang2020reliable,zhang2022graphless} and to extract relationships between them \cite{yao2020graph,song2021scgcn,qian2021distilling}. Lacking a method in GNNs that can capture high-quality intermediate-layer representations to achieve effective graph distillation. In this paper, we devise an attention representation method for the intermediate layers of graphs, rendering knowledge distillation-based backdoor defense effective in GNNs.

%% file: data/3-threat_model.tex
\section{Threat Model}

\textbf{Attacker's goals.}
Backdoor attacks are designed to stealthily embed harmful functions within a target model in the form of hidden neural trojans. These trojans lie dormant until specific pre-defined patterns, known as triggers $\delta$, are detected. Upon encountering these triggers, the backdoor is activated, causing the model to deviate from its intended behavior and produce outputs that align with the attacker's objectives $y_t$.

\textbf{Attacker's capability.}
Depending on different backgrounds and knowledge of the attack, we define attackers' capabilities into two cases: \ding{182}Attackers poison only a segment of the data, altering its associated labels $y_t$, and then disseminate this corrupted dataset. They anticipate that users will integrate these poisoned samples into their training sets, thereby inadvertently introducing a backdoor into the models they develop.
\ding{183}Attackers gain access to a legitimate model and utilize an auxiliary dataset to transform it into a backdoored version.
Regardless of the attacker's specific capabilities, their ultimate aim is to possess a model $\tilde{F}$ compromised by a backdoor.
In our considerations, we assume a powerful attacker. This attacker provides the trained model to the defender, retaining full knowledge of the model's internal mechanisms and the dataset used for training. With this deep understanding, the attacker has the ability to manipulate the model $F$ in any manner necessary to create a backdoored version $\tilde{F}$. The trigger $\delta$ for activating the backdoor can assume various forms, locations, and sizes, showcasing the attacker's versatility and adaptability.

\textbf{Defender's capability and goal.}
Building upon previous studies, we refined the simulation of real-world scenarios by imposing limitations on the defender's capabilities: \ding{182}The defender is kept unaware of both the tainted graph identities and the attacker's intended target label. \ding{183}The defender is granted access to only a portion of the dataset, thus precluding full possession of the training data. In an even more stringent scenario, the defender's resources are limited to a small subset sourced from publicly available auxiliary datasets. We assume that the defender obtains a trained model from an untrusted third party, the goal of the defender is to decrease the success rate of backdoor attacks while maintaining the performance of the model on regular tasks.

%% file: data/4-method.tex
\section{Design of \sysname}
\label{Method}

\subsection{Defense Intuition and Challenges}

Our objective is to mitigate the backdoor embedded within the backdoored model, intending to restore the performance of the backdoored model to a level comparable to what would be achieved by training directly on the original clean dataset. Inspired by advancements in backdoor defense within the CV domain, a straightforward and intuitive method is known as neuronal attention distillation. This technique employs the finetuned backdoor model as a teacher to guide the original backdoored student network on a small subset of clean data. It introduces the attention map as intermediate-layer knowledge and enforces the intermediate-layer attention of the student well aligned with that of the teacher, thus purifying the backdoor neurons. However, applying this distillation to defend against graph-based backdoors still poses the following challenges:

\textbf{\textit{Challenge 1: how to find the proper attention representations in graphs to distill?}}

Although attention alignment has emerged as a standard component in knowledge distillation, its adaptation and refinement for GNNs remain largely uncharted territory. Utilizing attention mechanisms designed for image data in a graph context overlooks the unique topological structure inherent to graphs. Consequently, there is a pressing requirement to develop a specialized approach for calculating graph attention that ensures the rich structural information within graphs is not only preserved but also leveraged effectively during the distillation process.

\textbf{\textit{Challenge 2: how to improve distillation with limited data?}}

Both fine-tuning and distillation can degrade the accuracy of the main task under limited data constraints. Moreover, our objective is to sanitize backdoored models, transforming them into clean models, while concurrently maintaining high classification accuracy. This intricate demand adds an additional layer of complexity to the challenge at hand, necessitating a more nuanced and precise approach to the distillation process.

\subsection{Overview}

\begin{figure*}[htbp]
	\centering
	\includegraphics[width=0.98\linewidth]{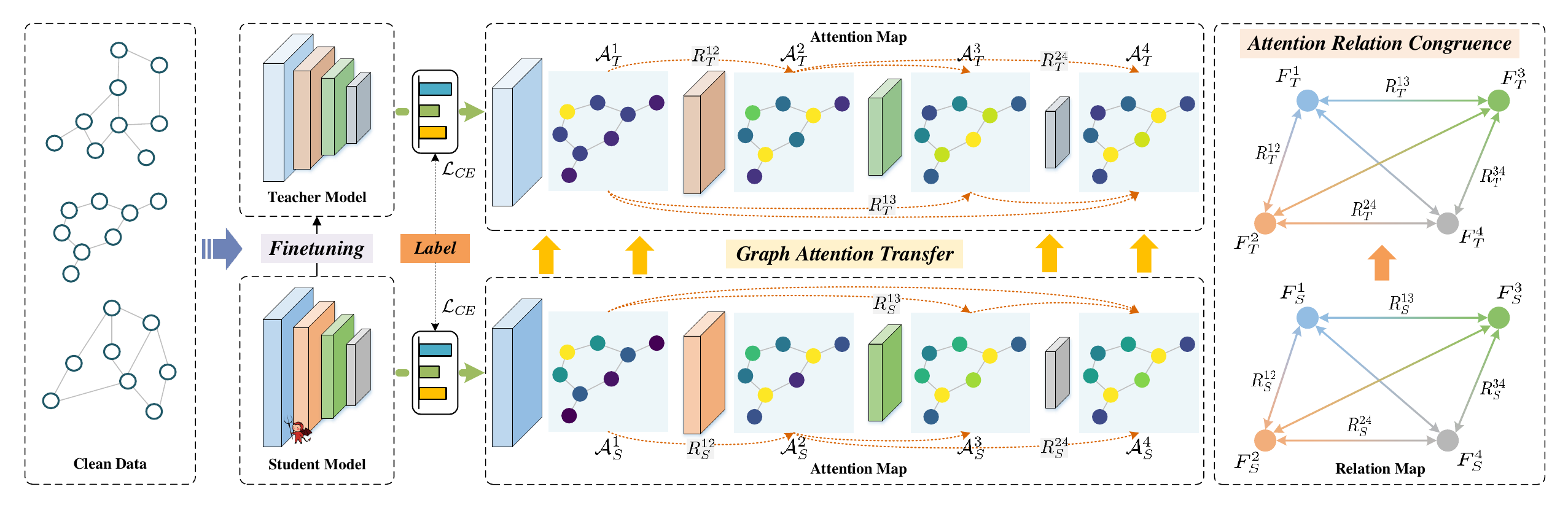}
	\caption{Framework of the proposed \sysname.}
	\Description{Framework of the proposed \sysname} 
	\label{framework}
	\vspace{-5mm}
\end{figure*}

Fig. \ref{framework} presents the framework of our proposed \sysname. By leveraging limited clean data, \sysname first fine-tunes a backdoored model as a teacher model and uses the backdoored model as the student model for knowledge distillation. During the distillation process, \textbf{\textit{graph attention transfer}} is performed to identify proper attention representations in graphs and align the intermediate-layer attention of the backdoored model with that of the teacher model to purify abnormal neurons. Furthermore, \sysname employs \textbf{\textit{attention relation congruence}} to fully exploit the relationships between attention maps, enforcing the layer transformation of the backdoored model consistent with that of the teacher model, guaranteeing a high model performance.

\subsection{Graph Attention Transfer}
\label{W-attention}

In the realm of CV, backdoor mitigation via distillation succeeds by crafting a knowledge representation in the middle layers akin to human attention, emphasizing the input's focus region for a particular model layer. Usually, the normal model focuses on the key features of the input sample, while the backdoor model focuses on the abnormal region. Our core idea is to achieve backdoor mitigation by correcting the attention of the backdoor model, that is, adjusting the focus region of the backdoor model to the normal region.

Given a GNN model $F$, we denote the activation tensor at the $l$-th layer as $F\in \mathbb{R}^{C_l\times N}$, which $C_l$ and $N$ as the output channel dimension and node numbers of the the $l$-th layer respectively. We define a function $\mathcal{A}: \mathbb{R}^{C_l\times N } \to \mathbb{R}^{N}$, which maps an activation map to an attention representation. In the map, we count the contribution of each element to the final classification results. Considering the topological properties of the graph, we calculated the degree for each node $D=deg(V)$. We define formulation of the attention operator $\mathcal{A}$ to be
\begin{equation}
	\begin{split}
        \mathcal{A}^p(F^l)=\frac{D}{max(D)} \sum_{i=1}^C \left | F^{l(i)} \right |^p,
	\end{split}
	\label{EQ-Attention}
\end{equation}
where $F^{l(i)}$ is the activation map of the $i$-th channel, $max(D)$ is the maximum value in $D$, $|\cdot|$ is the absolute value function and $p>1$. By adjusting the value of $p$, we can further amplify the differences between the backdoor neurons and benign neurons. Experiments have provided insights into the impact of different $p$ values on defensive performance.

We use the distilled attention knowledge to guide the purification of the model. To ensure a consistent representation of attention representations across different layers, the normalization function $\psi(\cdot)$ is applied to the attention representation, which is a pivotal step in the success of the distillation process. Formally,
\begin{equation}
	\begin{split}
        \psi(\mathcal{A}^l)=\frac{\mathcal{A}^l}{\left \| \mathcal{A}^l \right \|_2}.
	\end{split}
	\label{EQ2}
\end{equation}

To mitigate the impact of backdoor neurons, we align neurons susceptible to triggers with clean neurons. We measure the distributional discrepancies between attention maps of different neuron layers by computing the $L^2$-norm between the attention graphs of the student and teacher models. Formally,

\begin{equation}
	\begin{split}
\mathcal{L}_{AD}(F_T^l,F_S^l)=\left \|\psi(\mathcal{A}^l_T)-\psi(\mathcal{A}^l_S)\right \|_2.
	\end{split}
\end{equation}

\subsection{Attention Relation Congruence}
\label{W-relation}

By aligning the intermediate-layer attention features between the teacher and student models, the backdoor triggers can be effectively mitigated. Nonetheless, aligning the attention features at the same layer between the two models demands enough clean data to provide reliable and effective knowledge, thereby achieving a more thorough removal of backdoors. However, in practical scenarios, due to privacy concerns or various access limitations, the availability of clean data for fine-tuning and distillation is often restricted. Consequently, there is a requirement for supplementary knowledge to enhance the distillation process.

In light of the information propagation mechanism inherent in GNNs, the relation of transforms between different layers cannot be disregarded. Therefore, we consider not only the layer level congruence but also the relation congruence between layers. For relation matrix $R$, each element represents the correlation between $F^i$ and $F^j$ in attention representation, which is defined as
\begin{equation}
	\begin{split}
        R^{ij}=\frac{F^i-F^j}{\left \| F^i-F^j \right \| _2},\ \ \ \  R^{ij} \in \mathbb{R}.
	\end{split}
\end{equation}

Our objective is to measure the distributional discrepancies between teacher and student model pairs, with the intent of reducing these variances to a minimum. Drawing inspiration from generative adversarial network models \cite{Deshpande2019MaxslicedWasserstein}, we employ the Wasserstein distance to compute these disparities, which is a widely used metric for measuring the distributional differences between two potential distributions. The Wasserstein-p distance between distributions $R^{ij}_T$ and $R^{ij}_S$ is defined as follows:
\begin{equation}
    \begin{split}
    W_p(R^{ij}_T,R^{ij}_S)=\inf_{\gamma\in\Pi(R^{ij}_T,R^{ij}_S)}(\mathbb{E}_{(x,y)\thicksim\gamma}[||x-y||^p])^{\frac1p},
    \end{split}
\end{equation}
where $\Pi(R^{ij}_T, R^{ij}_S)$ represents the set of all possible joint distributions over the points $(x, y)$.

Given the constraint of utilizing only a small portion of the test dataset, to ensure that the model fully absorbs all the knowledge from the teacher model, we forego the use of attention mechanisms and instead employ the representations directly. This also implies that our data has a high dimensionality. Estimating Wasserstein distance on high-dimensional data is not straightforward. To alleviate the computational complexity, we adopt the sliced version of the Wasserstein-2 distance, which requires estimating the distance between one-dimensional distributions, thus enhancing efficiency. Formally,
\begin{equation}
    \begin{split}
    \tilde{W}_2(R^{ij}_T,R^{ij}_S)=\left[\int_{0}^1W_2^2(R^{ij}_T(\omega),R^{ij}_S(\omega))d\omega\right]^{\frac12},
    \end{split}
\end{equation}
where $R^{ij}_T(\omega)$ and $R^{ij}_S(\omega)$ respectively represent the distributional relationship embeddings between layers $i$ and $j$ of the teacher and student models, projected onto one-dimensional data points along the slicing direction $\omega$.

In the relation congruence loss, for each pair of layers chosen within our model, we refer to these as ``pairs''. For instance, in a 4-layer model, $pairs=\{<i,i+1>\}$ would represent considering the relationships between $\{<1,2>,<2,3>,<3,4>\}$. Particularly, when $pairs=\{full\}$ is employed, it signifies the inclusion of all inter-layer relationships. In the experimental section, we will explore which selections of pairs lead to the most effective defensive performance. The attention relation congruence loss $\mathcal{L}_{RC}$ is denoted as
\begin{equation}
    \begin{split}
    \mathcal{L}_{RC} = \sum_{i,j\in <pairs>}\tilde{W}_2(R^{ij}_T,R^{ij}_S).
    \end{split}
\end{equation}

Using only the attention distillation loss $\mathcal{L}_{AD}$ and the relation congruence loss $\mathcal{L}_{RC}$ during retraining will diminishes the prediction accuracy of clean samples. The rationale behind this decline is that attention distillation focuses solely on narrowing the gaps between the attention maps across various layers, without ensuring that the aligned attention maps align with the accurate prediction outcomes. So, we use the cross entropy (CE) loss $\mathcal{L}_{CE}$ to guarantee the precision of prediction results for clean samples.
Then, the optimization goal of \sysname is to minimize the following loss function:
\begin{equation}
	\begin{split}
\mathcal{L} = \mathcal{L}_{CE}\left(F_S\left(x\right),y\right) + \beta \sum^k_{l=1}\mathcal{L}_{AD}\left(F_T^l\left(x\right),F_S^l\left(x\right)\right)+ \gamma \mathcal{L}_{RC},
	\end{split}
\end{equation}
where $\beta$ and $\gamma$ is a hyperparameter that balances distillation and prediction accuracy preservation. The overall algorithm of \sysname is summarized in Algorithm \ref{Aprocess} (see Appendix \ref{A-Algorithm}).

%% file: data/5-evaluation.tex
\section{Evaluation}

In this section, we evaluate \sysname under targeted  attacks for graph-level classification tasks from different perspectives. First, we introduce  experimental settings and compare its effectiveness with SOTA methods. Then, we measure the performance of \sysname under various backdoor settings and investigate the mechanism behind \sysname in depth. Finally, we do ablation studies to find out how the modules influence the performance of \sysname.

\subsection{Experimental Setup}
\subsubsection{Graph Datasets}
We evaluate our approach on four real-world datasets, where basic statistics for each dataset are shown in Table \ref{TDataset}, including the graph numbers in the dataset, the average number of nodes per graph, the average number of edges per graph, the number of classes, the number of graphs in each class, and the target class for attacks. These datasets are class-imbalanced, which makes defense more difficult. More details are in Appendix \ref{A-Dataset}.

\begin{table*}[t]
\centering
\caption{Dataset statistics.}
\renewcommand\tabcolsep{5pt}
	\renewcommand\arraystretch{1.1}
\begin{tabular}{ccccccc}
 \hline \hline
\textbf{Datasets} & \textbf{\# Graphs} & \textbf{Avg.\# Nodes} & \textbf{Avg.\# Edges} & \textbf{\# Classes} & \textbf{\# Graphs[Class]}     & \textbf{\# Target Label} \\ \hline 
PROTEINS       & 1113              & 39.06                 & 72.82                 & 2                  & 663[0], 450[1] & 1                       \\
COLLAB            & 5000              & 73.49                & 2457.78              & 3                  & 2600[0],775[1],1625[2]       & 1                       \\ 
AIDS              & 2000              & 15.69                & 16.2                 & 2                  & 400[0],1600[1]               & 0                       \\
Fingerprint       & 1661              & 8.15                 & 6.81                 & 4                  & 538[0], 517[1],109[2],497[3] & 1                       \\
\hline \hline 
\end{tabular}
\label{TDataset}
\end{table*}

\subsubsection{Attack Methods}
We investigate three SOTA backdoor attack methods: Subgraph-based Backdoor (Sub-BA) \cite{badsub}, GTA \cite{GTA}, and Motif-Backdoor (Motif-BA) \cite{motifBA}. Those attack methods are briefly described in Appendix \ref{A-Attack}.

\subsubsection{Models}
To assess the defensive capabilities of \sysname, we have selected four popular GNN models, i.e., Graph Convolutional Networks (GCN) \cite{GCN}, Graph Isomorphism Network (GIN) \cite{GIN}, Graph Attention Networks (GAT) \cite{GAT}, and Graph Sample and Aggregate (GraphSAGE) \cite{GSA} as the target models. We utilize the openly accessible codebases for these models. Note that GIN is the default model unless otherwise mentioned.

\subsubsection{Baseline}
We implement the backdoor defense strategies Prune \cite{dai2023unnoticeable} and randomized smoothing (RS) \cite{Wang2021CertifiedRobustnessGraph}. Additionally, we transfer the anti-backdoor learning (ABL) \cite{Li2021AntibackdoorLearning} to graph fields, which is a popular backdoor defense method in CV domains.

\subsubsection{Evaluation Metrics}
We evaluate the performance of defense mechanisms with two metrics: \ding{182}attack success rate (ASR), which is the ratio of backdoored inputs misclassified by the backdoor model as the target labels specified by attackers:
\begin{equation}
\begin{split}
    \text{Attack Success Rate (ASR)}=\frac{\#\text{successful attacks}}{\#\text{total trials}},
\end{split}
\label{EASR}
\end{equation}
and \ding{183}the accuracy of the main classification task on normal samples (ACC). An effective defense method means significantly reducing the ASR while maintaining a high ACC.

\subsubsection{Implementation Details}
All experiments are implemented \sysname in Python using the PyTorch framework. The experimental environment consists of 13th Gen Intel(R) Core(TM) i7-13700KF, NVIDIA GeForce RTX 4070 Ti, 32GiB memory, and Ubuntu 20.04 (OS). The data was divided into a training set and a testing set in a ratio of 80:20, and \sysname is assumed to be able to access 3$\%$ of the clean data randomly selected from the testing set. The fine-tuning process was conducted for a total of 10 epochs. For the distillation process, the loss term is set to $\beta = \gamma = 1$, the batch size is $B = 64$, SGD is employed as the optimizer with a learning rate of $\eta = 0.001$, and the process is run for $E = 30$ epochs. For all the baseline attacks and defenses, we adopt the default hyperparameters recommended by the corresponding papers. Specifically, attacks have the common parameters: trigger size $t$ and injection ratio $\varphi$. Given a dataset $D_{train}$ with an average node number $N_{avg}$, the number of nodes in the subgraph trigger is equal to $N_{avg} \times t$. Unless otherwise mentioned, the backdoor injection ratio is set to $\varphi = 5\%$ and the trigger size $t$ is set to $20\%$. We test the performance of \sysname as well as other baselines five times and report the mean and standard deviation results to eliminate the effects of randomness.

\subsection{Experimental Results}

\begin{table*}[t]
\centering
\caption{Performance of \sysname compared with baseline methods on four datasets.}
\vspace{1mm}
\small
\renewcommand\tabcolsep{9pt}
	\renewcommand\arraystretch{1}
\begin{tabular}{cc|cc|cccccccc}
	\hline
	\multirow{2}{*}{\textbf{Datasets}}    & \multirow{2}{*}{\textbf{Attacks}} & \multicolumn{2}{c|}{\textbf{Before}} & \multicolumn{2}{c}{\textbf{ABL}} & \multicolumn{2}{c}{\textbf{Prune}} & \multicolumn{2}{c}{\textbf{RS}} & \multicolumn{2}{c}{\textbf{\sysname}} \\ \cline{3-12} 
	&                                   & \textbf{ASR}     & \textbf{ACC}     & \textbf{ASR}    & \textbf{ACC}   & \textbf{ASR}     & \textbf{ACC}    & \textbf{ASR}   & \textbf{ACC}   & \textbf{ASR}      & \textbf{ACC}      \\ \hline
\multirow{3}{*}{\textbf{PROTEINS}}    & \textbf{Sub-BA}                   & 96.34            & 68.42            & 74.39           & 68.95          & 90.65            & 62.33           & 85.37          & 57.37          & \cellcolor{lightgreen}1.47              & \cellcolor{lightgreen}71.34             \\
	& \textbf{GTA}                      & 95.52            & 73.25            & 5.00            & 61.05          & 97.14            & 40.81           & 92.54          & 57.96          & \cellcolor{lightgreen}1.49              & \cellcolor{lightgreen}72.61             \\
	& \textbf{Motif-BA}                 & 42.68            & 73.16            & 14.63           & 63.68          & 29.91            & \cellcolor{lightgreen}69.96           & 78.05          & 55.26          & \cellcolor{lightgreen}1.47              & 69.43             \\ \hline
\multirow{3}{*}{\textbf{COLLAB}}      & \textbf{Sub-BA}                   & 77.49            & 79.52            & 21.42           & \cellcolor{lightgreen}75.86          & 42.60            & 64.20           & 32.02          & 67.71          & \cellcolor{lightgreen}0.74              & 75.14             \\
	& \textbf{GTA}                      & 92.36            & 80.43            & 2.71            & 62.29          & 10.02            & 63.90           & 99.51          & 71.43          & \cellcolor{lightgreen}0.49              & \cellcolor{lightgreen}77.57             \\
	& \textbf{Motif-BA}                 & 81.46            & 79.71            & 24.64           & 70.79          & 22.97            & 66.64           & 39.48          & 71.55          & \cellcolor{lightgreen}0.49              & \cellcolor{lightgreen}74.00             \\ \hline
\multirow{3}{*}{\textbf{AIDS}}        & \textbf{Sub-BA}                   & 80.86            & 97.86            & 9.88            & 82.50          & 70.43            & 68.00           & 48.15          & 90.36          & \cellcolor{lightgreen}0.62              & \cellcolor{lightgreen}97.14             \\
	& \textbf{GTA}                      & 100.00           & 99.25            & 9.88            & 82.50          & 96.09            & 60.50           & 54.94          & 90.71          & \cellcolor{lightgreen}0.62              & \cellcolor{lightgreen}99.29             \\
	& \textbf{Motif-BA}                 & 77.50            & 98.21            & 21.88           & 81.79          & 24.89            & 83.50           & 23.13          & 95.71          & \cellcolor{lightgreen}0.63              & \cellcolor{lightgreen}98.93             \\ \hline
\multirow{3}{*}{\textbf{Fingerprint}} & \textbf{Sub-BA}                   & 91.67            & 61.62            & 17.31           & 48.59          & 64.80            & 42.64           & 87.82          & 42.61          & \cellcolor{lightgreen}0.79              & \cellcolor{lightgreen}59.83             \\
	& \textbf{GTA}                      & 99.13            & 60.26            & 44.35           & 41.88          & 99.38            & 30.33           & 59.13          & 57.69          & \cellcolor{lightgreen}3.48              & \cellcolor{lightgreen}58.97             \\
	& \textbf{Motif-BA}                 & 100.00           & 61.11            & 23.33           & 50.00          & 75.71            & 54.95           & 44.17          & 51.76          & \cellcolor{lightgreen}1.03              & \cellcolor{lightgreen}59.40             \\ \hline
\end{tabular}
\label{Tall}
\end{table*}

\subsubsection{Comparison}

Table \ref{Tall} showcases a comparative analysis of \sysname's performance against three baseline methods across four distinct datasets. In the ``Before'' column, the results of attacks without any defense mechanisms are displayed. The best results are highlighted in \colorbox{lightgreen}{Green} color. Overall, our devised backdoor defense mechanism boasts the lowest ASR and better clean accuracy across all datasets and attack methods. On the one hand, \sysname achieves an ASR below $5\%$ across all datasets, indicating a robust mitigation strategy that effectively curtails the model's susceptibility to backdoor contamination. On the other hand, \sysname maintains a negligible reduction in ACC, as compared to an unprotected model, demonstrating that our system preserves the precision of clean node predictions, thereby ensuring the model's continued functionality.

In particular, although the ABL approach effectively eliminates the backdoor, it concurrently causes a notable decline in the accuracy of the primary task. We suspect this discrepancy arises from the distinct structural differences between image and graph data. The unique information dissemination process inherent in graph data means that simply eradicating the triggered features through gradient ascent can inadvertently result in the loss of essential features. This, in turn, impairs the model’s overall performance on the primary task. The Prune technique is capable of identifying nodes with notably low cosine similarity to the features of their adjacent nodes and subsequently removing them. Nonetheless, attackers typically target nodes with similar characteristics for their malicious injections. Moreover, advanced attacks like GTA can further alter the features of the compromised nodes to mimic others, rendering Prune ineffective. RS has demonstrated a degree of efficacy on certain datasets by randomly eliminating nodes and edges, yet this comes at the cost of reduced accuracy. 

In contrast, our method attains comparable defense outcomes by integrating graph attention transfer and attentional relation congruence strategies, which not only yield the highest rate of backdoor removal but also minimize accuracy loss. Ablation studies have confirmed the substantial contribution of these strategies to improved prediction accuracy. In certain scenarios, \sysname can even surpass the prediction accuracy of clean models, enhancing the performance of backdoored models. According to our analysis, this is attributed to the fact that our Attention Relation Congruence can extract the relation maps and enforce layer transformation of the backdoored model to be consistent with that of the clean model, thereby ensuring model accuracy. Subsequent ablation experiments also prove this. Other standard defense mechanisms either fail to preserve high prediction accuracy or are insufficient in lowering the success rate of backdoor attacks. Furthermore, we examine the impact of distillation on the accuracy of class-specific categorization in Appendix \ref{A-various_cate}.

\subsubsection{Impact of Injection Ratio} 
\label{N_injection_ratio}
\begin{figure}
	\centering
	\subfigure[ASR with Sub-BA method.]{
		\begin{minipage}[t]{0.485\linewidth}
		\centering
			\includegraphics[width=1\linewidth]{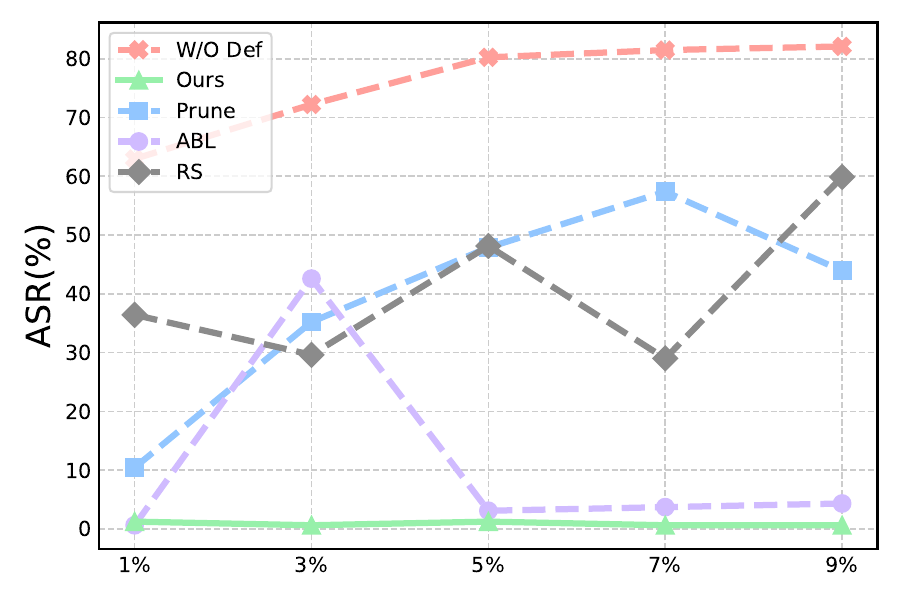}
	\end{minipage}}
        \subfigure[ACC with Sub-BA method.]{
		\begin{minipage}[t]{0.485\linewidth}
		\centering
			\includegraphics[width=1\linewidth]{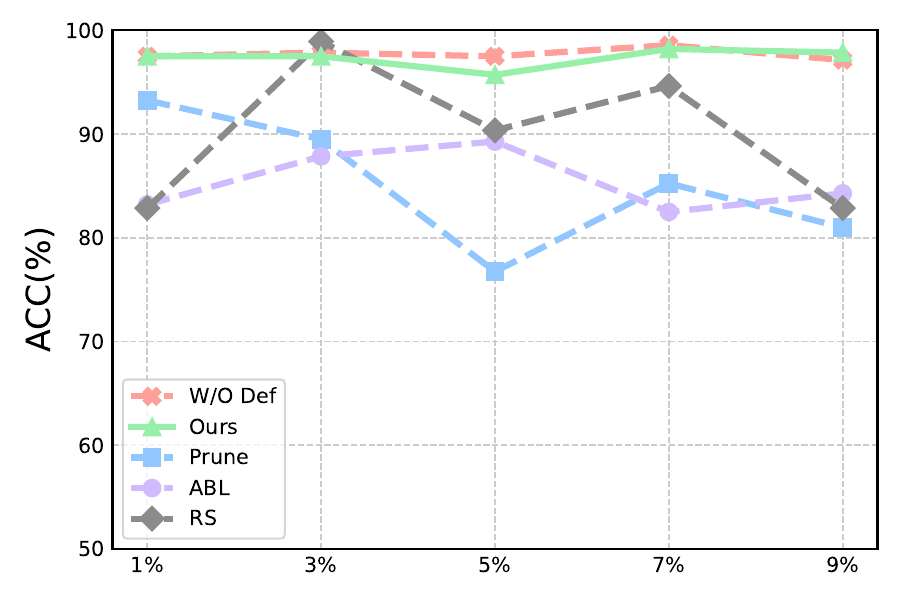}
	\end{minipage}}
 \\
	\subfigure[ASR with GTA method.]{
		\begin{minipage}[t]{0.485\linewidth}
			\includegraphics[width=1\linewidth]{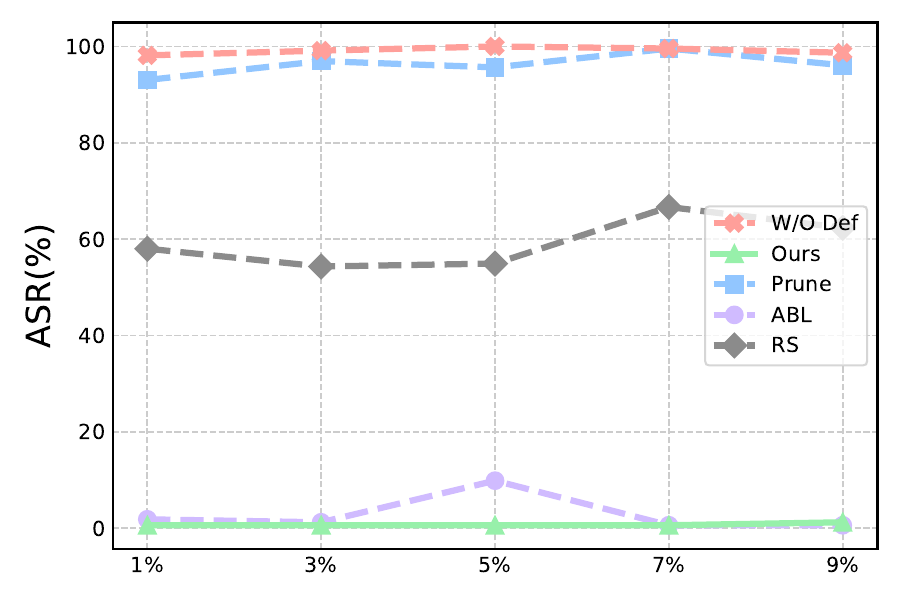}
	\end{minipage}}
	\subfigure[ACC with GTA method.]{
		\begin{minipage}[t]{0.485\linewidth}
			\includegraphics[width=1\linewidth]{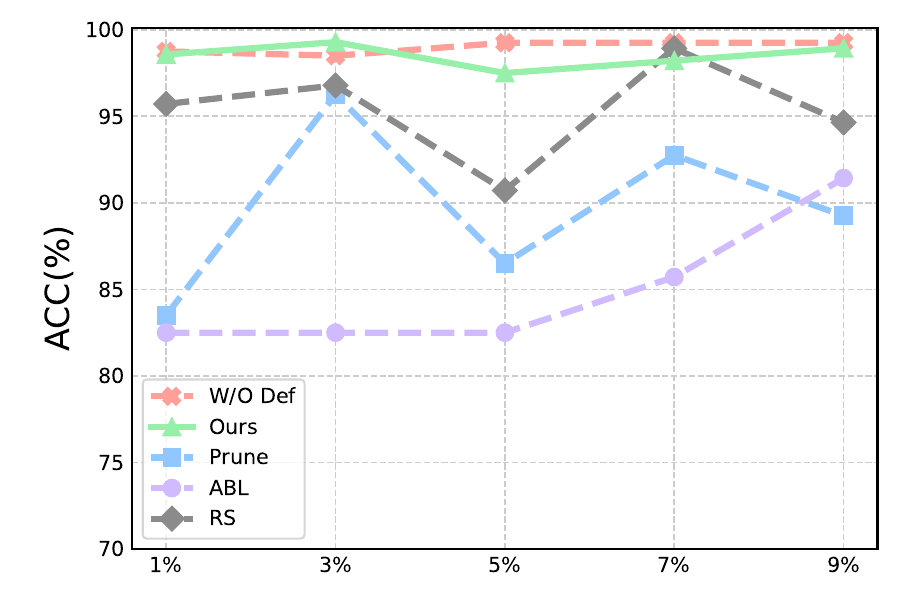}
	\end{minipage}}
 \\
	\subfigure[ASR with Motif-BA method.]{
		\begin{minipage}[t]{0.485\linewidth}
			\includegraphics[width=1\linewidth]{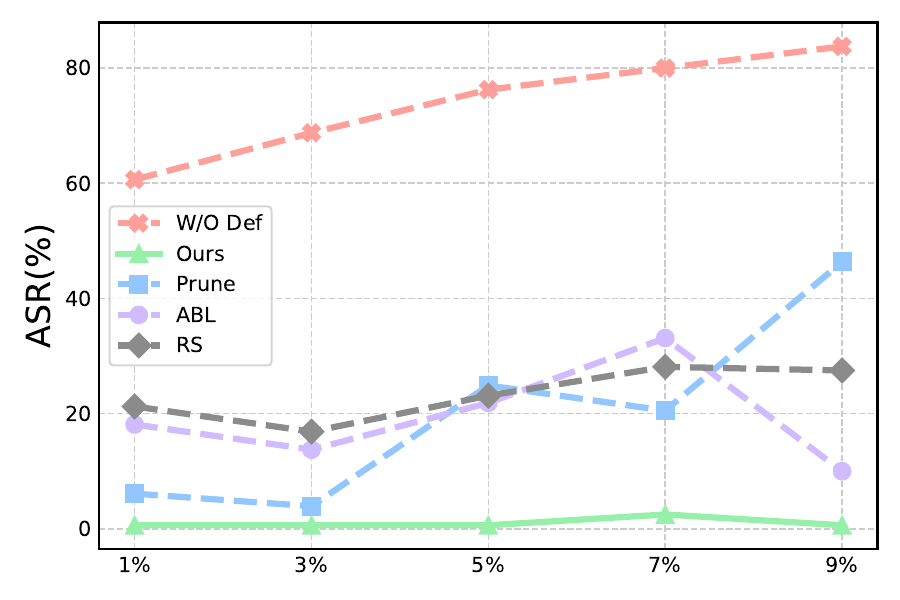}
	\end{minipage}}
	\subfigure[ACC with Motif-BA method.]{
		\begin{minipage}[t]{0.485\linewidth}
			\includegraphics[width=1\linewidth]{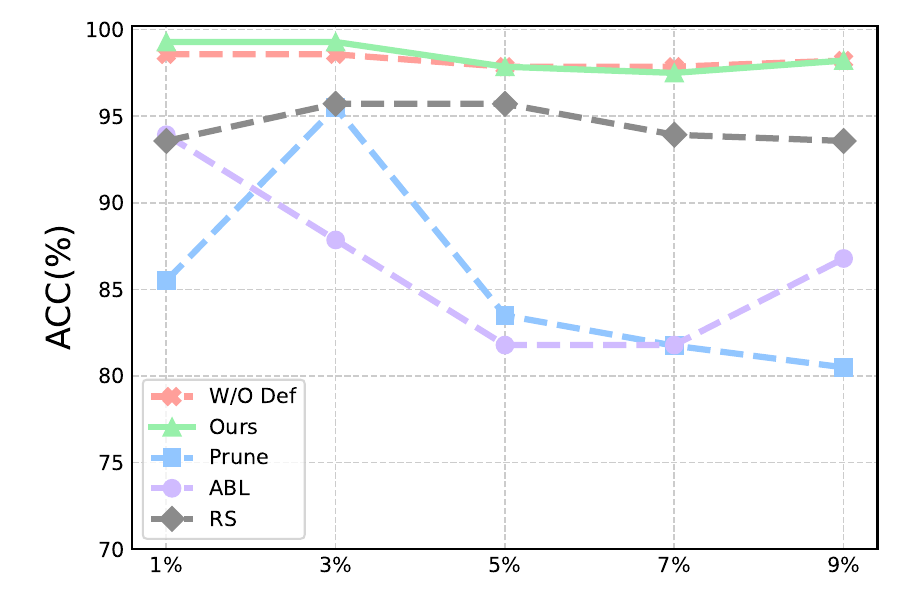}
	\end{minipage}}%
	\caption{Impact of injection ratio with three SOTA methods on AIDS dataset.}
	\Description{Impact of injection ratio with three SOTA methods on AIDS dataset.}
	\label{F_injection_ratio}
\end{figure}

Given that the defender has sole access to the trained backdoor model, the proportion of backdoor injections within the training dataset remains unknown. To assess the robustness of our technique against backdoor incursions with varying proportions, we have benchmarked the efficacy of diverse defense strategies against three distinct attacks utilizing the AIDS dataset. The outcomes are depicted in Figure \ref{F_injection_ratio}, where the injection rate fluctuates between $1\%$ and $9\%$, increasing in $2\%$ increments. The results for other datasets can be reviewed in the Fig. \ref{F-injection_ratio_Fingerprint} and Fig. \ref{F-injection_ratio_PROTEINS} (see Appendix \ref{A_injection-ratio}).

Our findings indicate that, without the application of any defensive measures, the ASR escalates in tandem with the intensity of backdoor injections, creating a balancing act between the infusion of more contaminated data and a decline in the primary task’s accuracy. In general, the greater the proportion of backdoor injections, the more formidable the challenge of backdoor defense becomes. Significantly, when the backdoor injection rate surpasses $5\%$, our method stands out with superior defensive capabilities, as the adverse effects on the main task are nearly imperceptible. This robust performance can be attributed to the higher success rate of backdoor attacks at elevated injection rates, which widens the discrepancy between the backdoor model and the original teacher model. This larger gap effectively enhances the impact of attentional alignment, thereby strengthening the defense against backdoor attacks.

\subsubsection{Impact of Holding Rate}
\begin{figure*}
	\centering
	\subfigure[ASR on PROTEINS dataset.]{
		\begin{minipage}[t]{0.235\linewidth}
			\centering
			\includegraphics[width=1\linewidth]{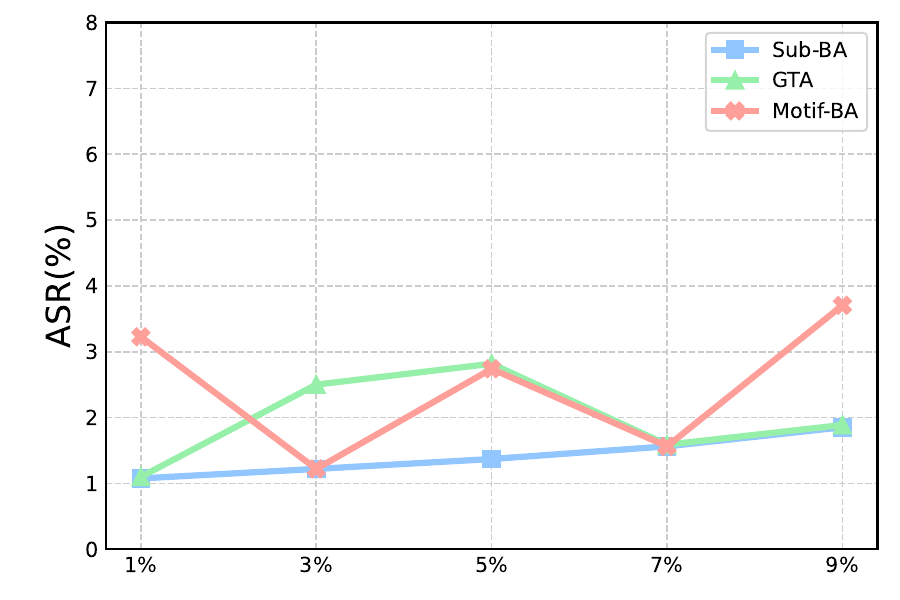}
	\end{minipage}}
	\subfigure[ASR on Fingerprint dataset.]{
		\begin{minipage}[t]{0.235\linewidth}
			\centering
			\includegraphics[width=1\linewidth]{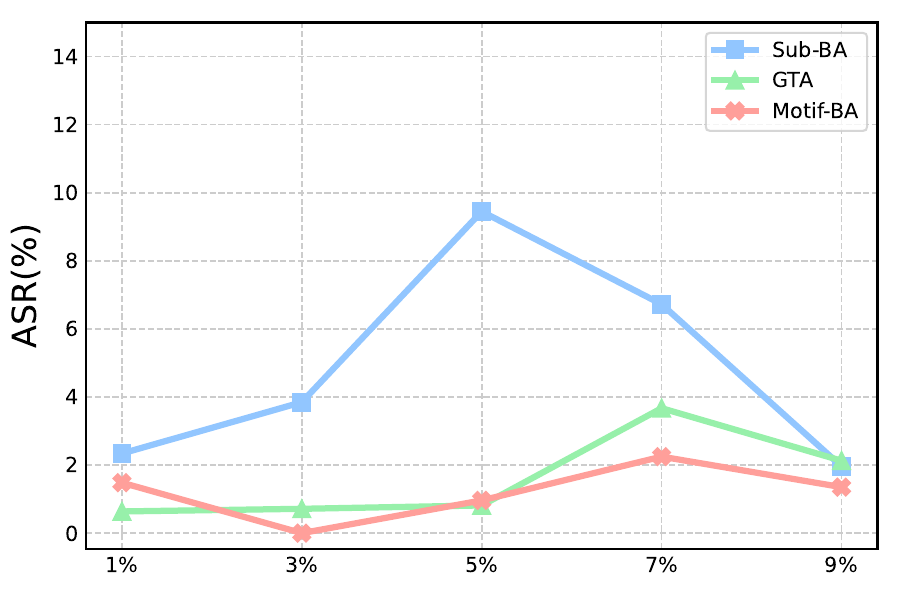}
	\end{minipage}}
	\subfigure[ASR on AIDS dataset.]{
		\begin{minipage}[t]{0.235\linewidth}
			\centering
			\includegraphics[width=1\linewidth]{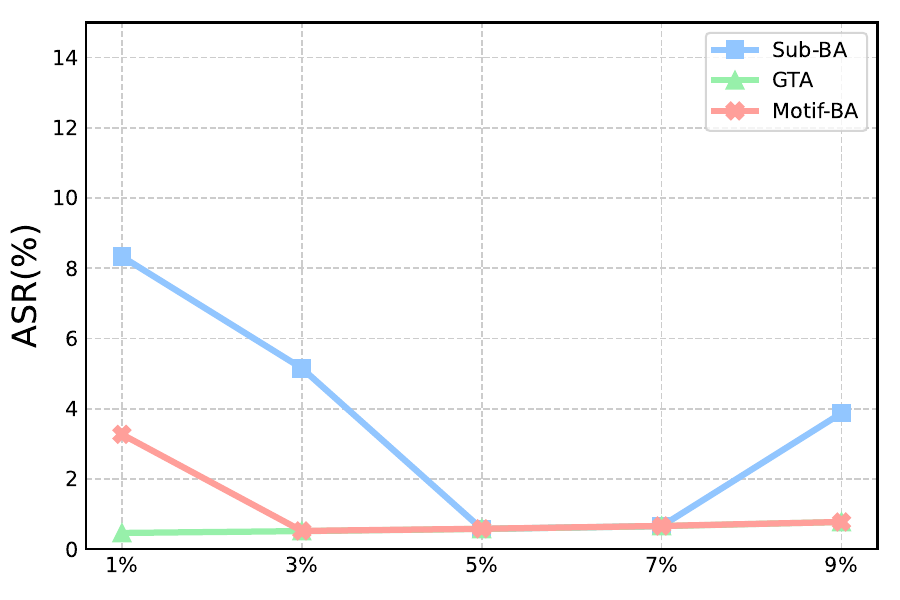}
	\end{minipage}}
	\subfigure[ASR on COLLAB dataset.]{
		\begin{minipage}[t]{0.235\linewidth}
			\centering
			\includegraphics[width=1\linewidth]{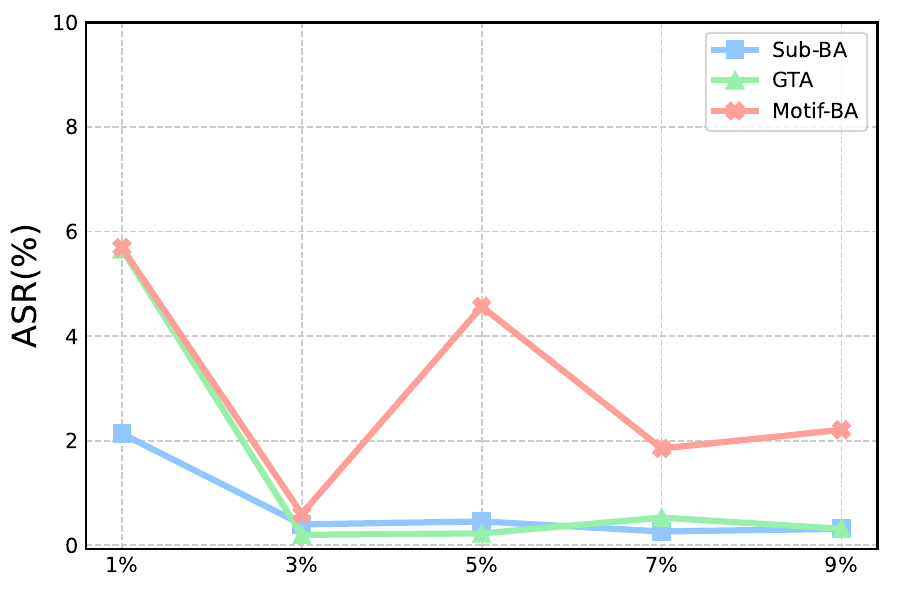}
	\end{minipage}}%
 \\
	\subfigure[ACC on PROTEINS dataset.]{
		\begin{minipage}[t]{0.235\linewidth}
			\centering
			\includegraphics[width=1\linewidth]{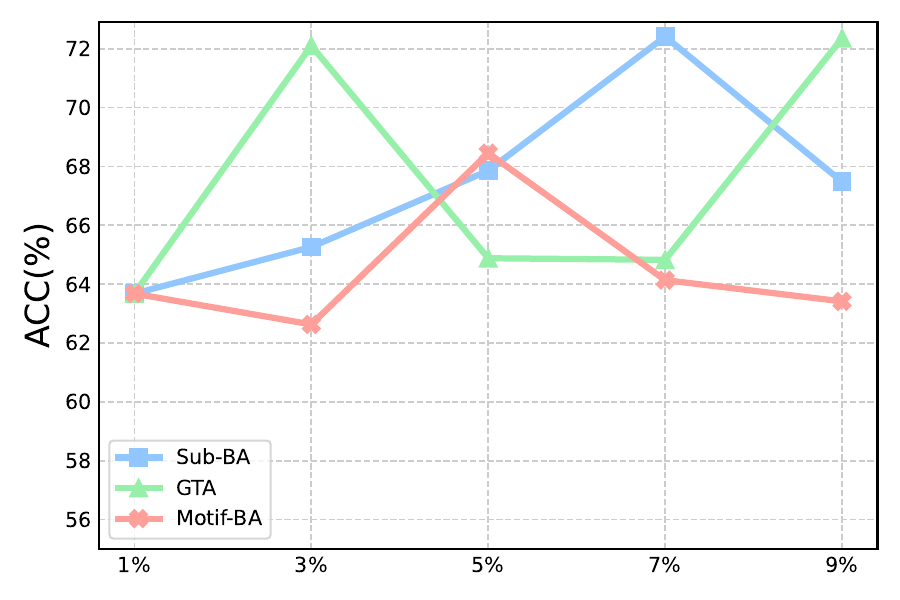}
	\end{minipage}}
	\subfigure[ACC on Fingerprint dataset.]{
		\begin{minipage}[t]{0.235\linewidth}
			\centering
			\includegraphics[width=1\linewidth]{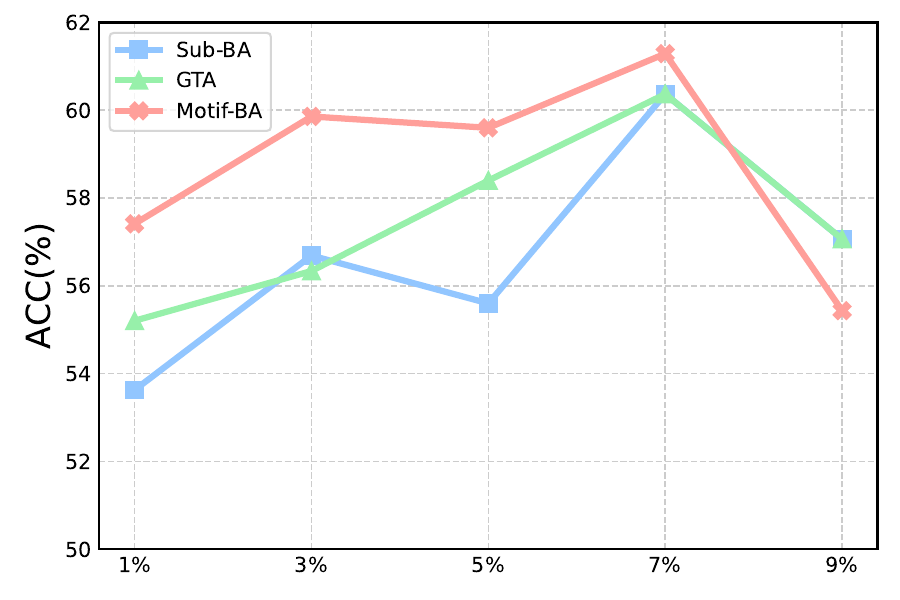}
	\end{minipage}}
	\subfigure[ACC on AIDS dataset.]{
		\begin{minipage}[t]{0.235\linewidth}
			\centering
			\includegraphics[width=1\linewidth]{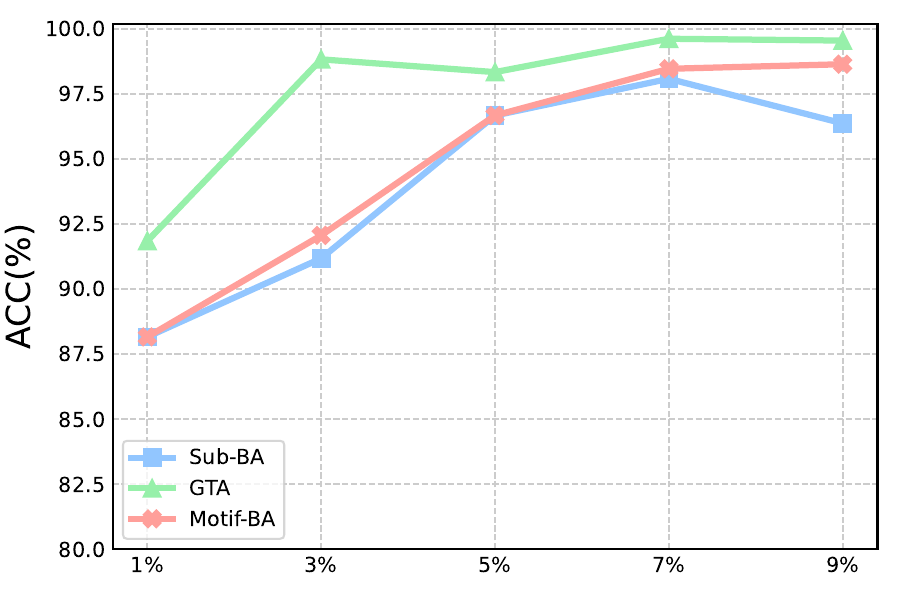}
	\end{minipage}}
	\subfigure[ACC on COLLAB dataset.]{
		\begin{minipage}[t]{0.235\linewidth}
			\centering
			\includegraphics[width=1\linewidth]{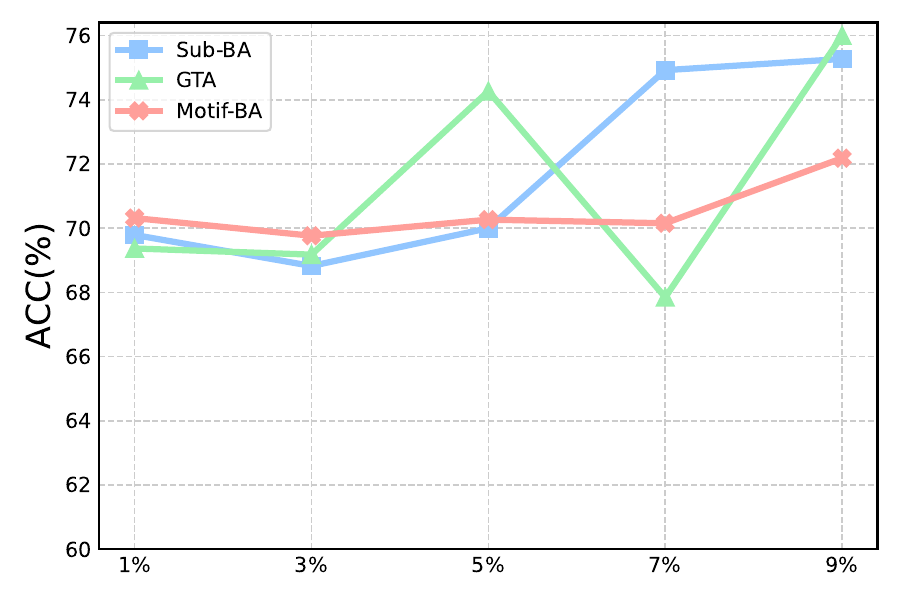}
	\end{minipage}}
	\caption{Impact of holding rate on four datasets.}
	\Description{Impact of holding rate on four datasets.}
	\label{Fclean_ratio}
\end{figure*}

As mentioned in Section \ref{Method}, a clean dataset is essential for the process of model fine-tuning and attention transfer. Within the experimental framework, defenders isolate a certain proportion of clean data from the test set, quantified as the clean data retention rate. This metric is calculated by dividing the number of clean data instances by the overall dataset size. Indeed, we restrict the clean data retention rate to a maximum of $10\%$ to more closely align with the challenges encountered in real-world applications, as shown in Fig \ref{Fclean_ratio}.

As anticipated, \sysname exhibits enhanced defensive capabilities as the availability of clean data samples increases. However, our approach continues to demonstrate acceptable performance even when utilizing only a small subset of such data. The ASR of backdoor attacks can be significantly reduced to below 10$\%$, as evidenced across four datasets with just 1$\%$ of clean data. This aligns with our motivation as our method recharacterizes features in the model's intermediate layers, effectively amplifying the discrepancies between the backdoored model and the fine-tuned model in the intermediate-layer representations. This enhancement makes the inactive backdoor neurons more prominent. Moreover, by incorporating changes between layers as additional knowledge, we thoroughly explore the potential of limited datasets.

\begin{table}[t]
 \caption{Performance of \sysname on alternative datasets.}
 \vspace{1mm}
  \centering
  \small
  \renewcommand\tabcolsep{7pt}
  \renewcommand\arraystretch{1}
\begin{tabular}{c|c|cc|cc}
\hline
\multirow{2}{*}{\textbf{Setting}}                                                              & \multirow{2}{*}{\textbf{Attack}} & \multicolumn{2}{c|}{\textbf{Origin}}                                                                                                                        & \multicolumn{2}{c}{\textbf{Auxiliary}} \\ \cline{3-6} 
                                                                                               &                                  & \textbf{ASR}                                                                & \textbf{ACC}                                                                  & \textbf{ASR}       & \textbf{ACC}      \\ \hline
\multirow{3}{*}{\textbf{\begin{tabular}[c]{@{}c@{}}NCI1\\ ->AIDS\end{tabular}}}   & \textbf{Sub-BA}                  & \multirow{6}{*}{\begin{tabular}[c]{@{}c@{}}0.62\\ 0.62\\ 0.63\end{tabular}} & \multirow{6}{*}{\begin{tabular}[c]{@{}c@{}}97.14\\ 99.29\\ 98.93\end{tabular}} & 18.04              & 95.59             \\
                                                                                               & \textbf{GTA}                     &                                                                             &                                                                               & 0.52               & 90.88             \\
                                                                                               & \textbf{Motif-BA}                &                                                                             &                                                                               & 0.52               & 82.06             \\ \cline{1-2} \cline{5-6} 
\multirow{3}{*}{\textbf{\begin{tabular}[c]{@{}c@{}}NCI109\\ ->AIDS\end{tabular}}} & \textbf{Sub-BA}                  &                                                                             &                                                                               & 8.24               & 88.82             \\
                                                                                               & \textbf{GTA}                     &                                                                             &                                                                               & 0.00                  & 81.76             \\
                                                                                               & \textbf{Motif-BA}                &                                                                             &                                                                               & 1.04               & 81.18             \\ \hline
\multicolumn{6}{l}{\small $\bullet$ Origin is the result of distillation with AIDS data.}\\
\end{tabular}
  \label{T-alternative}
\end{table}

In more extreme scenarios, defenders may find themselves unable to access any datasets directly related to the original data, relying solely on publicly available datasets that may have different distributions from the original data. We validated the effectiveness of our method in such cases. Specifically, we utilized two commonly used public datasets, NCI1 and NCI109, as auxiliary datasets to counteract backdoors implanted in the AIDS dataset by three attack methods. To maintain dimensionality consistency, we augmented the NCI1 dataset with a one-dimensional constant of 0, while no further modifications were required for NCI109 to align their dimensions. The experimental results presented in Table \ref{T-alternative} demonstrate that by leveraging these auxiliary datasets, our approach can reduce the ASR to approximately 10$\%$. Our analysis indicates that this reduction is due to the persistent differences between the intermediate-layer representations of the backdoored model and the fine-tuned model, regardless of whether the clean samples are drawn from the same distribution as the training model. By efficiently utilizing these feature discrepancies through attention mechanisms, we correct malicious neurons within the backdoored model. However, we also note that using alternative datasets may lead to a decrease in ACC compared to using clean test or validation datasets that match the distribution of the original data. This is akin to fine-tuning a pre-trained model where maintaining the original accuracy, especially when the feature distributions of the datasets differ, can be challenging. In other words, our approach necessitates additional knowledge of the features from the original dataset to sustain the model's performance on that dataset.

\subsubsection{Impact of Model Types} 
\begin{figure*}
	\centering
	\subfigure[Results on PROTEINS dataset.]{
		\begin{minipage}[t]{0.23\linewidth}
			\centering
			\includegraphics[width=1\linewidth]{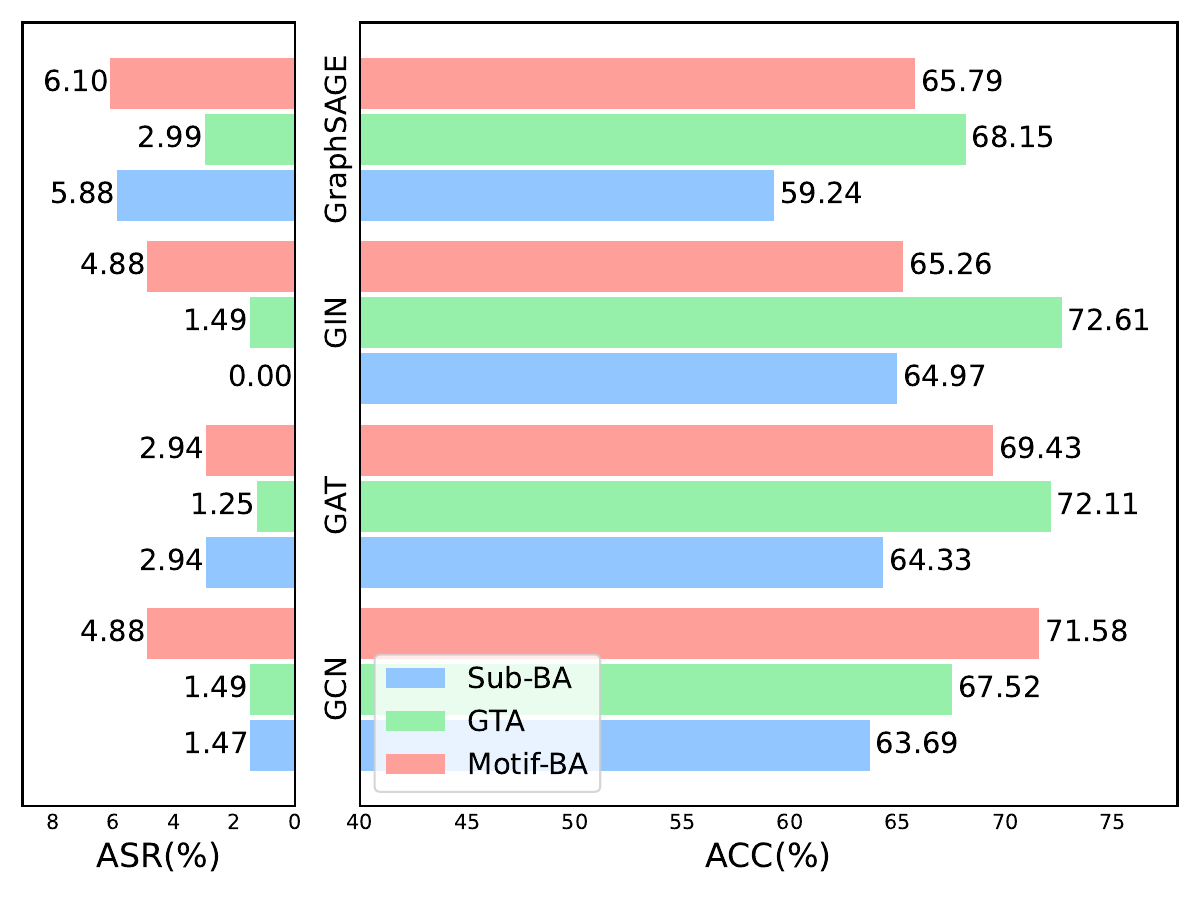}
	\end{minipage}}
	\subfigure[Results on Fingerprint dataset.]{
		\begin{minipage}[t]{0.23\linewidth}
			\centering
			\includegraphics[width=1\linewidth]{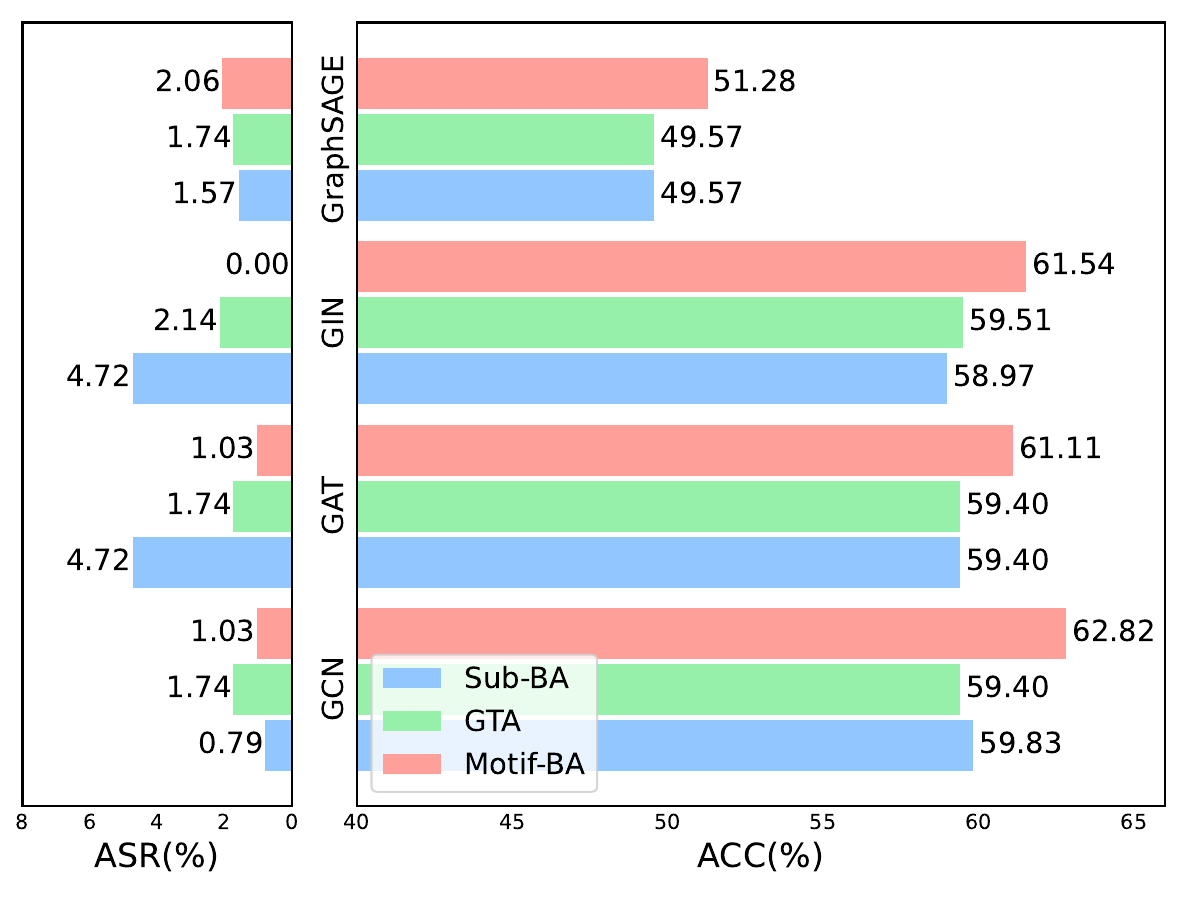}
	\end{minipage}}
	\subfigure[Results on AIDS dataset.]{
		\begin{minipage}[t]{0.23\linewidth}
			\centering
			\includegraphics[width=1\linewidth]{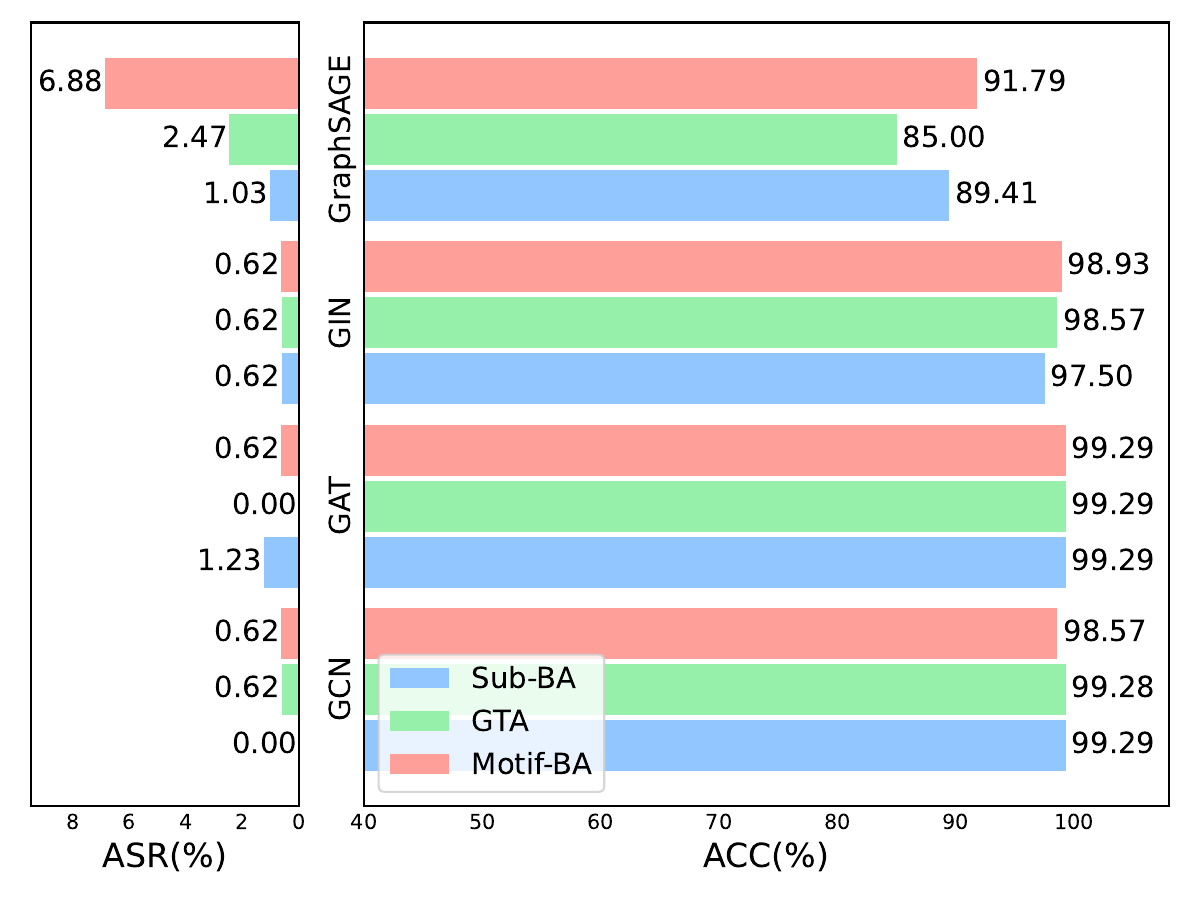}
	\end{minipage}}
	\subfigure[Results on COLLAB dataset.]{
		\begin{minipage}[t]{0.23\linewidth}
			\centering
			\includegraphics[width=1\linewidth]{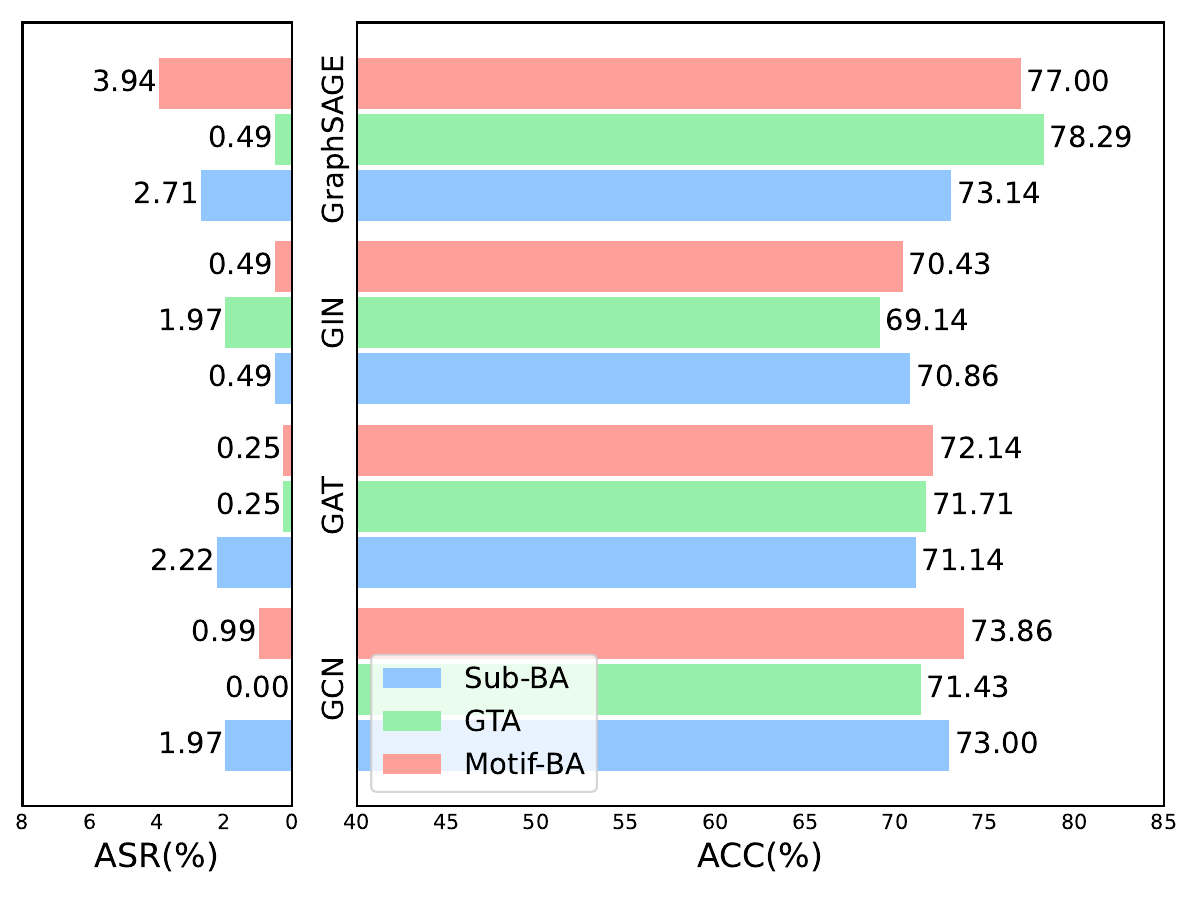}
	\end{minipage}}
	\caption{Impact of model types on four datasets.}
	\Description{Impact of model types on four datasets.}
	\label{F-model_type}
\end{figure*}

To assess the broad applicability of our method, we investigated the impact of different model structures on our approach. In addition to the default GIN model, we extended our evaluation to include the GAT, GCN, and GraphSAGE models, with experimental results illustrated in Fig. \ref{F-model_type}. Overall, our method consistently demonstrated satisfactory performance across a diverse range of attack vectors and the complexities inherent in various model structures. Remarkably, across different attack scenarios, ASR for all four models consistently remained at a reassuringly low level, well below the threshold of 7$\%$. The ACC of the primary task varied across different model structures and datasets, showing no distinct patterns. For instance, the GraphSAGE model performed less effectively than the other three models on PROTEINS, Fingerprint, and AIDS datasets, but outperformed them on COLLAB. This indicates the importance of selecting an appropriate model structure based on the nature of the data at hand. In conclusion, the outcomes of the experiments underscore the robustness of our strategy and emphasize its adaptability across a spectrum of model architectures. 

\subsubsection{Impact of Attention Functions}
In our scenario, the defenders have limited access to clean datasets, and the backdoor neurons are only weakly activated by clean samples, resulting in minimal representation differences between the backdoored model and the fine-tuned model on clean data. Therefore, the attention mechanism we employ to amplify these differences and enhance the quality of intermediate-layer representations is crucial. We explored the defensive capabilities of \sysname under various attention functions by adjusting the parameter $p$ in the formula outlined in Eq. \ref{EQ-Attention}. Increasing the value of $p$ amplifies the attention variances among model neurons. To reflect real-world conditions, we tested $p$ values within the range of 1-4.

As shown in Table \ref{T-attention_fuction}, all $p$ values prove highly effective in mitigating backdoors from the model. Notably, setting $p=2$ yields optimal defense performance across almost all datasets. Additionally, we find that setting $p=4$ results in satisfactory defense outcomes, attributed to the heightened attention differences between neurons. However, excessive amplification can lead to performance degradation on smaller datasets, such as Fingerprint, as larger $p$ values may cause the model to focus solely on certain nodes while neglecting other crucial information. This underscores the significance of selecting an appropriate attention parameter. Furthermore, we visually represented the attention maps at different $p$ values to provide a more intuitive understanding of our findings. Please refer to Appendix \ref{A-Visualizations} for these visualizations.
\begin{table*}
\centering
\caption{Impact of attention functions compared with baseline methods on four datasets.}
\label{T-attention_fuction}
\begin{tabular}{cc|cc|cccccccc}
\hline
\multirow{2}{*}{\textbf{Dataset}}     & \multirow{2}{*}{\textbf{Attack}} & \multicolumn{2}{c|}{Original} & \multicolumn{2}{c}{$p=1$}   & \multicolumn{2}{c}{$p=2$}                                & \multicolumn{2}{c}{$p=3$}                 & \multicolumn{2}{c}{$p=4$}                                \\ \cline{3-12} 
                                      &                                  & \textbf{ASR}  & \textbf{ACC}  & \textbf{ASR} & \textbf{ACC} & \textbf{ASR}               & \textbf{ACC}                & \textbf{ASR}               & \textbf{ACC} & \textbf{ASR}               & \textbf{ACC}                \\ \hline
\multirow{3}{*}{\textbf{PROTEINS}}    & \textbf{Sub-BA}                  & 96.34         & 68.42         & 4.41         & 65.61        & \cellcolor{lightgreen}1.47 & \cellcolor{lightgreen}71.34 & 2.94                       & 70.06        & \cellcolor{lightgreen}1.47 & 67.52                       \\
                                      & \textbf{GTA}                     & 95.52         & 73.25         & 4.48         & 63.06        & \cellcolor{lightgreen}1.49 & \cellcolor{lightgreen}72.61 & \cellcolor{lightgreen}1.49 & 68.79        & 2.99                       & 68.79                       \\
                                      & \textbf{Motif-BA}                & 42.68         & 73.16         & 7.35         & 66.24        & \cellcolor{lightgreen}1.47 & 69.43                       & 5.88                       & 68.79        & 2.94                       & \cellcolor{lightgreen}70.70 \\ \hline
\multirow{3}{*}{\textbf{COLLAB}}      & \textbf{Sub-BA}                  & 77.49         & 79.52         & 0.49         & 74.57        & 0.74                       & \cellcolor{lightgreen}75.14 & 0.49                       & 70.14        & \cellcolor{lightgreen}0.25 & 73.71                       \\
                                      & \textbf{GTA}                     & 92.36         & 80.43         & 0.99         & 72.00        & \cellcolor{lightgreen}0.49 & \cellcolor{lightgreen}77.57 & 1.72                       & 70.71        & \cellcolor{lightgreen}0.49 & 68.43                       \\
                                      & \textbf{Motif-BA}                & 81.46         & 79.71         & 1.48         & 72.29        & \cellcolor{lightgreen}0.49 & \cellcolor{lightgreen}74.00 & 1.48                       & 71.71        & 1.48                       & \cellcolor{lightgreen}74.00 \\ \hline
\multirow{3}{*}{\textbf{AIDS}}        & \textbf{Sub-BA}                  & 80.86         & 97.86         & 3.09         & 89.64        & \cellcolor{lightgreen}0.62 & \cellcolor{lightgreen}97.14 & 1.23                       & 98.21        & \cellcolor{lightgreen}0.62 & \cellcolor{lightgreen}97.14 \\
                                      & \textbf{GTA}                     & 100.00        & 99.25         & 1.85         & 98.57        & \cellcolor{lightgreen}0.62 & \cellcolor{lightgreen}99.29 & 1.23                       & 98.93        & \cellcolor{lightgreen}0.62 & 98.57                       \\
                                      & \textbf{Motif-BA}                & 77.50         & 98.21         & 1.25         & 98.21        & \cellcolor{lightgreen}0.63 & \cellcolor{lightgreen}98.93 & 1.25                       & 97.50        & \cellcolor{lightgreen}0.63 & 97.86                       \\ \hline
\multirow{3}{*}{\textbf{Fingerprint}} & \textbf{Sub-BA}                  & 91.67         & 61.62         & 4.72         & 58.97        & \cellcolor{lightgreen}0.79 & \cellcolor{lightgreen}59.83 & 4.72                       & 58.12        & 3.94                       & \cellcolor{lightgreen}59.83 \\
                                      & \textbf{GTA}                     & 99.13         & 60.26         & 6.09         & 58.97        & \cellcolor{lightgreen}3.48 & \cellcolor{lightgreen}58.97 & 4.35                       & 52.14        & 8.70                       & \cellcolor{lightgreen}58.97 \\
                                      & \textbf{Motif-BA}                & 100.00        & 61.11         & 2.06         & 59.40        & \cellcolor{lightgreen}1.03 & \cellcolor{lightgreen}59.40 & 2.06                       & 55.56        & 8.25                       & 58.55                       \\ \hline
\end{tabular}
\end{table*}

\subsubsection{Impact of Relationship Pairs}

As detailed in Section \ref{W-relation}, the loss associated with attention relation congruence is computed based on the interrelationships between the layers of the model, denoted as $R^{ij}$. In Table \ref{T-relationship_pair}, we delve into the influence of selecting different relationship pairs on the defensive performance. The notation $pairs=\{full\}$ indicates that all relationship pairs between the teacher and student models are considered, which amounts to $\frac{l(l-1)}{2}$ pairs, where $l$ is the number of layers in the model. Conversely, $pairs=\{<i,i+k>\}$ signifies that only the relationships between layer $i$ and layer $i+k$ of the teacher and student models are taken into account.

The results indicate that the best performance is achieved when $pairs=\{full\}$ is employed. This is because considering all the relationships between the layers of the model allows for a more accurate capture of the distributional characteristics of clean data. When $pairs=\{<i,i+2>\}$ is used, the model purification effect is still good, but $pairs=\{full\}$ improves ACC on AIDS compared to the original model. To ensure the accuracy of the main task, we ultimately selected $pairs=\{full\}$ for our approach.

\begin{table*}
\centering
\caption{Impact of relationship pairs compared with baseline methods on four datasets.}
\label{T-relationship_pair}
\begin{tabular}{cc|cc|cccccccc}
\hline
\multirow{2}{*}{\textbf{Dataset}}     & \multirow{2}{*}{\textbf{Attack}} & \multicolumn{2}{c|}{Original} & \multicolumn{2}{c}{$pairs=\{full\}$}                     & \multicolumn{2}{c}{$pairs=\{<i,i+1>\}$}   & \multicolumn{2}{c}{$pairs=\{<i,i+2>\}$}    & \multicolumn{2}{c}{$pairs=\{<i,i+3>\}$}    \\ \cline{3-12} 
                                      &                                  & \textbf{ASR}  & \textbf{ACC}  & \textbf{ASR}               & \textbf{ACC}                & \textbf{ASR}               & \textbf{ACC} & \textbf{ASR} & \textbf{ACC}                & \textbf{ASR} & \textbf{ACC}                \\ \hline
\multirow{3}{*}{\textbf{PROTEINS}}    & \textbf{Sub-BA}                  & 96.34         & 68.42         & \cellcolor{lightgreen}1.47 & \cellcolor{lightgreen}71.34 & 2.94                       & 66.24        & 4.41         & 68.15                       & 2.94         & 69.43                       \\
                                      & \textbf{GTA}                     & 95.52         & 73.25         & \cellcolor{lightgreen}1.49 & \cellcolor{lightgreen}72.61 & 2.99                       & 71.97        & 4.48         & 65.61                       & 4.48         & 68.79                       \\
                                      & \textbf{Motif-BA}                & 42.68         & 73.16         & \cellcolor{lightgreen}1.47 & \cellcolor{lightgreen}69.43 & 2.94                       & 66.24        & 2.94         & 64.97                       & 4.41         & 64.97                       \\ \hline
\multirow{3}{*}{\textbf{COLLAB}}      & \textbf{Sub-BA}                  & 77.49         & 79.52         & \cellcolor{lightgreen}0.74 & \cellcolor{lightgreen}75.14 & 3.60                       & 72.82        & 4.60         & 73.29                       & 5.00         & 70.94                       \\
                                      & \textbf{GTA}                     & 92.36         & 80.43         & \cellcolor{lightgreen}0.49 & \cellcolor{lightgreen}77.57 & 3.69                       & 71.29        & 5.80         & 69.18                       & 0.99         & 67.57                       \\
                                      & \textbf{Motif-BA}                & 81.46         & 79.71         & \cellcolor{lightgreen}0.49 & \cellcolor{lightgreen}74.00 & 0.99                       & 71.29        & 1.60         & 70.47                       & 0.74         & 70.14                       \\ \hline
\multirow{3}{*}{\textbf{AIDS}}        & \textbf{Sub-BA}                  & 80.86         & 97.86         & \cellcolor{lightgreen}0.62 & 97.14                       & 1.23                       & 97.14        & 1.23         & \cellcolor{lightgreen}98.21 & 1.85         & 97.50                       \\
                                      & \textbf{GTA}                     & 100.00        & 99.25         & \cellcolor{lightgreen}0.62 & \cellcolor{lightgreen}99.29 & 1.23                       & 97.50        & 1.23         & 92.14                       & 4.94         & 96.43                       \\
                                      & \textbf{Motif-BA}                & 77.50         & 98.21         & \cellcolor{lightgreen}0.63 & \cellcolor{lightgreen}98.93 & 1.25                       & 96.43        & 1.88         & 97.50                       & 1.25         & 96.79                       \\ \hline
\multirow{3}{*}{\textbf{Fingerprint}} & \textbf{Sub-BA}                  & 91.67         & 61.62         & \cellcolor{lightgreen}0.79 & 59.83                       & 4.72                       & 58.55        & 1.57         & 53.85                       & 4.72         & \cellcolor{lightgreen}60.68 \\
                                      & \textbf{GTA}                     & 99.13         & 60.26         & 3.48                       & 58.97                       & \cellcolor{lightgreen}2.61 & 54.27        & 5.22         & 59.40                       & 6.09         & \cellcolor{lightgreen}60.68 \\
                                      & \textbf{Motif-BA}                & 100.00        & 61.11         & \cellcolor{lightgreen}1.03 & \cellcolor{lightgreen}59.40 & 6.19                       & 58.55        & 3.09         & 58.97                       & 2.06         & 58.55                       \\ \hline
\end{tabular}
\end{table*}

\subsection{Attention Maps}

\begin{figure*}
	\centering
	\includegraphics[width=1\linewidth]{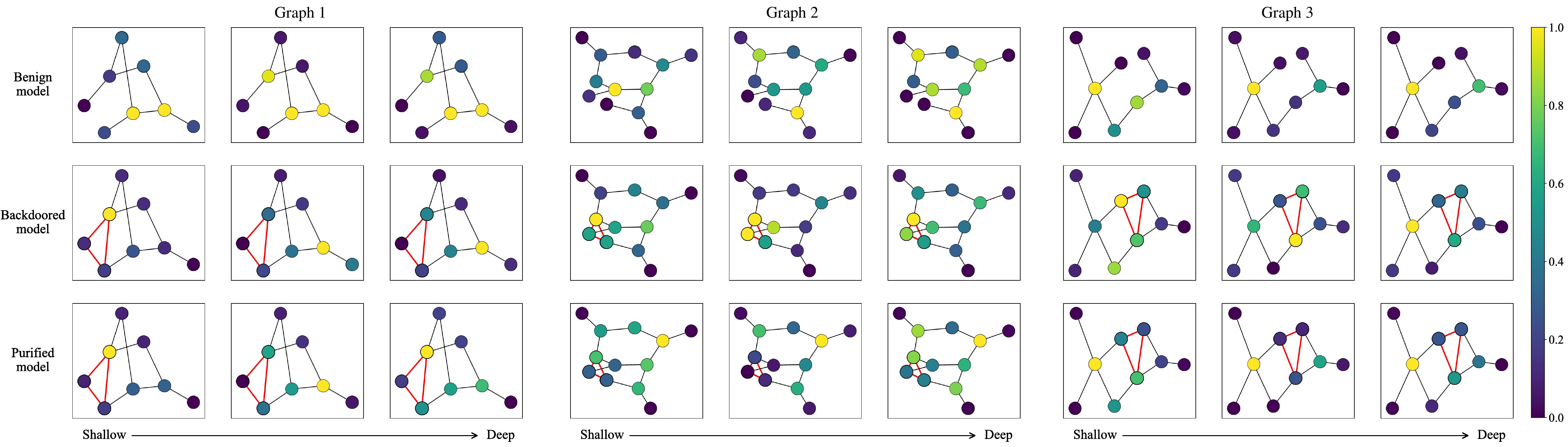}
	\caption{Attention maps of different layers. The color of the nodes represents the model's attention to them, while the red edges denote the presence of a backdoor trigger.}
	\Description{Attention maps of different layers.}
	\label{F-attention}
\end{figure*}

In the preceding sections \ref{W-attention}, we have examined the prevalent techniques for knowledge distillation within the graph domain. When it comes to distilling middle-layer knowledge, these approaches typically harness the intermediate-layer information directly, neglecting the utility of an Attention Map. However, by leveraging the Attention Map, we gain insight into the most salient regions within the network's topology. To illustrate the role of the Attention Map in assisting \sysname in obliterating backdoor triggers, we present a visual comparison of the Attention Maps before and after the removal process in Figure \ref{F-attention}. 

We employ a color gradient within the graph to represent the varying degrees of the model's attention allocated to each node. In the case of the backdoored model, the attention graphs uniformly exhibit a strong bias towards the backdoor trigger region, indicating that the triggers can effortlessly manipulate the network's behavior. Consequently, the objective of the backdoor erasure technique is to mitigate the excessive focus on this area, thereby restoring the network's integrity.

\subsection{Ablation Studies}

\subsubsection{Loss Terms}
\begin{figure}
	\centering
	\subfigure[Heatmap of ASR.]{
		\begin{minipage}[t]{0.485\linewidth}
			\centering
			\includegraphics[width=1\linewidth]{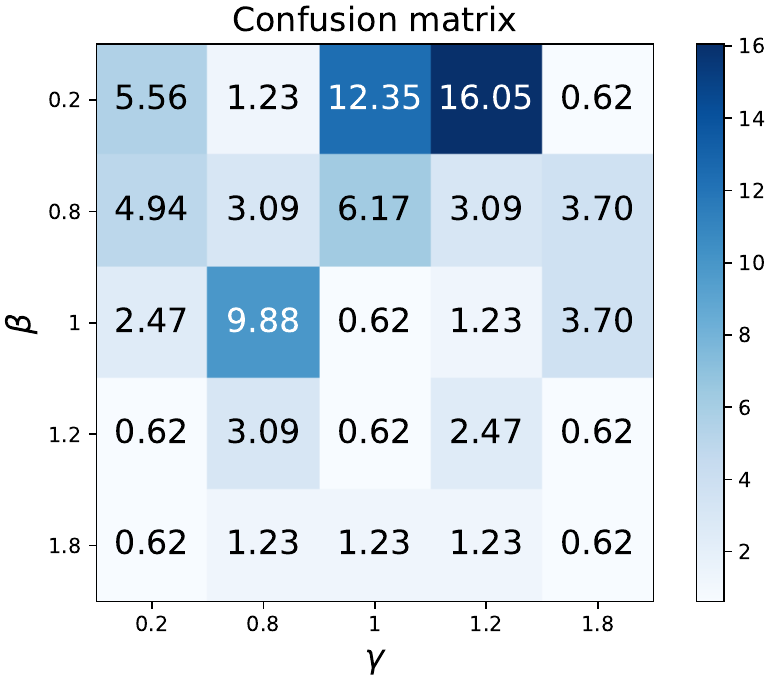}
	\end{minipage}}
        \subfigure[Heatmap of ACC.]{
		\begin{minipage}[t]{0.485\linewidth}
			\centering
			\includegraphics[width=1\linewidth]{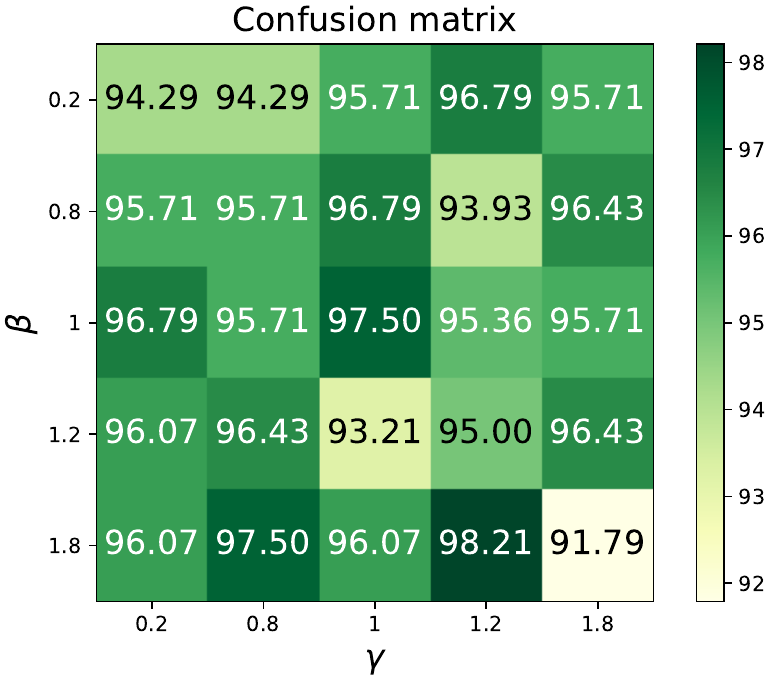}
	\end{minipage}}
	\caption{Impact of loss term $\beta$ and $\gamma$.}
	\Description{Impact of loss term $\beta$ and $\gamma$.}
	\label{F-loss_term}
\end{figure}

In this section, we study the impact of different hyperparameters on \sysname. We focus on the hyperparameters $\beta$ and $\gamma$, which are pivotal in maintaining an equilibrium between $\mathcal{L}_{AD}$ and $\mathcal{L}_{RC}$. We consider that both graph attention transfer and attention relation congruence play a steady role. Consequently, our baseline configuration sets $\beta$ and $\gamma$ to unity. Within our framework, we examine the repercussions of assigning different constant values to these hyperparameters. Specifically, we vary the values of $\beta$ and $\gamma$ as $\{0.2,0.8,1,1.2,1.8\}$. The attack method used is Sub-BA. The outcomes of this exploration are depicted in Fig \ref{F-loss_term}.

In summary, the efficacy of mitigating the backdoor effect is enhanced as the values of $\beta$ and $\gamma$ are incremented. However, excessively high values can inversely impair performance, potentially diminishing the accuracy of the primary task. Optimal distillation outcomes are achieved through a prudent selection of moderate $\beta$ and $\gamma$ values.

\subsubsection{Component Contributions}

\begin{table}[t]
 \caption{\small Performance of \sysname with different components.}
  \vspace{1mm}
  \centering
  \renewcommand\tabcolsep{10pt}
  \renewcommand\arraystretch{1}
\begin{tabular}{cc|cc}
\hline
$\mathcal{L}_{AD}$           & $\mathcal{L}_{RC}$           & \textbf{ACC} & \textbf{ASR} \\ \hline
\textcolor{red}{\ding{55}}   & \textcolor{red}{\ding{55}}   & 91.18        & 38.66        \\ \hline
\textcolor{red}{\ding{55}}   & \textcolor{green}{\ding{51}} & 90.00         & 10.31        \\ \hline
\textcolor{green}{\ding{51}} & \textcolor{red}{\ding{55}}   & 89.71        & 3.61         \\ \hline
\textcolor{green}{\ding{51}} & \textcolor{green}{\ding{51}} & 97.50         & 0.62         \\ \hline
\end{tabular}
  \label{T-components}
\end{table}

 Our distillation losses are divided into two components: graph attention transfer loss $\mathcal{L}_{AD}$ and attention relation congruence loss $\mathcal{L}_{RC}$. We delve into the roles of each component to advance our understanding of the working mechanisms behind the distillation process. Table \ref{T-components} presents the contributions of different components.

Obviously, reliance solely on $\mathcal{L}_{AD}$ enables the forgetting of backdoor features, yet it comes at the cost of compromising ACC. Conversely, utilizing $\mathcal{L}_{RC}$ independently manages to sustain ACC but falls short in effectively eliminating the backdoor. The concurrent application of both $\mathcal{L}_{AD}$ and $\mathcal{L}_{RC}$ successfully unlearns the backdoor features while maintaining ACC. The finely tuned balance of these two parameters further enhances overall performance, underscoring the superior efficiency of our proposed method.

\begin{table}
    \centering
\caption{Performance of \sysname against adative backdoor attacks.}
\label{T-adative_backdoor}
\begin{tabular}{c|cccccc}
\hline
            & \multicolumn{2}{c}{\textbf{GTA}} & \multicolumn{2}{c}{\textbf{Adative GTA}} & \multicolumn{2}{c}{\textbf{\sysname}} \\ \cline{2-7}
            & \textbf{ASR}        & \textbf{ACC}        & \textbf{ASR}            & \textbf{ACC}            & \textbf{ASR}           & \textbf{ACC}          \\ \hline
\textbf{PROTEINS}    & 95.52& 73.25& 91.25& 75.79& 5.00& 65.79
\\
\textbf{COLLAB}      & 92.36& 77.57& 95.80& 78.24& 0.00& 66.00
\\
\textbf{AIDS}        & 100.00& 99.25& 97.94& 99.12& 0.52& 98.24
\\
\textbf{Fingerprint} & 99.13& 60.26& 100.00& 59.86& 7.86& 58.80
\\ \hline
\end{tabular}
\end{table}
 
\subsection{Adaptive Attacks}
Beyond the three backdoor assaults previously outlined, we have developed a tailored adaptive attack strategy intended to circumvent \sysname. This serves to assess \sysname's capability in effectively countering such an evolved threat. Given that \sysname utilizes an attentional distillation process to expunge backdoors, an adversary could theoretically harness this very mechanism—the attentional distillation loss $\mathcal{L}_{AD}$—to refine a backdoor model. By optimizing the triggers in this manner, the attacker aims to fulfill the objectives of the adaptive attack.

We have crafted a formidable adaptive attack leveraging the cutting-edge graph-based attack technique, known as GTA \cite{GTA}. In particular, adhering to the core architecture of GTA, which comprises a topology generator and a feature generator, we incorporate an attention distillation loss, denoted as $\mathcal{L}_{AD}$, into the training regimen of the attack triggers. This integration is designed to synchronize the attention mechanisms of the backdoored model with those of a pristine model, thereby enhancing the attack's efficacy.

We assessed the impact of the adaptive attack on four distinct datasets, as illustrated in Table \ref{T-adative_backdoor}. The ASR for the adaptive attack on the PROTEINS and AIDS datasets experiences a slight decrease but still remains substantial. By deploying \sysname against the adaptive GTA, we observe that our method effectively diminishes the ASR to below $10\%$ while incurring minimal decrements in the accuracy of the primary task.

%% file: data/6-related_work.tex
\section{Related Work}

\subsection{Backdoor Attack and Defense on DNNs}

Backdoor attacks have been largely investigated in DNNs. Gu et al. pioneered exploratory research on backdoor attacks (Badnets) \cite{Gu2019BadNetsEvaluating}, utilizing pixel patches as triggers for activating covert operations. Subsequent studies have focused on enhancing the stealthiness and effectiveness of such attacks, which can be categorized into Invisible attacks \cite{chen2017targeted,Liu2019ABSScanning,Liu2020ReflectionBackdoor,Zhong2020BackdoorEmbedding,Li2021InvisibleBackdoor,Nguyen2021WaNetImperceptible} and Natural attacks \cite{li2020rethinking,Doan2021BackdoorAttack,Ren2021SimtrojanStealthy,Zhang2022PoisonInk,Zhao2022DEFEATDeep}. \ding{182}\textit{Invisible attacks} introduce imperceptible changes as the trigger mechanism for backdoor attacks by minimizing pixel-level differences between the original and manipulated images. \ding{183}\textit{Natural attacks} adjust the stylistic elements of images as the trigger, aiming to ensure that the images maintain their natural appearance and thus reduce the likelihood of suspicion \cite{Liu2020ReflectionBackdoor,Wan2020FeatureConsistency,Cheng2021DeepFeature,Xue2023CompressionresistantBackdoor}.

Extensive defenses have been proposed to address the impact of backdoors and cleansing compromised models \cite{Wang2019NeuralCleanse,Gong2023RedeemMyself,Weber2023RABProvable,Kumari2023BayBFedBayesian,zhang2024badcleaner,zhang2024flpurifier}. For instance, NeuralCleanse (NC) \cite{Wang2019NeuralCleanse} utilizes reverse engineering to evaluate the trigger effects and identifies backdoors based on defined thresholds. SAGE \cite{Gong2023RedeemMyself} introduces a top-down attention distillation mechanism, which leverages benign shallow layers to guide the mitigation of harmful deep layers, thereby enhancing the defense capability through normalization and adaptive learning rate adjustments. However, all these defenses cannot be directly applied to the graph domain since they cannot capture the topological structure information inherent in graph data.

\subsection{Backdoor Attack and Defense on GNNs}
Current research on graph backdoor attacks can be classified by their trigger generation strategies as follows:

\ding{182} \textbf{Tampering with node features}. 
Xu et al. \cite{EXPBA} utilized GNN interpretability methods to identify optimal node positions for implanting triggers and assessed backdoor attacks on node classification tasks, though this node-traversal approach is time-consuming. To mitigate this, Chen et al. \cite{chen2022general} employed edge and feature explanation methods to target arbitrary nodes without affecting non-targeted ones. Similarly, Xu et al. \cite{EXP2} further analyzed the impact of node importance on backdoor attacks, while Dai et al. \cite{transferable} introduced transferred semantic backdoors, assigning triggers to specific node classes to evade detection.

\ding{183} \textbf{Disrupting graph topology}. 
Xi et al. \cite{GTA} developed GTA, using dynamically adjustable subgraphs as triggers to create scalable attacks independent of downstream tasks. Zheng et al. \cite{motifBA} explored triggers from a motif perspective, showing that the frequency of trigger subgraphs impacts attack effectiveness. Chen et al. \cite{neighboring} proposed neighboring backdoors with triggers as individual nodes linked to target nodes, ensuring normal model function if nodes are disconnected. However, these methods often produce outputs with incorrect labels, making them detectable. To overcome this, Xu et al. \cite{clean} introduced clean-label backdoor attacks, where poisoned inputs share consistent labels with clean ones, improving attack success rates during testing.

\ding{184} \textbf{Node information alteration}. 
Graph Contrastive Backdoor Attacks (GCBA) \cite{contras} introduced the first backdoor attack for graph contrastive learning (GCL), employing various strategies across different stages of the GCL pipeline, including data poisoning and post-pretraining encoder injection. Unnoticeable Graph Backdoor Attack (UGBA) \cite{dai2023unnoticeable} aimed to achieve covert attacks with limited budgets by selectively targeting nodes for backdoor insertion and using adaptive triggers. The limited exploration of defenses against backdoor attacks in GNNs has prompted the development of the first effective graph backdoor defense mechanism in our study.

Several promising defense ideas have been discussed in the attack proposals, such as model inspection \cite{Wang2019NeuralCleanse,wu2019adversarial}, Prune \cite{dai2023unnoticeable}, and RS \cite{Wang2021CertifiedRobustnessGraph}. Model inspection seeks to reverse-engineer the unified trigger in backdoored models to disrupt the link between trigger-laden inputs and target classes, but its effectiveness in GNNs is limited by the challenge of creating a universal trigger. Prune and RS, which dilutes backdoor features through random sampling of graph structure and node attributes, risks discarding vital features, potentially harming model performance. Other defense countermeasures have been developed to alleviate the backdoor attacks in GNNs, which can be categorized as detection \cite{Detect1,Detect2,Detect3} and mitigation \cite{Node1,Node2,Node3,Node4}. Backdoor detection methods attempt to distinguish the differences between benign and backdoor samples to determine whether a model is backdoored. Nevertheless, these approaches are not applicable to scenarios with diverse backdoor triggers nor capable of effectively eliminating backdoor triggers. Furthermore, existing mitigation strategies are only effective in the context of node classification tasks and commonly rely on the impractical assumption of having access to a large dataset of clean data. Instead, our method can simultaneously defend against both graph and node classification scenarios with only a small portion of clean data.

%% file: data/7-conclusion.tex
\section{Conclusion \& Discussion}

In this paper, we present a GNN backdoor mitigation approach, denoted as \sysname. \sysname effectively mitigates the negative effects through knowledge distillation. Specifically, \sysname comprises two primary modules: graph attention transfer and attention relation congruence. The first step involves fine-tuning the backdoored model with a small subset of clean data, which serves as the teacher model. Subsequently, on the same clean data, we employ our proposed graph attention representation method to align the attention of the backdoored model with that of the teacher model. To extract more effective information from limited data, we consider the relationships between layers during alignment, ensuring that the backdoored model is aligned with the teacher model. Extensive experiments conducted on four datasets against three SOTA attacks validate the superiority of \sysname compared to three baseline methods.

\textbf{Limitations \& Future work.} \sysname has several limitations that need to be addressed in future work. Firstly, \sysname is designed for neural networks with a multi-layer structure (3 layers), and its performance may be affected when faced with networks with fewer layers. Secondly, \sysname requires a portion of the original dataset to perform fine-tuning and distillation. Although our study has confirmed that even under extreme circumstances, using auxiliary public datasets with different distributions from the original dataset can still achieve effective backdoor defense, a related performance decline is observed. Lastly, our focus is primarily on the defense against backdoors in subgraph classification. How to transplant existing defense strategies to node prediction scenarios remains an open problem. In our upcoming work, our goal is to conduct in-depth research in scenarios where there is no dataset interaction, aiming to develop a general backdoor defense framework tailored to different graph tasks while achieving graph backdoor mitigation.

%% file: data/99-appendix.tex
\section{Details on Datasets}
\label{A-Dataset}
\begin{itemize}
    \item \textbf{PROTEINS} \cite{protein}: It is a widely used benchmark dataset in the field of protein structure prediction. It consists of 1,203 protein structures with lengths ranging from 20 to 304 residues. Each protein in the dataset is represented by a set of 9 features, including secondary structure, solvent accessibility, and residue depth.
    \item \textbf{Fingerprint} \cite{finger}: Fingerprint is a collection of fingerprints formatted as graph structures from the NIST-4 database, which consists of 4,000 grayscale images of fingerprints with class labels according to the five most common classes of the Galton-Henry classification scheme.
    \item \textbf{AIDS} \cite{aids}: It is a collection of anonymized medical records from patients diagnosed with acquired immune deficiency syndrome (AIDS). This dataset contains information on various demographic, clinical, and laboratory variables, such as age, sex, CD4 cell count, viral load, and the presence of opportunistic infections.
    \item \textbf{COLLAB} \cite{collab}: It is a scientific collaboration dataset. Each graph in the dataset represents a researcher's ego network, where the researcher and their collaborators are nodes, and collaboration between researchers is indicated by edges. The ego network of a researcher is classified into three labels: High Energy Physics, Condensed Matter Physics, and Astro Physics, representing their respective fields of study.
\end{itemize}

\section{Details on Attack Methods}
\label{A-Attack}
\begin{itemize}
    \item \textbf{Subgraph-based Backdoor (Sub-BA)} \cite{badsub}: Sub-BA involves injecting a subgraph trigger into a graph. In our experiments, we generate the subgraph trigger using the Erdős-Rényi (ER) model \cite{ER}. The poisoned nodes are randomly selected from the node set of the graph.
    \item \textbf{GTA} \cite{GTA}: To execute the GTA attack, we train a topology generator and a feature generator. These generators enable us to generate a range of candidate backdoor triggers. Poisoned nodes is randomly selected in GTA. We subsequently solve a bi-level optimization problem iteratively to poison the victim graphs with the backdoor triggers. For all experiments, we set the number of epochs as 20 for the bi-level optimization process.
    \item \textbf{Motif-Backdoor (Motif-BA)} \cite{motifBA}: Motif-BA first obtains the distribution of the motif through the motif extraction tool and analyzes it to select the suitable motif as the trigger. Then, the trigger injection position is determined by using network importance metrics, shadow models, and target node-dropping strategies. Finally, it injects the trigger into benign graphs and learns a backdoored model on them.
\end{itemize}

\section{Impact of Distillation on Various Categorization}
\label{A-various_cate}
\begin{figure}
	\centering
	\subfigure[Clean model.]{
		\begin{minipage}[t]{0.32\linewidth}
			\centering
			\includegraphics[width=1\linewidth]{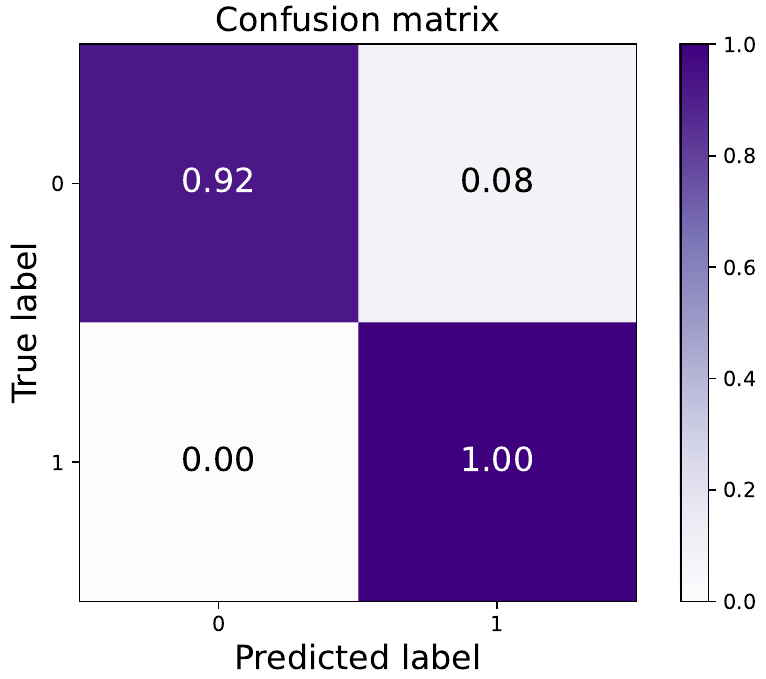}
	\end{minipage}}
    \subfigure[Backdoored model.]{
		\begin{minipage}[t]{0.32\linewidth}
			\centering
			\includegraphics[width=1\linewidth]{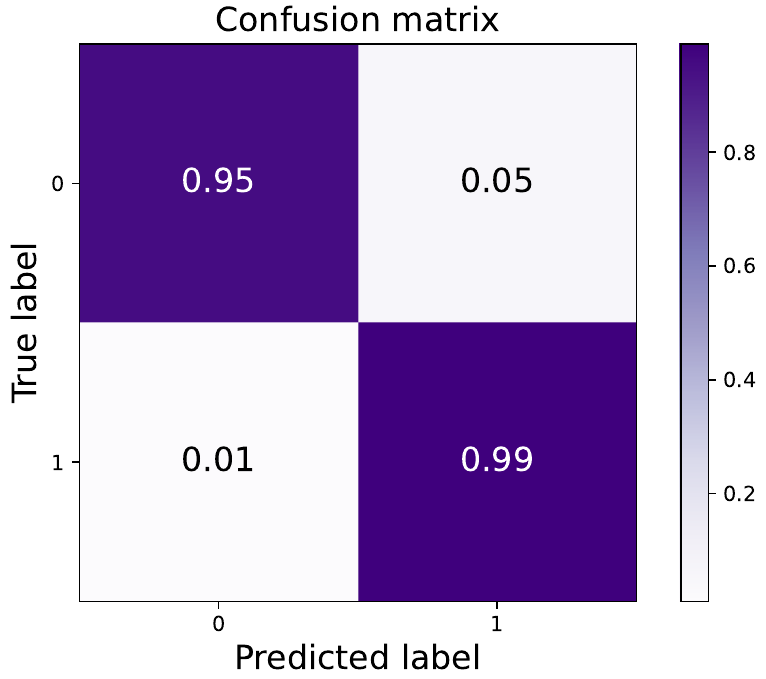}
	\end{minipage}}%
	\subfigure[Purified model.]{
		\begin{minipage}[t]{0.32\linewidth}
			\centering
			\includegraphics[width=1\linewidth]{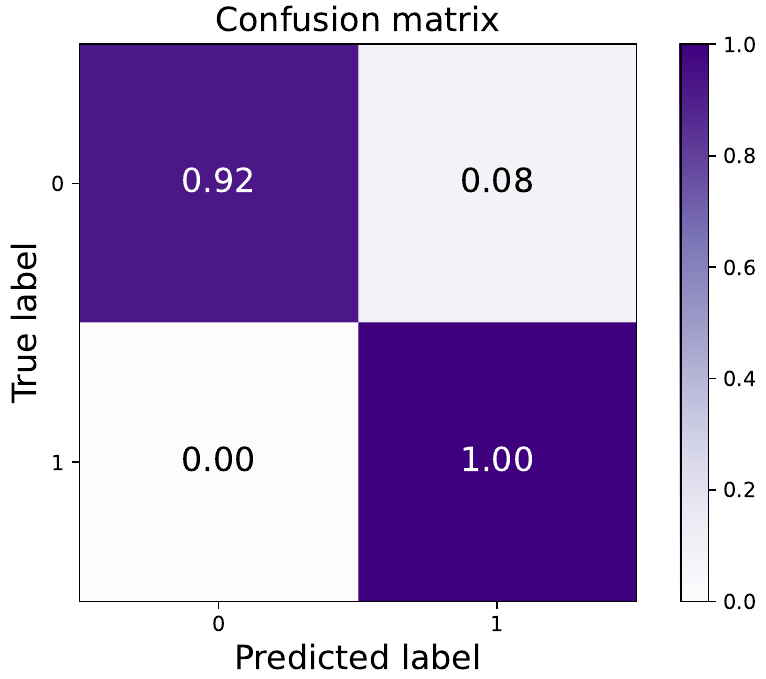}
	\end{minipage}}%
	\caption{Classification accuracy for each class of different models.}
	\Description{Classification accuracy for each class of different models.}
	\label{F-each_class}
\end{figure}

Both the insertion and removal of backdoors inherently compromise the inherent features of the affected classes, leading to diminished accuracy for the attacker's targeted classes. We illustrate these findings through a confusion matrix, as depicted in Fig \ref{F-each_class}. The empirical results indicate that while the classification accuracy for the target classes is somewhat diminished, we effectively address this issue by employing the attention relation congruence. By targeting the distillation process specifically at backdoor features, we can achieve a precise removal of these features, substantially alleviating the accuracy degradation.

\begin{figure*}
	\centering
	\subfigure[ASR with Sub-BA method.]{
		\begin{minipage}[t]{0.155\linewidth}
			\centering
			\includegraphics[width=1\linewidth]{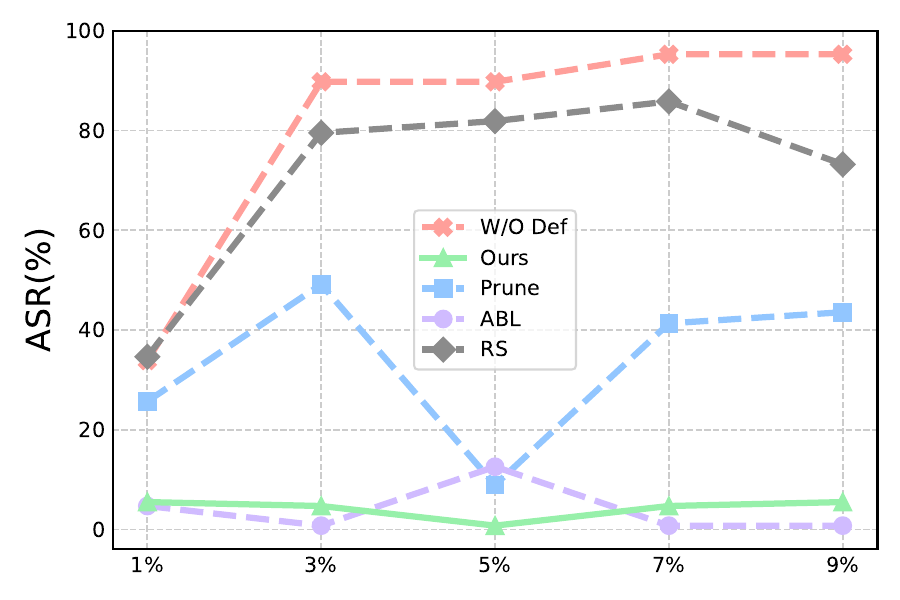}
	\end{minipage}}
        \subfigure[ACC with Sub-BA method.]{
		\begin{minipage}[t]{0.155\linewidth}
			\centering
			\includegraphics[width=1\linewidth]{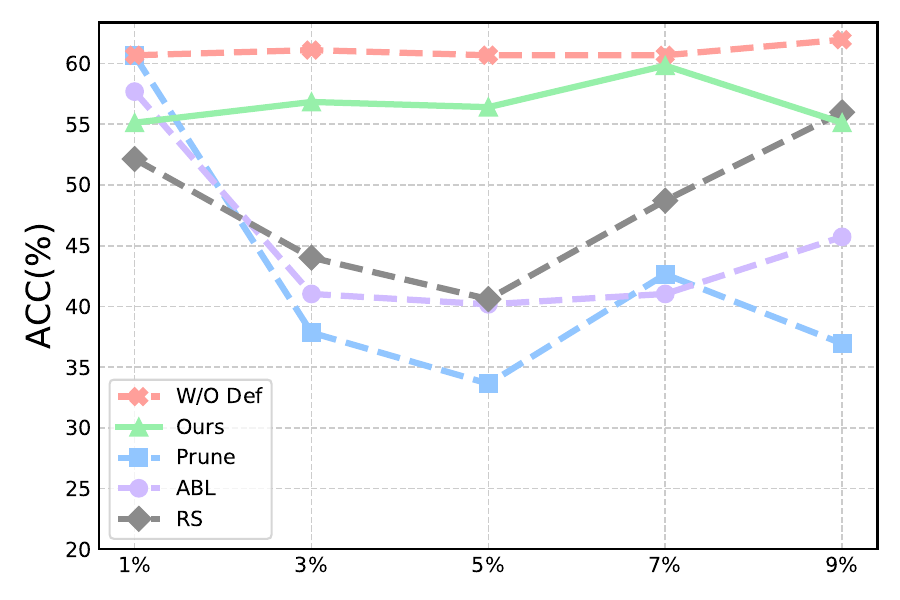}
	\end{minipage}}
	\subfigure[ASR with GTA method.]{
		\begin{minipage}[t]{0.155\linewidth}
			\centering
			\includegraphics[width=1\linewidth]{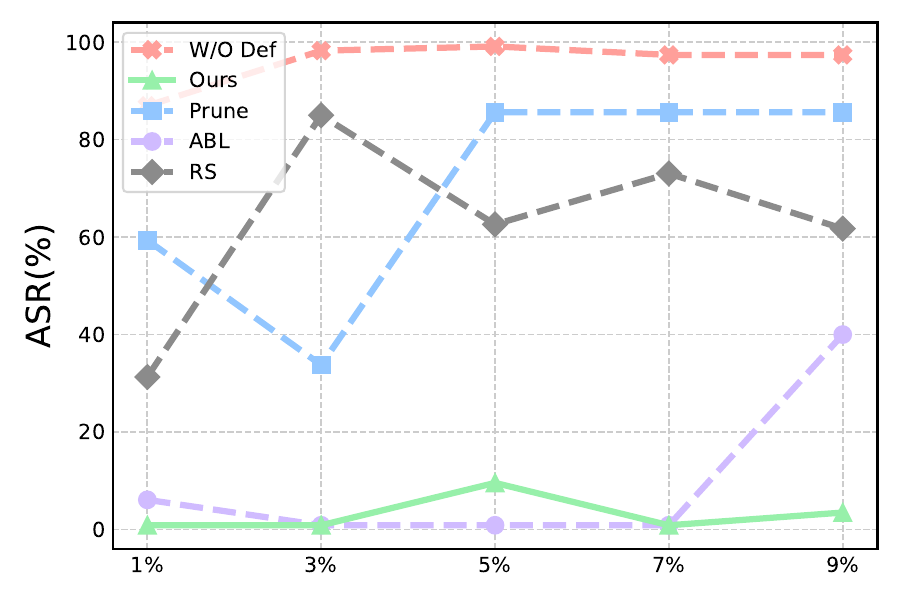}
	\end{minipage}}
	\subfigure[ACC with GTA method.]{
		\begin{minipage}[t]{0.155\linewidth}
			\centering
			\includegraphics[width=1\linewidth]{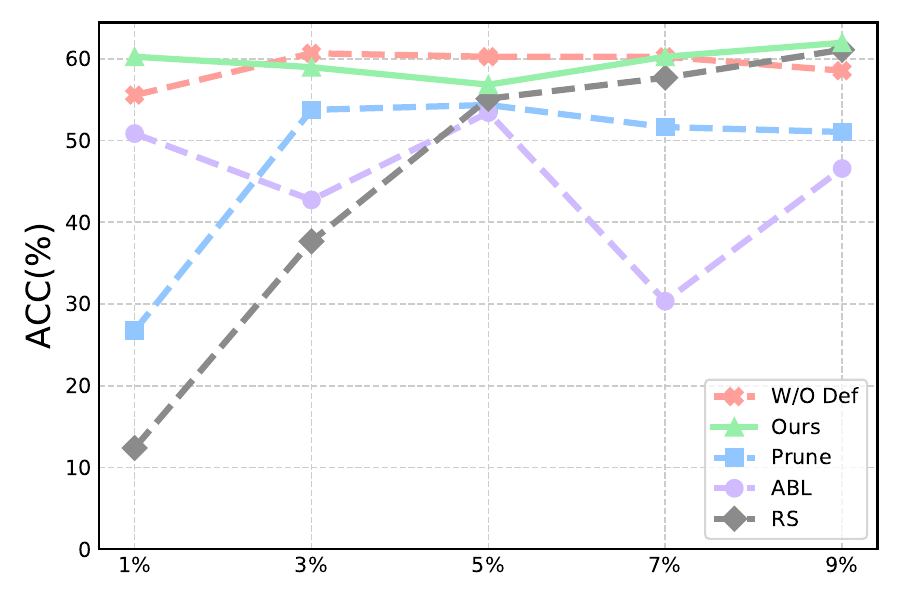}
	\end{minipage}}
	\subfigure[ASR with Motif-BA method.]{
		\begin{minipage}[t]{0.155\linewidth}
			\centering
			\includegraphics[width=1\linewidth]{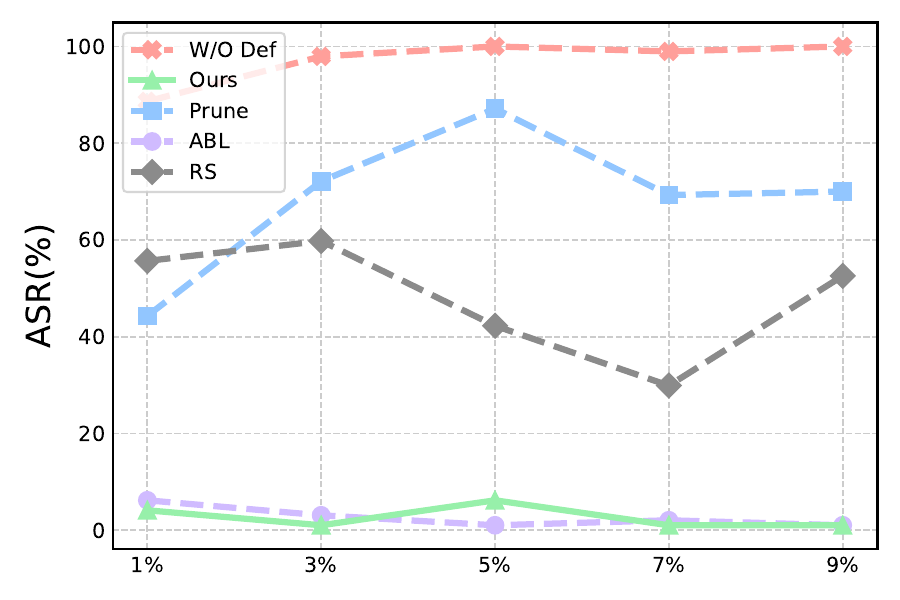}
	\end{minipage}}
	\subfigure[ACC with Motif-BA method.]{
		\begin{minipage}[t]{0.155\linewidth}
			\centering
			\includegraphics[width=1\linewidth]{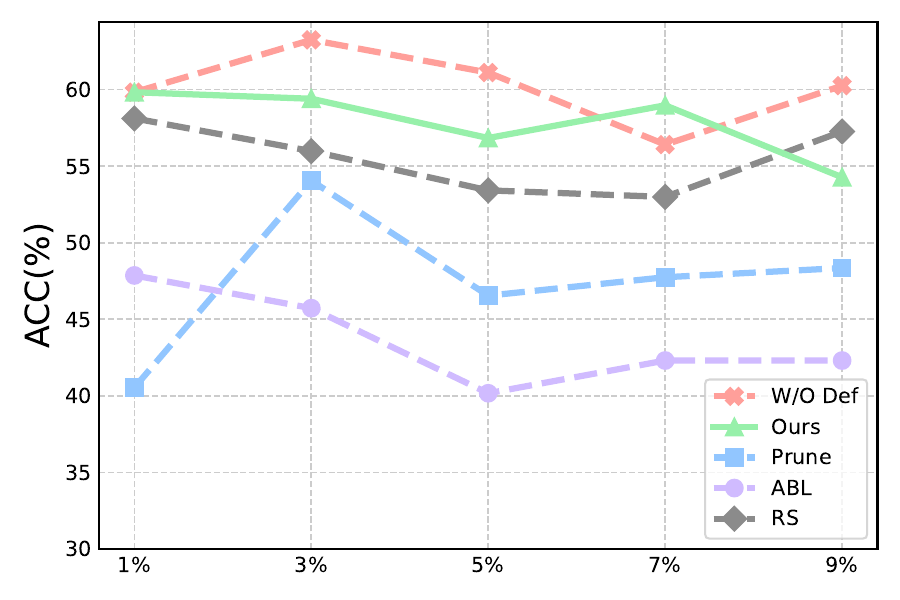}
	\end{minipage}}
	\caption{Impact of injection ratio with three SOTA methods on Fingerprint dataset.}
	\Description{Impact of injection ratio with three SOTA methods on Fingerprint dataset.}
	\label{F-injection_ratio_Fingerprint}
\end{figure*}

\begin{figure*}
	\centering
	\subfigure[ASR with Sub-BA method.]{
		\begin{minipage}[t]{0.15\linewidth}
			\centering
			\includegraphics[width=1\linewidth]{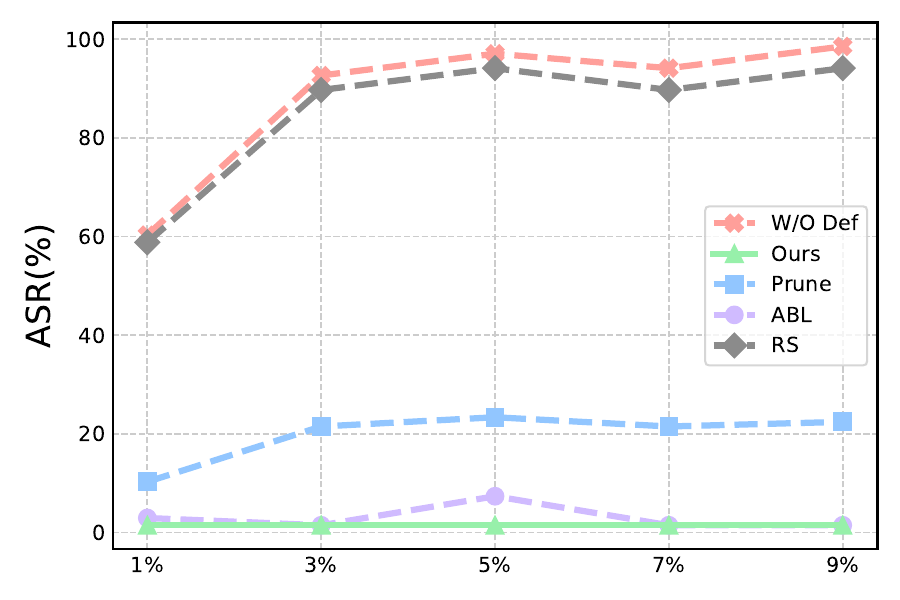}
	\end{minipage}}
        \subfigure[ACC with Sub-BA method.]{
		\begin{minipage}[t]{0.15\linewidth}
			\centering
			\includegraphics[width=1\linewidth]{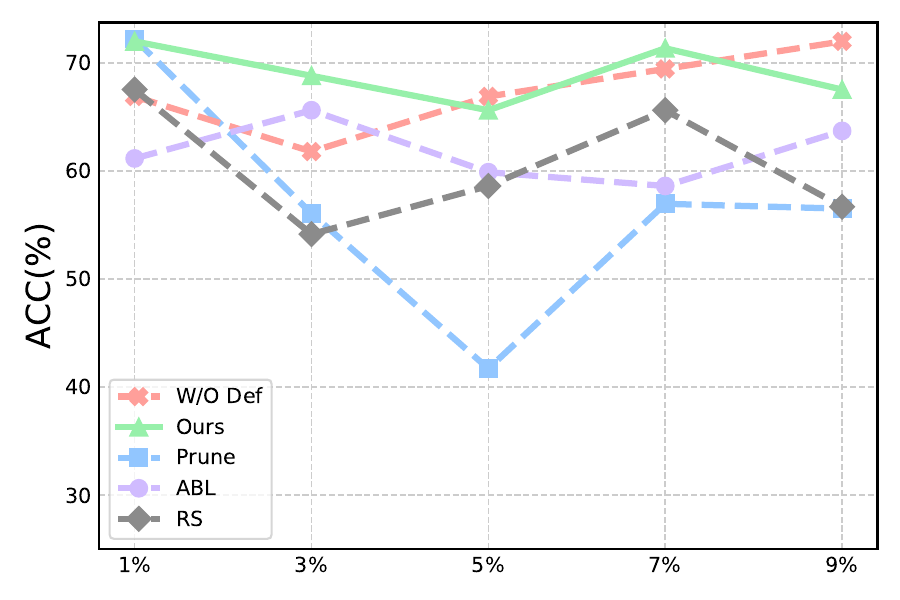}
	\end{minipage}}
	\subfigure[ASR with GTA method.]{
		\begin{minipage}[t]{0.15\linewidth}
			\centering
			\includegraphics[width=1\linewidth]{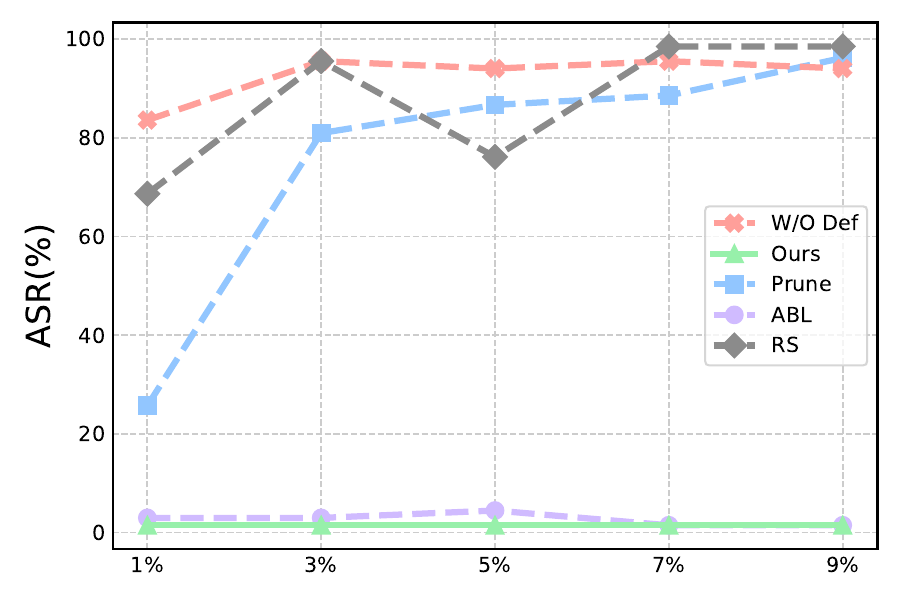}
	\end{minipage}}
	\subfigure[ACC with GTA method.]{
		\begin{minipage}[t]{0.15\linewidth}
			\centering
			\includegraphics[width=1\linewidth]{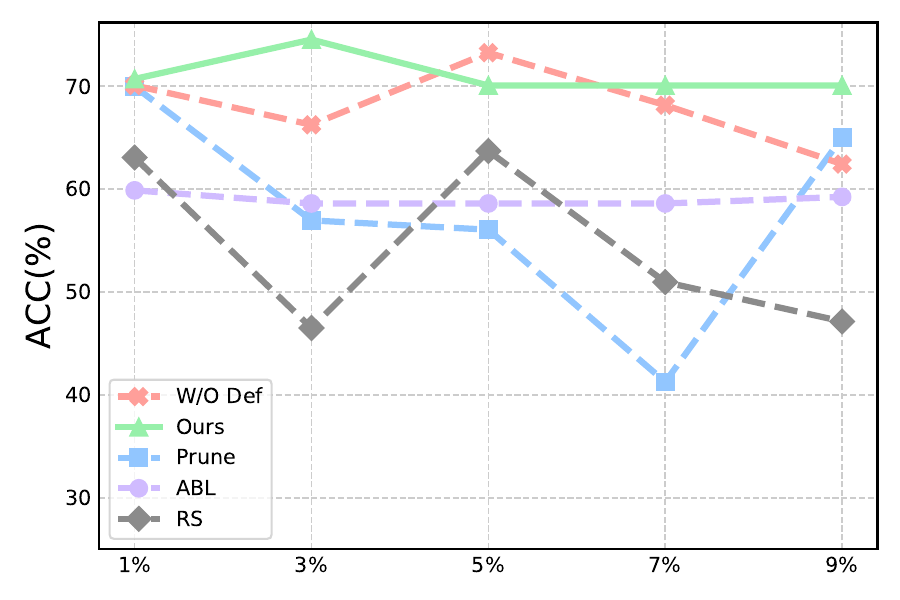}
	\end{minipage}}
	\subfigure[ASR with Motif-BA method.]{
		\begin{minipage}[t]{0.15\linewidth}
			\centering
			\includegraphics[width=1\linewidth]{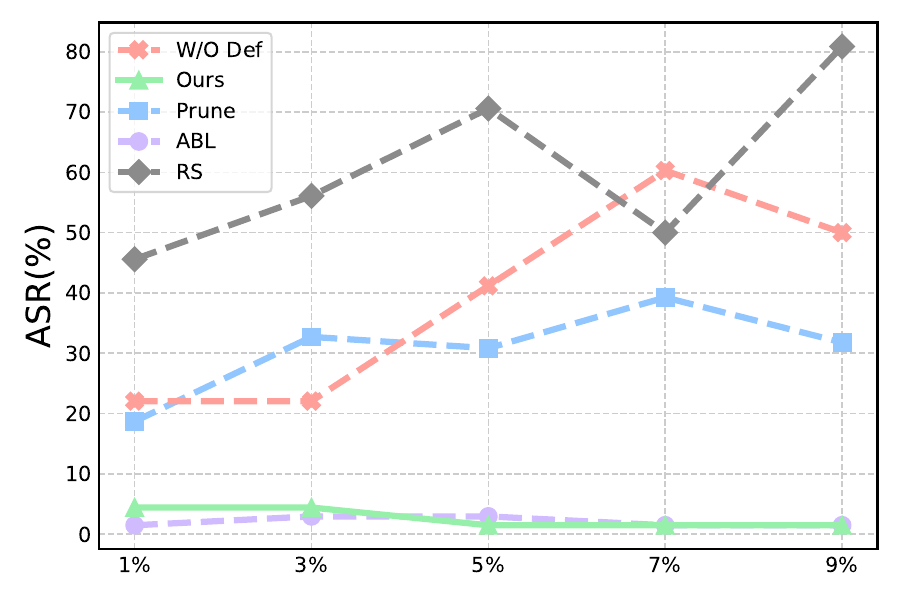}
	\end{minipage}}
	\subfigure[ACC with Motif-BA method.]{
		\begin{minipage}[t]{0.15\linewidth}
			\centering
			\includegraphics[width=1\linewidth]{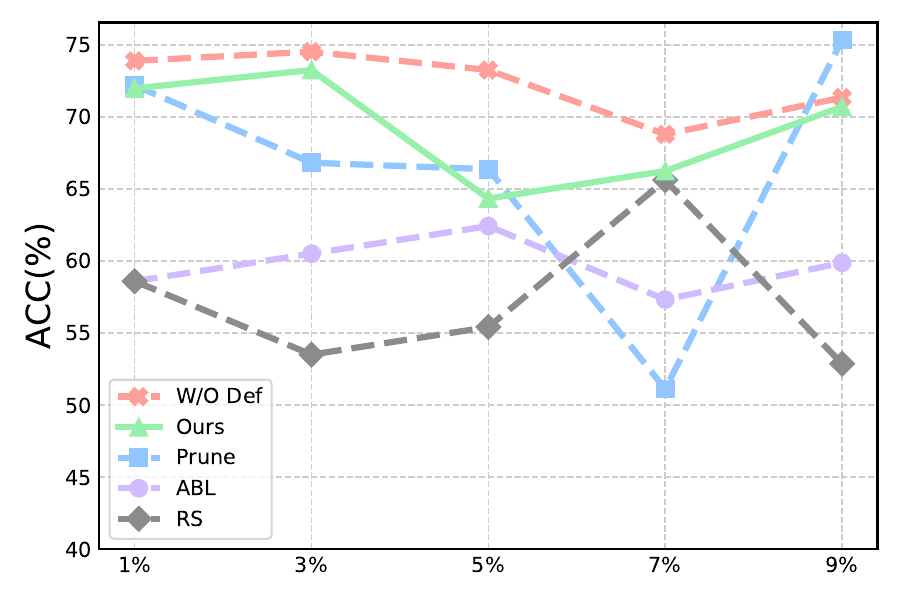}
	\end{minipage}}
	\caption{Impact of injection ratio with three SOTA methods on PROTEINS dataset.}
	\Description{Impact of injection ratio with three SOTA methods on PROTEINS dataset.}
	\label{F-injection_ratio_PROTEINS}
\end{figure*}

\begin{figure*}
	\centering
	\subfigure[ASR on PROTEINS dataset.]{
		\begin{minipage}[t]{0.235\linewidth}
			\centering
			\includegraphics[width=1\linewidth]{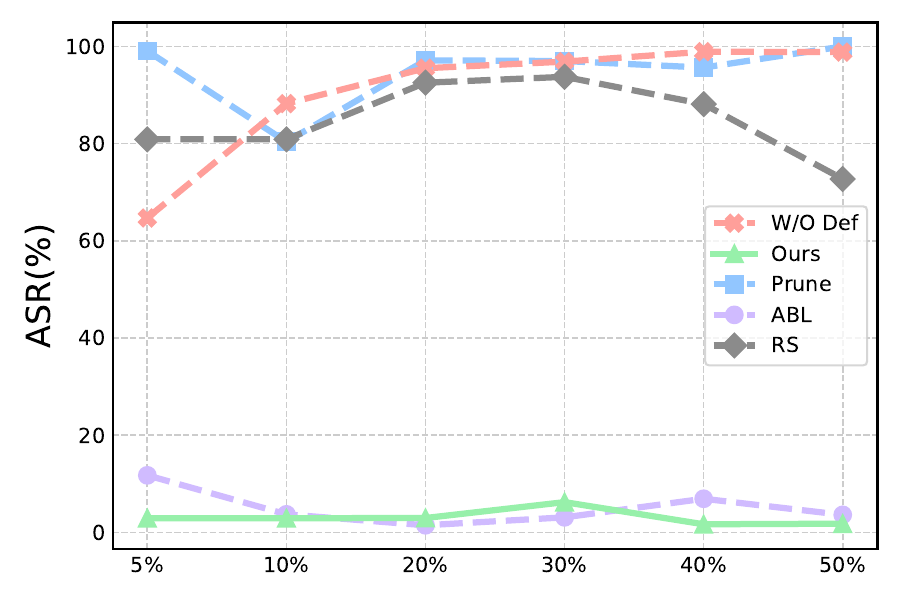}
	\end{minipage}}
	\subfigure[ASR on Fingerprint dataset.]{
		\begin{minipage}[t]{0.235\linewidth}
			\centering
			\includegraphics[width=1\linewidth]{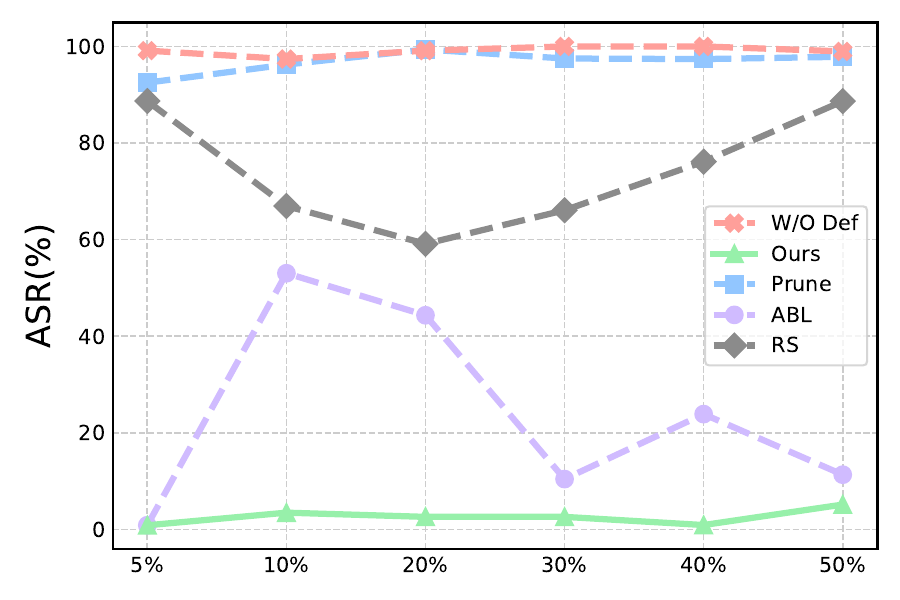}
	\end{minipage}}
	\subfigure[ASR on AIDS dataset.]{
		\begin{minipage}[t]{0.235\linewidth}
			\centering
			\includegraphics[width=1\linewidth]{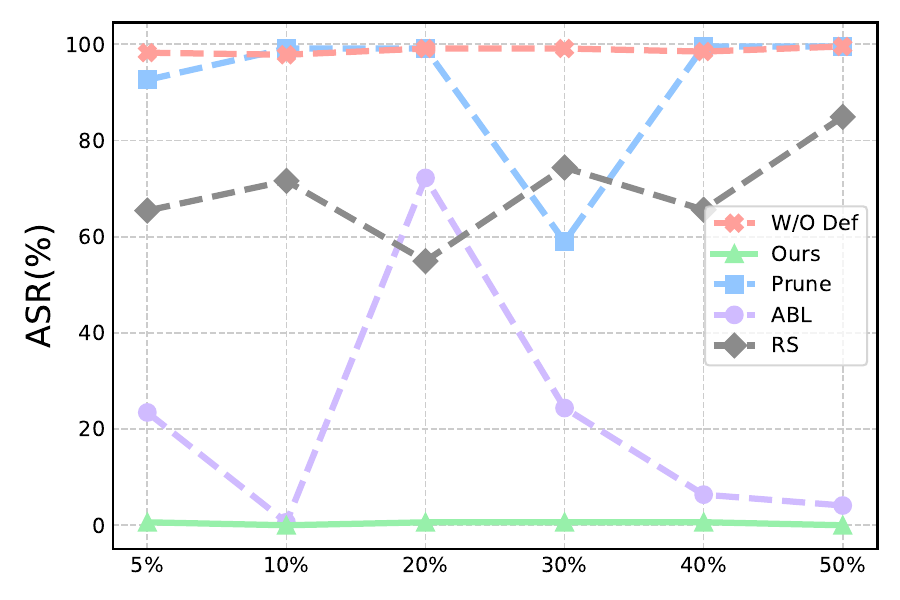}
	\end{minipage}}
	\subfigure[ASR on COLLAB dataset.]{
		\begin{minipage}[t]{0.235\linewidth}
			\centering
			\includegraphics[width=1\linewidth]{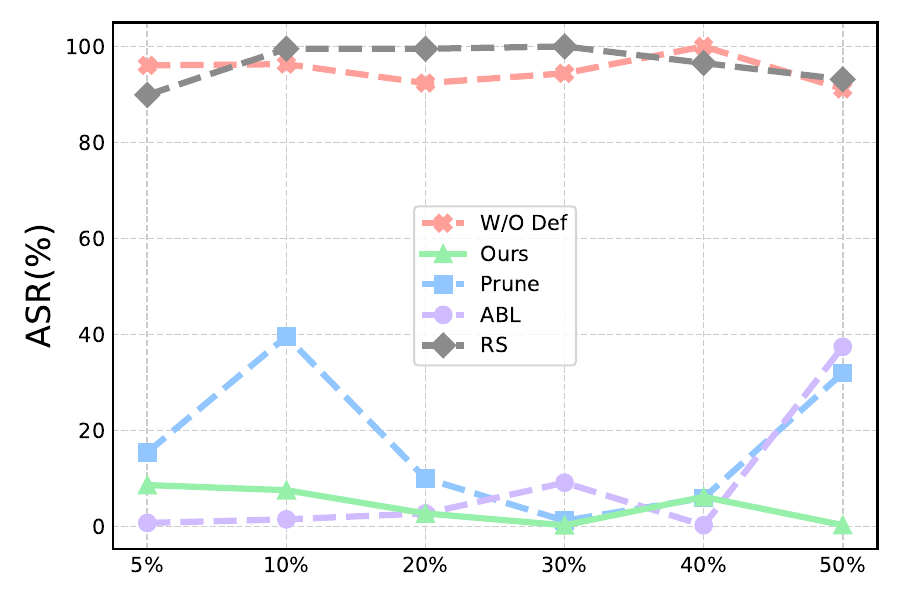}
	\end{minipage}}%
 \\
 	\subfigure[ACC on PROTEINS dataset.]{
		\begin{minipage}[t]{0.235\linewidth}
			\centering
			\includegraphics[width=1\linewidth]{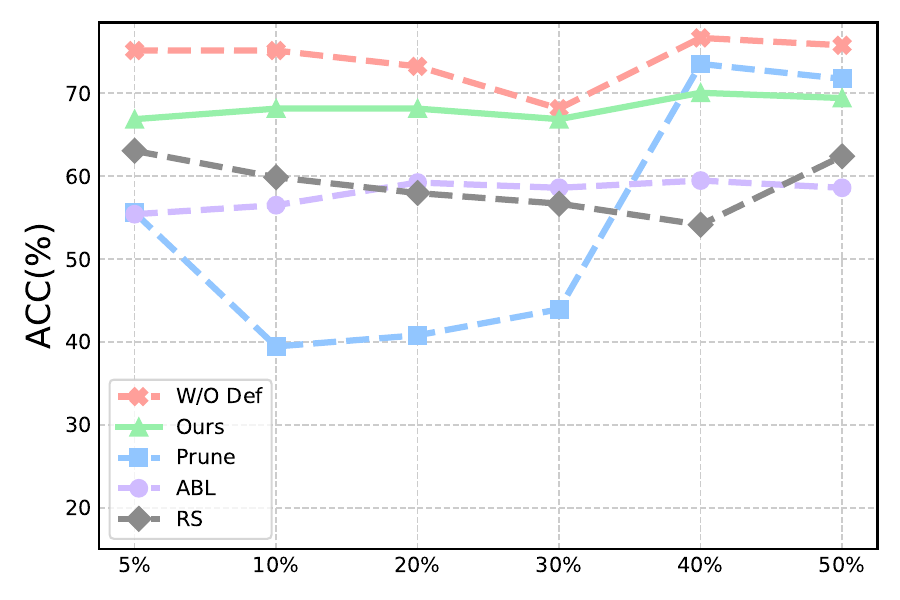}
	\end{minipage}}
	\subfigure[ACC on Fingerprint dataset.]{
		\begin{minipage}[t]{0.235\linewidth}
			\centering
			\includegraphics[width=1\linewidth]{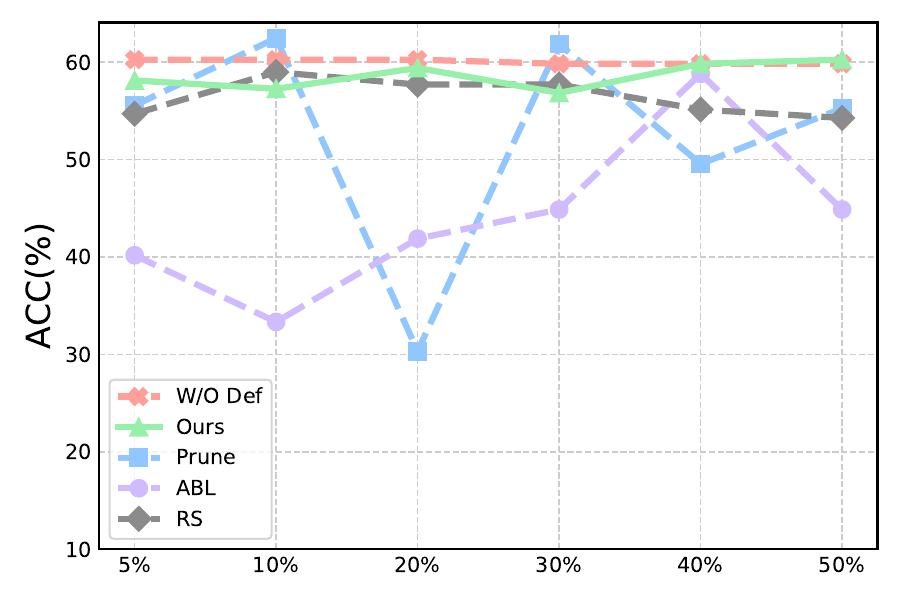}
	\end{minipage}}
	\subfigure[ACC on AIDS dataset.]{
		\begin{minipage}[t]{0.235\linewidth}
			\centering
			\includegraphics[width=1\linewidth]{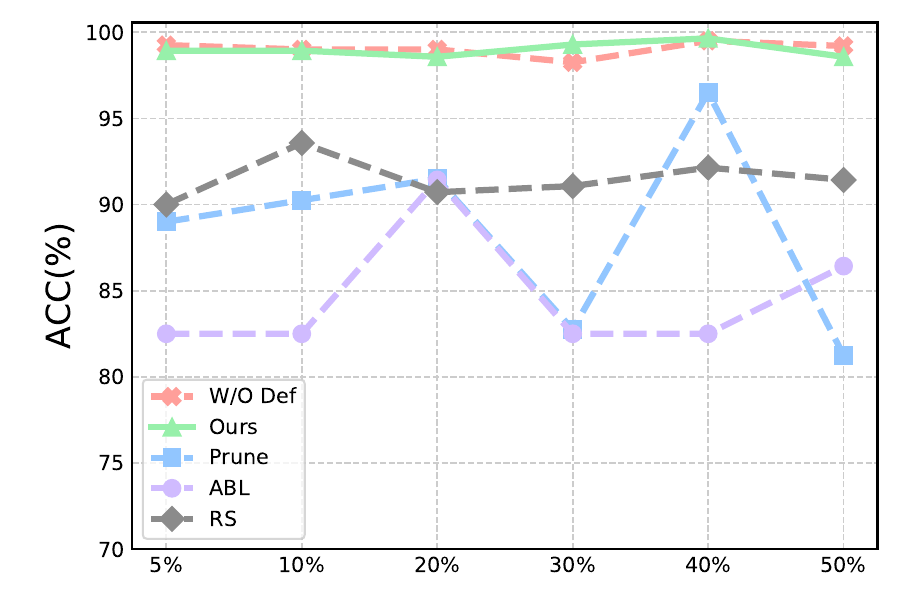}
	\end{minipage}}
	\subfigure[ACC on COLLAB dataset.]{
		\begin{minipage}[t]{0.235\linewidth}
			\centering
			\includegraphics[width=1\linewidth]{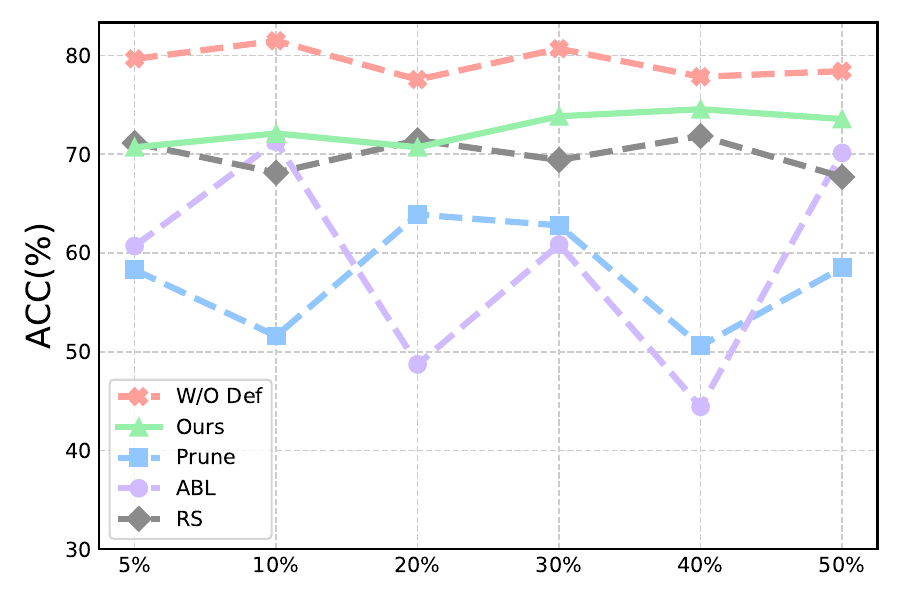}
	\end{minipage}}%
	\caption{Impact of trigger size on four datasets with GTA attack method.}
	\Description{Impact of trigger size on four datasets with GTA attack method.}
	\label{F-trigger_size}
\end{figure*}

\section{Impact of Trigger Size}
We examined the influence of trigger size on \sysname across various datasets, with the findings depicted in Figure \ref{F-trigger_size}. The X-axis denotes trigger size, ranging from $5\%$ to $50\%$ of the original graph size.In the absence of any defense mechanisms, we observed an interesting trend regarding the success rate of backdoor attacks across different datasets. For datasets where the original backdoor ASR is relatively low (e.g., PROTEINS), an increase in the size of the trigger leads to a corresponding rise in the backdoor attack success rate. However, for datasets where the ASR is already high, no clear pattern emerges with respect to trigger size. Correspondingly, the size of the trigger also has a certain impact on the performance of the main task. Overall, our proposed method is robust to variations in trigger size and can consistently reduce the ASR to approximately $0\%$ under any trigger size. This robustness stems from the fact that, regardless of the trigger size, the successful injection of backdoor results in notable discrepancies between the intermediate layer outputs (or the activation patterns of backdoor neurons) of the backdoor model and a fine-tuned clean model. Utilizing graph-based attention mechanisms to accentuate these disparities, our approach effectively counters backdoor influences without being contingent on trigger size. In contrast, the other three defense methods exhibit significant fluctuations in performance and even complete failure in some cases.

Specifically, ABL attempts to eliminate backdoors by unlearning backdoor patterns during the training phase. However, when the trigger size of backdoor samples varies, the effectiveness of ABL's forgetting mechanism also fluctuates. This inconsistency disrupts its ability to maintain a balance between accuracy on ACC and ASR. As the trigger size increases, ABL struggles to balance the two metrics, often leading to excessive forgetting. That is, while ABL may successfully reduce ASR, it does so at the cost of severely degrading ACC.
Conversely, both the Prune and RS defenses experience a downturn in their protective capabilities with the enlargement of the trigger size. The Prune method, which employs node pruning, suffers from an increased rate of false positives as the number of trigger nodes expands, thus impairing the defense's potency. Meanwhile, RS, which incorporates random sampling in its training phase, sees a rise in the probability of encountering trigger nodes as the trigger size grows, leading to an uptick in the ASR.

In summary, our strategy has demonstrated outstanding resilience across a spectrum of backdoor injection scenarios, showcasing its robustness and efficacy in countering backdoor threats.

\section{Effect of Injection Ratio}
\label{A_injection-ratio}
In this section, we delve into the impact of injection ratios on our method across several other datasets. As illustrated in Fig. \ref{F-injection_ratio_Fingerprint} and Fig. \ref{F-injection_ratio_PROTEINS}, the results indicate that regardless of the injection ratio employed by the attacker on different datasets, our method consistently reduces the success rate of backdoor attacks to around 10$\%$. This demonstrates the method's robustness and its ability to mitigate backdoor attacks effectively, showcasing its broad applicability independent of the injection ratio and dataset characteristics.

\begin{figure*}[t]
	\centering
	\subfigure[Backdoored model with Sub-BA.]{
		\begin{minipage}[t]{0.15\linewidth}
			\centering
			\includegraphics[width=1\linewidth]{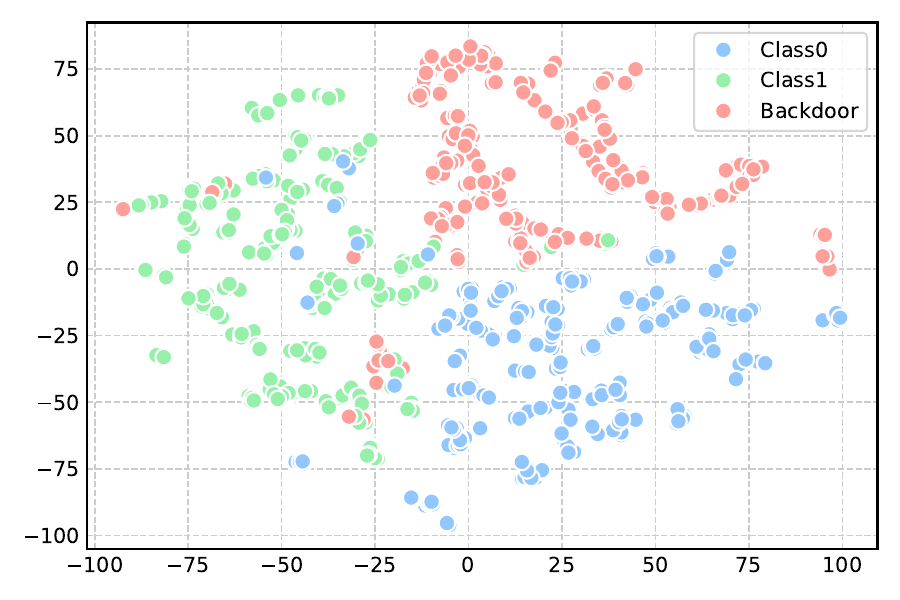}
	\end{minipage}}
	\subfigure[Purified model with Sub-BA.]{
		\begin{minipage}[t]{0.15\linewidth}
			\centering
			\includegraphics[width=1\linewidth]{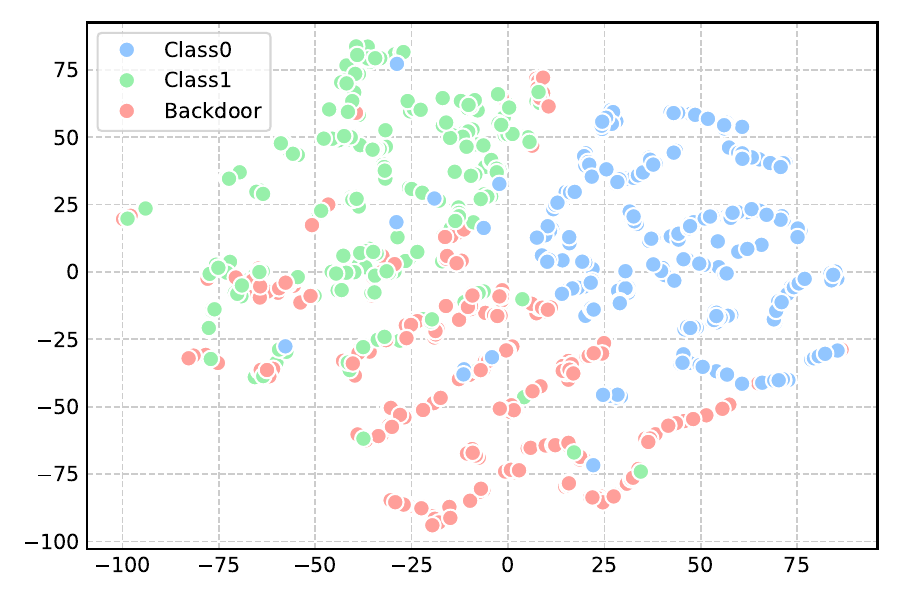}
	\end{minipage}}
	\subfigure[Backdoored model with GTA.]{
		\begin{minipage}[t]{0.15\linewidth}
			\centering
			\includegraphics[width=1\linewidth]{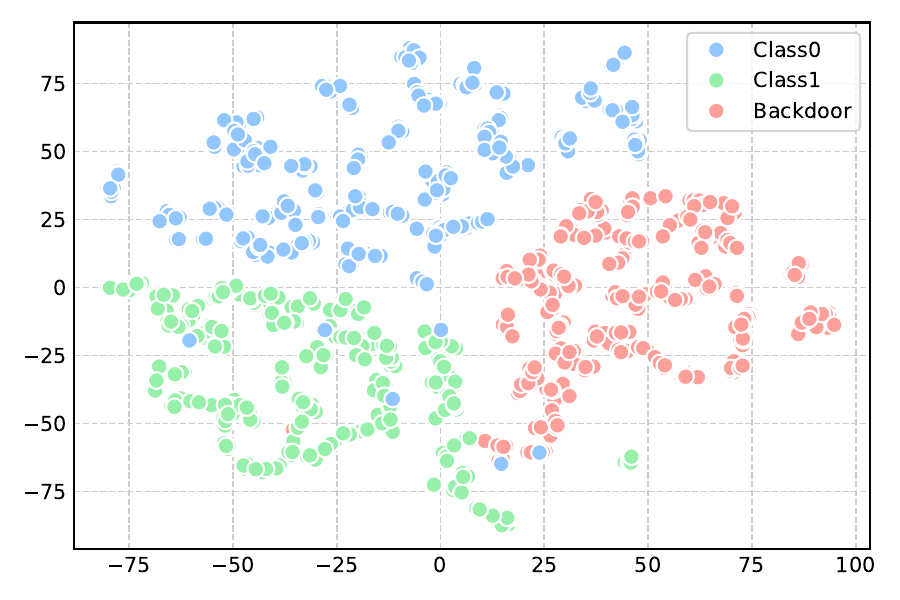}
	\end{minipage}}
	\subfigure[Purified model with GTA.]{
		\begin{minipage}[t]{0.15\linewidth}
			\centering
			\includegraphics[width=1\linewidth]{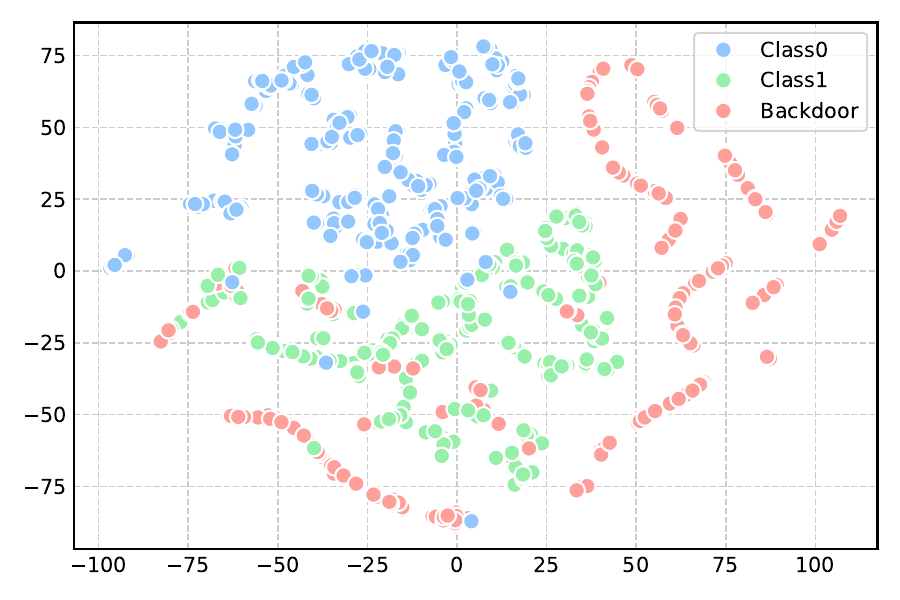}
	\end{minipage}}
	\subfigure[Backdoored model with Motif-BA.]{
		\begin{minipage}[t]{0.15\linewidth}
			\centering
			\includegraphics[width=1\linewidth]{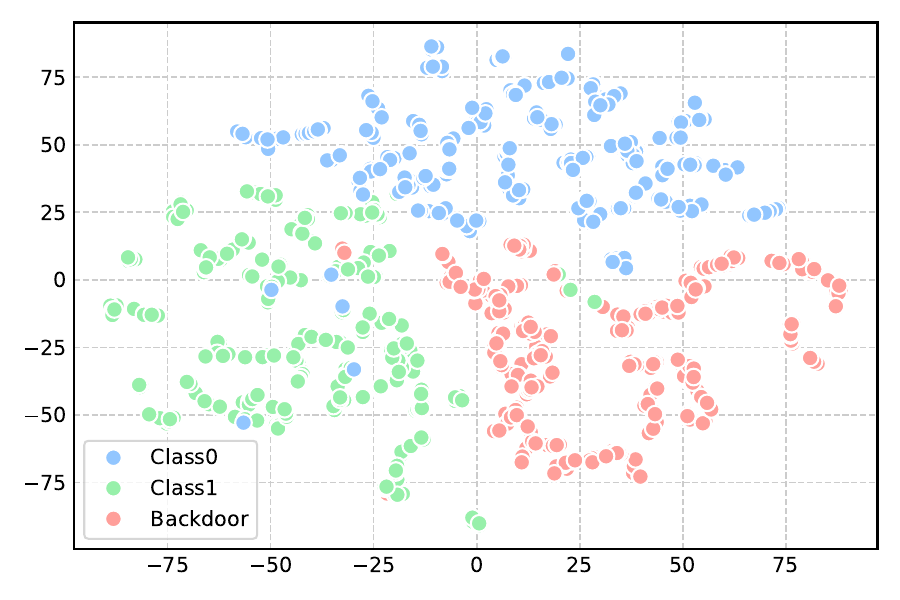}
	\end{minipage}}
	\subfigure[Purified model with Motif-BA.]{
		\begin{minipage}[t]{0.15\linewidth}
			\centering
			\includegraphics[width=1\linewidth]{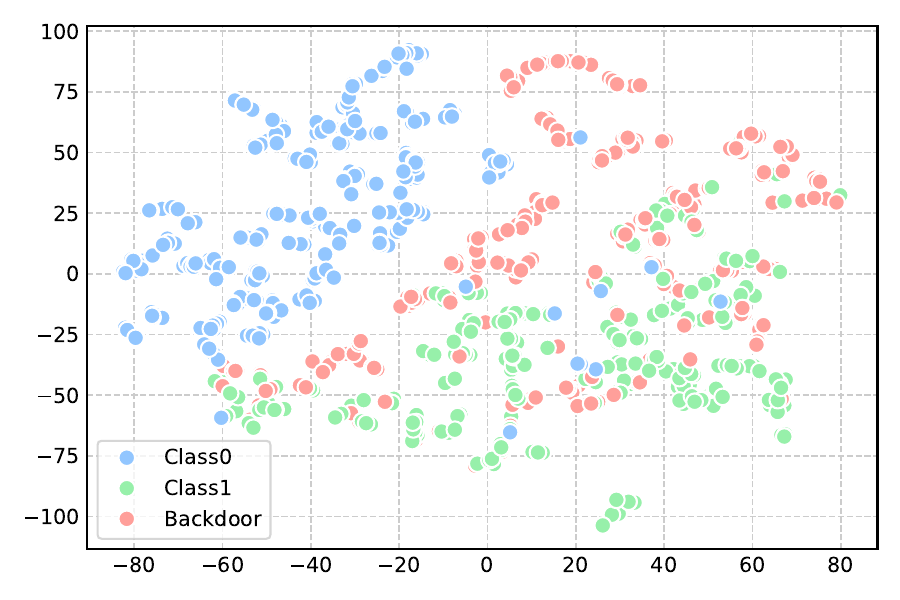}
	\end{minipage}}
	\caption{Visualizations of feature embedding of the backdoored model and purified model.}
	\Description{Visualizations of feature embedding of the backdoored model and purified model.}
	\label{F-TSNE}
\end{figure*}

\begin{figure*}
	\centering
	\subfigure[ASR on PROTEINS dataset.]{
		\begin{minipage}[t]{0.235\linewidth}
			\centering
			\includegraphics[width=1\linewidth]{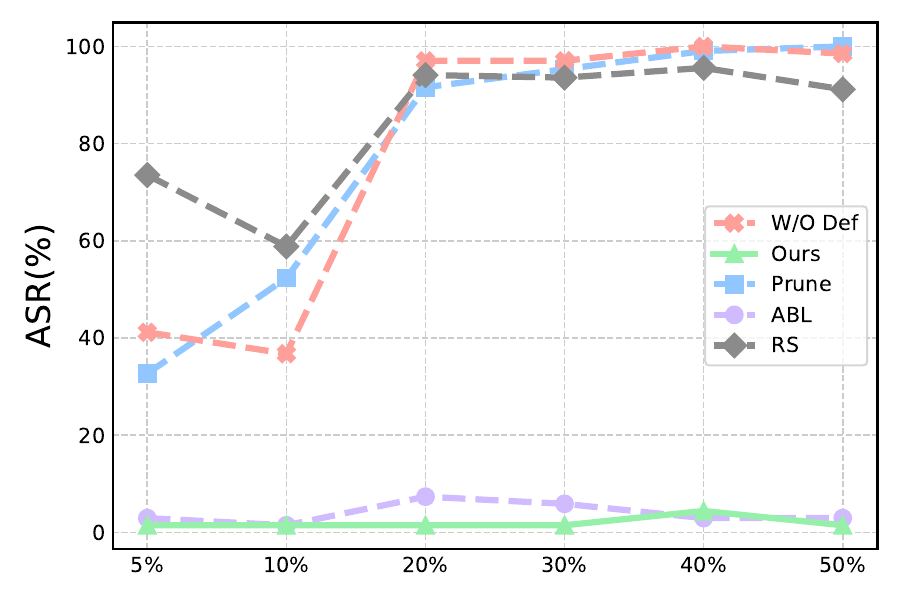}
	\end{minipage}}
	\subfigure[ASR on Fingerprint dataset.]{
		\begin{minipage}[t]{0.235\linewidth}
			\includegraphics[width=1\linewidth]{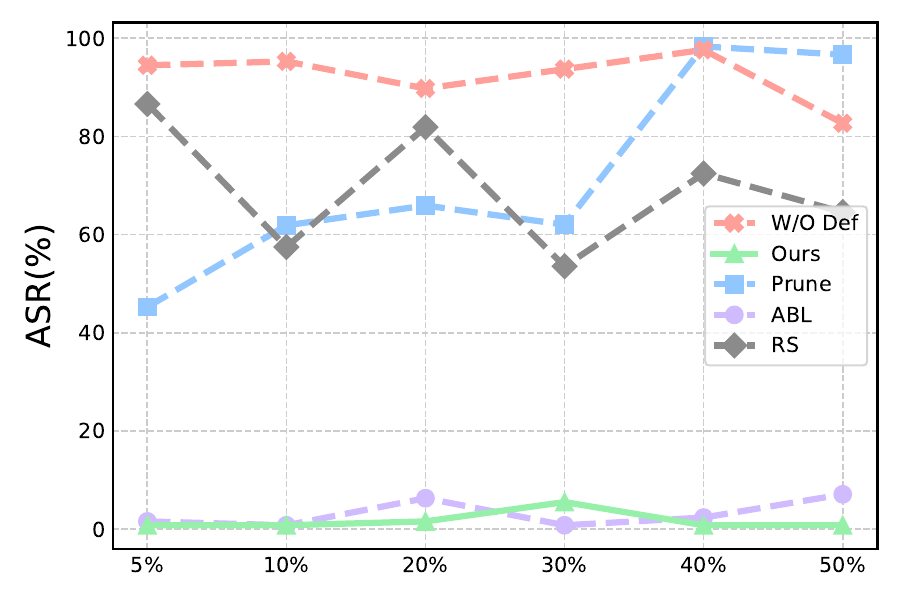}
	\end{minipage}}
	\subfigure[ASR on AIDS dataset.]{
		\begin{minipage}[t]{0.235\linewidth}
			\centering
			\includegraphics[width=1\linewidth]{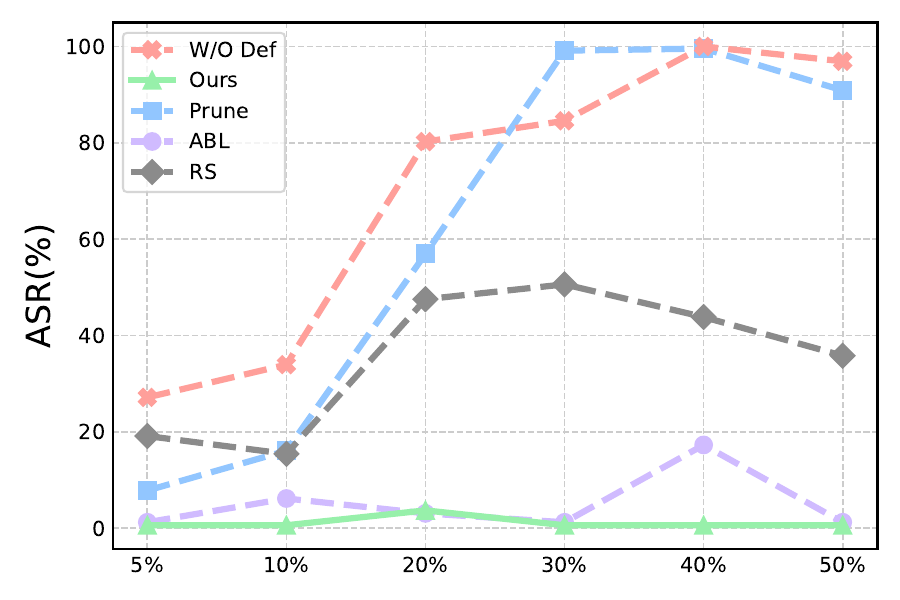}
	\end{minipage}}
	\subfigure[ASR on COLLAB dataset.]{
		\begin{minipage}[t]{0.235\linewidth}
			\includegraphics[width=1\linewidth]{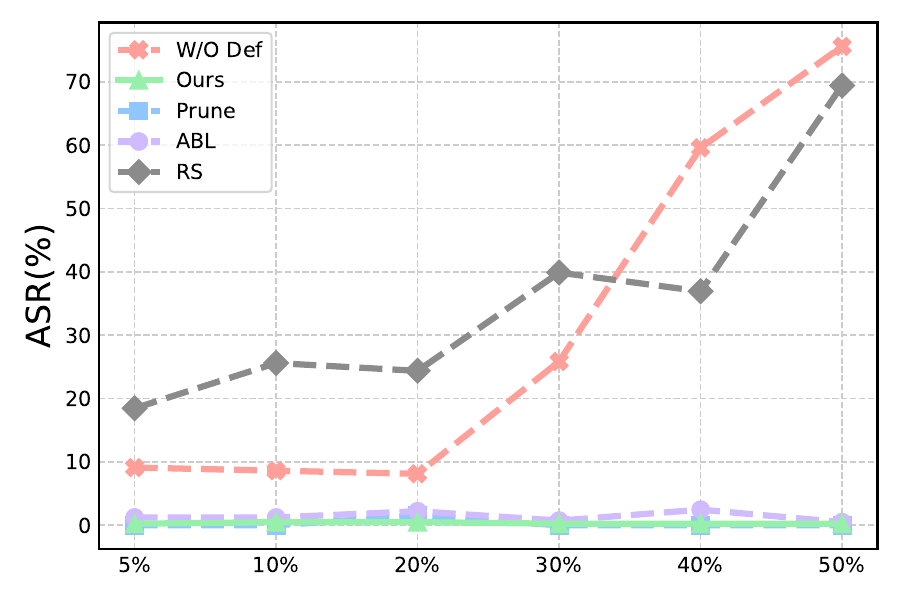}
	\end{minipage}}%
 \\
 	\subfigure[ACC on PROTEINS dataset.]{
		\begin{minipage}[t]{0.235\linewidth}
			\centering
			\includegraphics[width=1\linewidth]{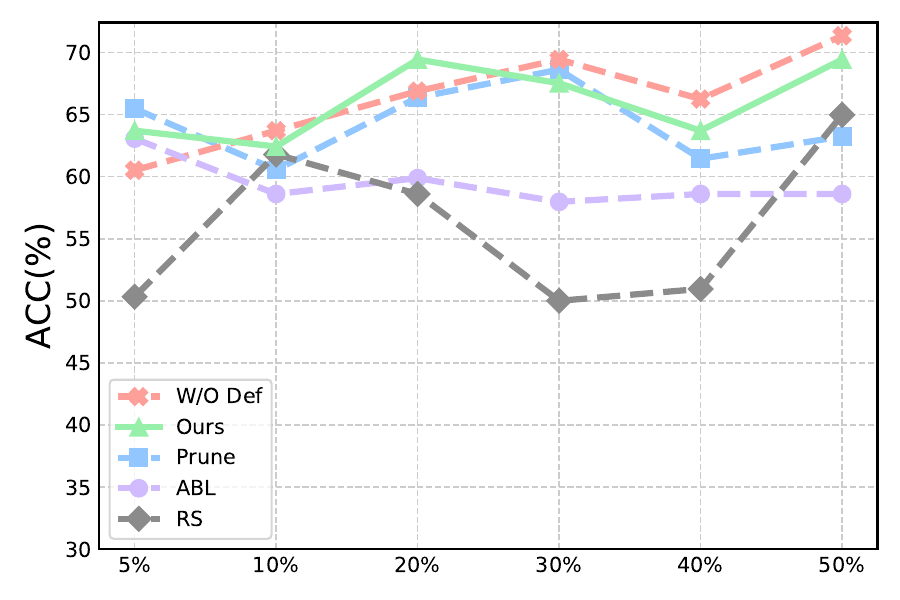}
	\end{minipage}}
	\subfigure[ACC on Fingerprint dataset.]{
		\begin{minipage}[t]{0.235\linewidth}
			\centering
			\includegraphics[width=1\linewidth]{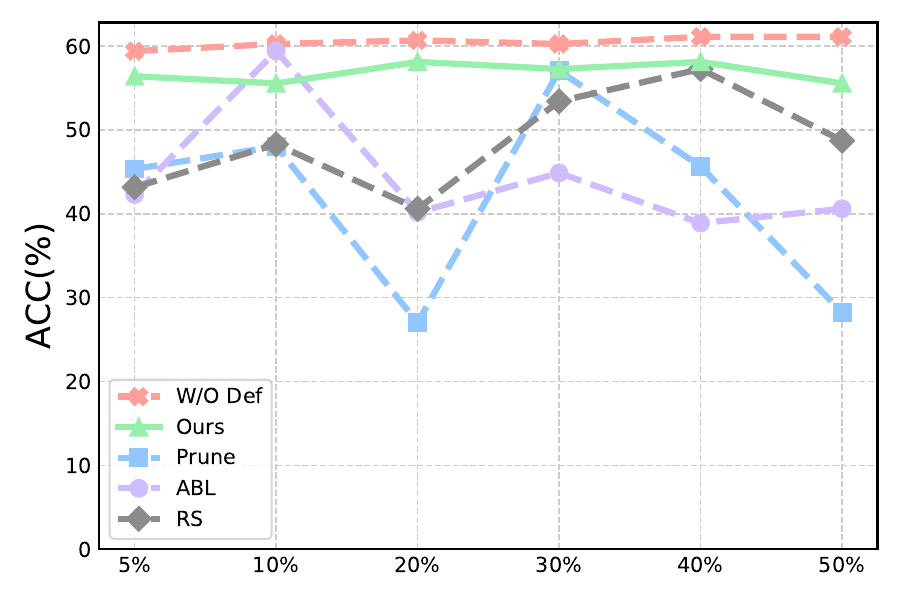}
	\end{minipage}}
	\subfigure[ACC on AIDS dataset.]{
		\begin{minipage}[t]{0.235\linewidth}
			\centering
			\includegraphics[width=1\linewidth]{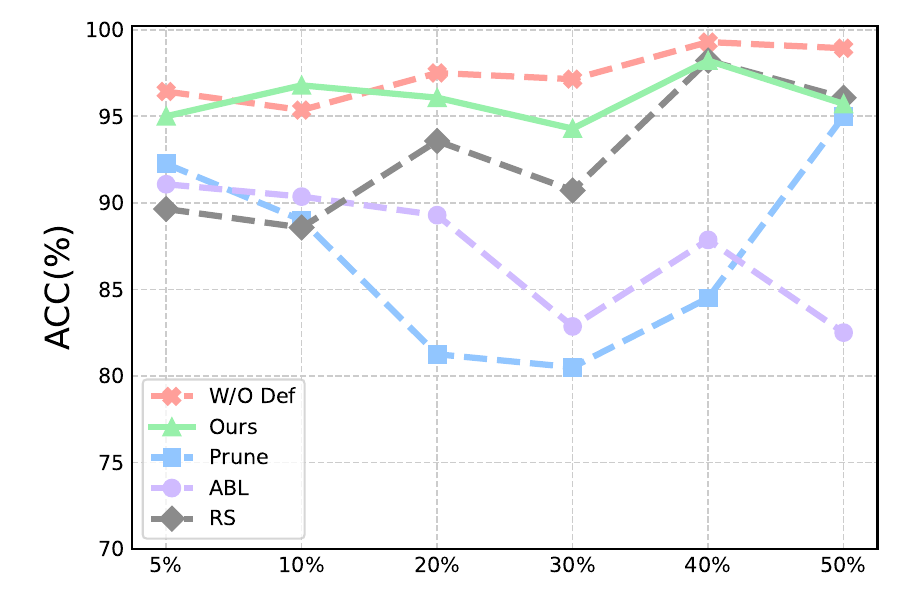}
	\end{minipage}}
	\subfigure[ACC on COLLAB dataset.]{
		\begin{minipage}[t]{0.235\linewidth}
			\includegraphics[width=1\linewidth]{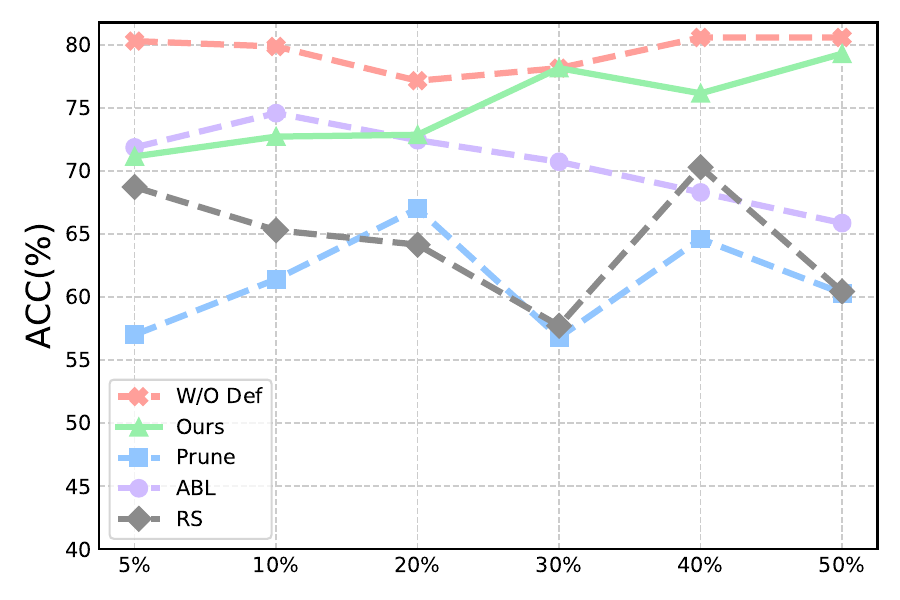}
	\end{minipage}}%
	\caption{Impact of trigger size on four datasets using the Sub-BA attack.}
	\Description{Impact of trigger size on four datasets using the Sub-BA attack.}
	\label{F-trigger_size_sub-BA}
\end{figure*}

\section{Algorithm}
\label{A-Algorithm}
Algorithm 1 summarizes the workflow of our method. Our approach takes a backdoor model as input and assumes a certain proportion of clean datasets, sourced from either the test set or the validation set, maintaining the same distribution as the original training data. In extreme cases, these clean datasets could originate from publicly available datasets, representing different distributions.

The process involves fine-tuning the backdoored model as the teacher model and the backdoored model as the student model. This is followed by the distillation process of knowledge using attention maps and relation congruence. Ultimately, the output is a clean model that has been refined through this process.

\begin{algorithm}
\caption{Algorithm of \sysname process}\label{Aprocess}
\KwData{Backdoored model $\tilde{F}$,clean dataset $\mathcal{D}$, \textit{fintuning epochs} $E_{fine}$, \textit{distillation epochs $E_{distill}$}, number of model layers $l$, relation \textit{pairs} }
\KwResult{Purified $F_S$}
\tcp{\footnotesize{Employing the backdoor model as the student model}}
Student model $F_S \leftarrow \tilde{F}$\;
$k \leftarrow 1$\;
\tcp{\footnotesize{Finetuning with clean samples}}
\For{each finetuning epoch $k\in[1,2,\cdots,E_{fine}]$}
{
    \For{each batch $b=\{x,y\}\in D$}
    {
    $\mathcal{L}\leftarrow CrossEntropy(\tilde{F}(x),y)$\;
    $\tilde{F} \leftarrow \tilde{F}-\eta\nabla\mathcal{L}$\;
    }
}
\tcp{\footnotesize{Employing the Finetuned model as the teacher model}}
Teacher model $F_T \leftarrow  freeze(\tilde{F})$\;
\tcp{\footnotesize{Attention Distillation}}
\For{each distillation epoch 
$k\in[1,2,\cdots,E_{distill}]$}
{
    \For{each batch $b=\{x,y\}\in D$}
    {
    $\mathcal{L}_{CE}\leftarrow CrossEntropy(F_b(x),y)$\;
    $\mathcal{L}_{AD}\leftarrow \sum^k_{l=1}\mathcal{L}_{AD}(F_T^l(x),F_S^l(x))$\;
    $\mathcal{L}_{RC}\leftarrow \sum_{i,j\in <pairs>}\tilde{W}_2(R^{ij}_T,R^{ij}_S)$\;
    $\mathcal{L}_{Total} = \mathcal{L}_{CE} + \beta \mathcal{L}_{AD}+ \gamma \mathcal{L}_{RC}$\;
    $F_{S} \leftarrow F_{S} -\eta\nabla\mathcal{L}_{Total}$\;
    }
}
\end{algorithm}

\begin{figure*}
	\centering
	\subfigure[Trigger Sample on Sub-BA attack.]{
		\label{F-backdoored_vis}
		\begin{minipage}[t]{0.49\linewidth}
			\centering
			\includegraphics[width=1\linewidth]{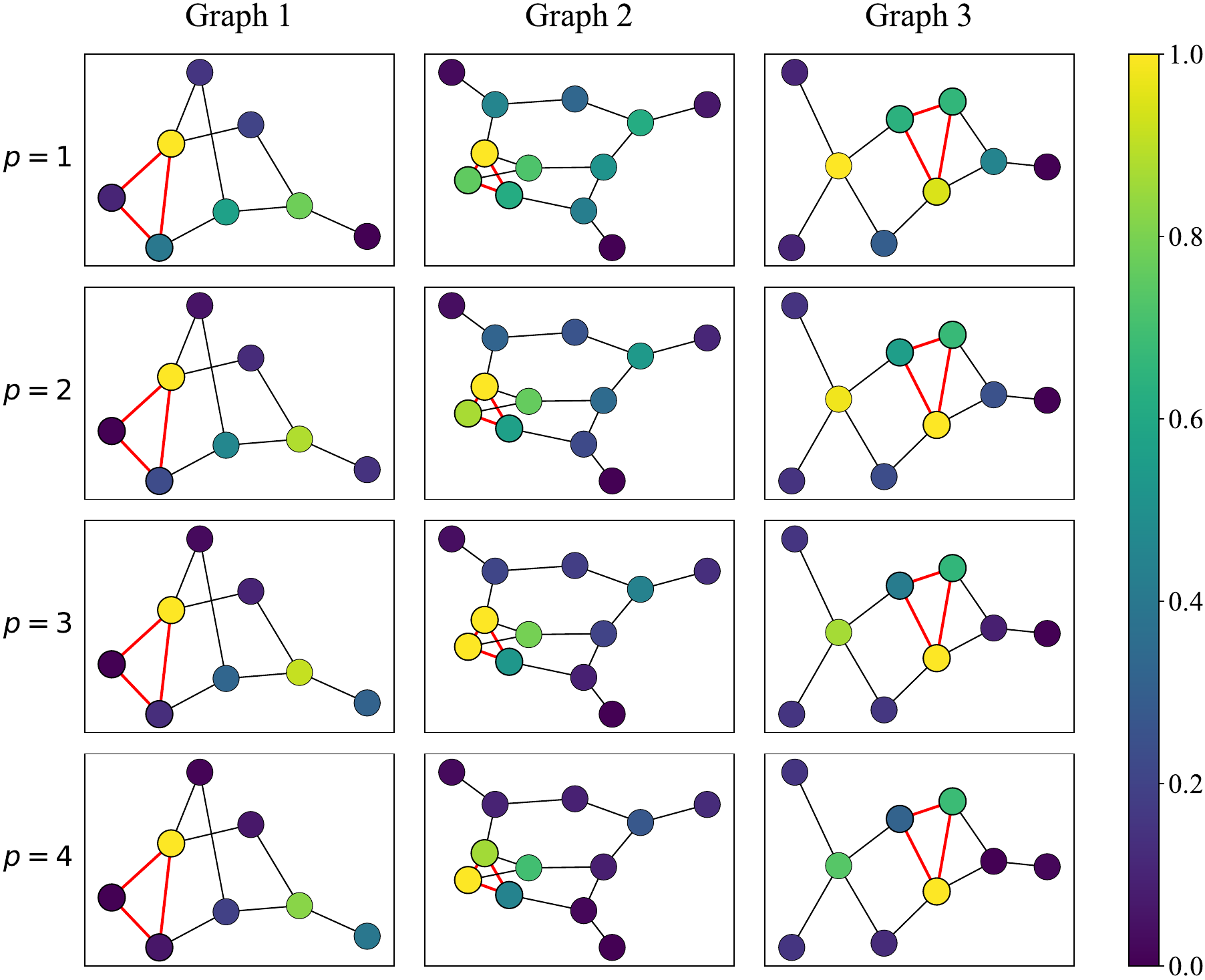}
	\end{minipage}}
	\subfigure[Trigger Sample on Sub-BA attack purified by \sysname.]{
		\label{F-purified_vis}
			\begin{minipage}[t]{0.49\linewidth}
			\includegraphics[width=1\linewidth]{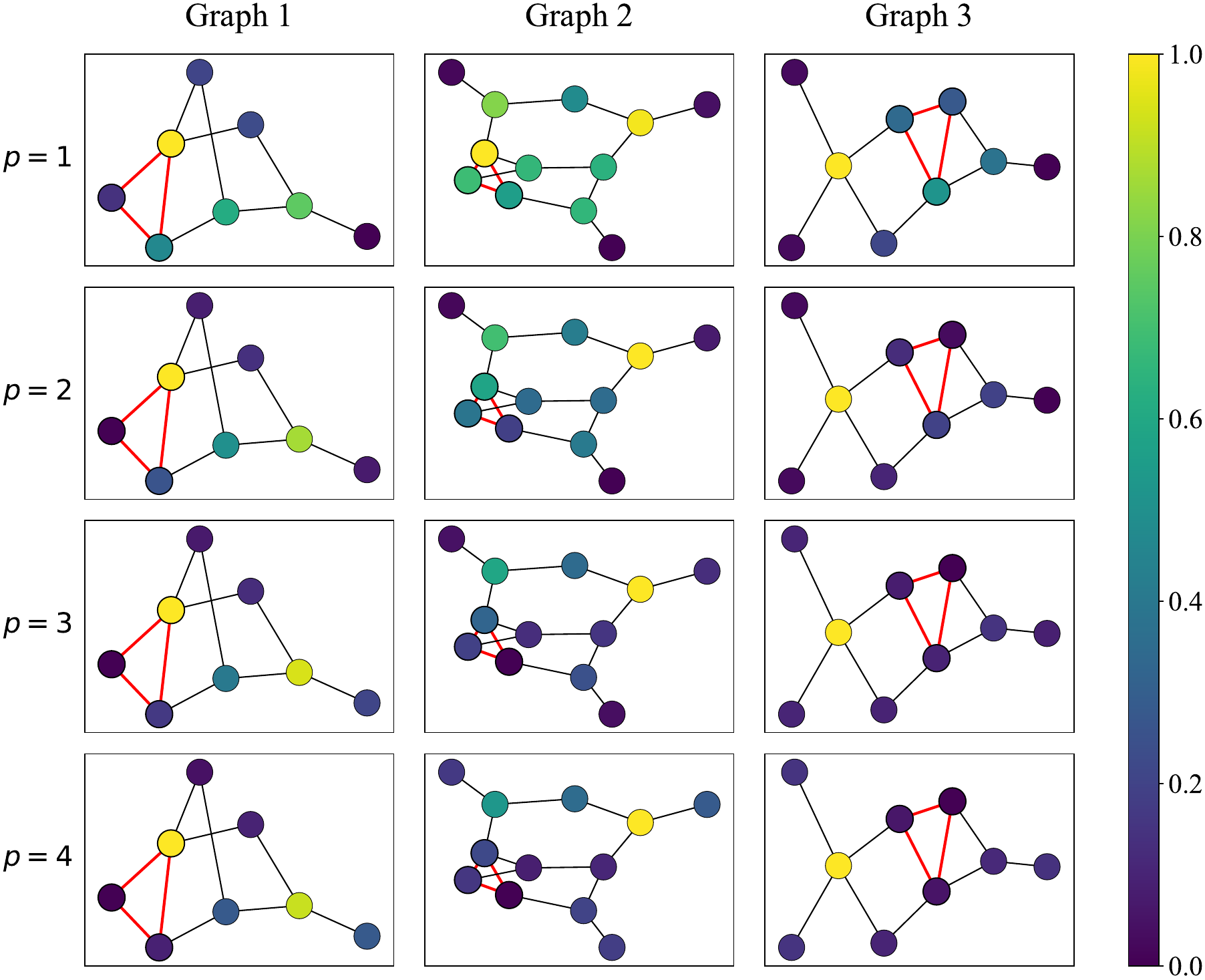}
	\end{minipage}}%
	\caption{The attention maps derived by 4 different attention functions are shown for backdoored and purified.}
	\Description{The attention maps derived by 4 different attention functions are shown for backdoored and purified.}
\end{figure*}

\section{Explanation Visualizations}
\label{A-Visualizations}
Figure \ref{F-TSNE} presents a visualization of the feature space distributions of the model, comparing its behavior with and without \sysname under three backdoor attack scenarios. The experimental results demonstrate that, after applying graph attention-based distillation, the features of trigger-embedded samples realign with their original class distributions, effectively dissolving the backdoor clusters.

Specifically, we sampled 1800 instances from the AIDS dataset, consisting of 600 samples from class 0 and 1200 samples from class 1, with half of the class 1 samples containing embedded triggers. These data points were processed by the backdoored models to obtain their feature embeddings. The embeddings were then reduced to two dimensions using the T-SNE algorithm and visualized in Figure \ref{F-TSNE}. The figure reveals that benign samples from the same category form well-defined and compact clusters in the feature space, while poisoned samples initially form separate backdoor-induced clusters (highlighted in red). However, after employing \sysname, these backdoor-induced clusters are disrupted, and the poisoned samples are re-aligned with the benign samples from their corresponding categories. This demonstrates \sysname's capability to effectively break the distinctive clustering patterns of backdoor features.

Moreover, benign samples remain tightly grouped within their respective clusters, characterized by minimal intra-cluster variance and clear inter-cluster boundaries. This indicates that the model, after purification, maintains high performance on the primary classification task. These observations underscore \sysname's dual ability to preserve the model's accuracy while successfully mitigating backdoor threats.

\section{Visualization of attention with different values of \texorpdfstring{$p$}{p}}
In this section, we visualize attention maps under different $p$-values to provide a more intuitive and in-depth understanding of our graph attention mechanism. Specifically, we analyze three sample graphs and present the attention distributions in the intermediate layers of the model for $p = 1, 2, 3, 4$. Fig. \ref{F-backdoored_vis} and \ref{F-purified_vis} illustrate the results under the Sub-BA attack without defenses and with our proposed defense mechanism deployed, respectively. 

The color scale ranges from 0 to 1, where visually ``brighter'' colors indicate higher attention scores, meaning the model assigns more importance to those nodes. Vertically, as the $p$-value increases, the model's attention becomes more concentrated on a few critical nodes. In other words, a small subset of nodes becomes brighter, while the majority of nodes dim as their scores decrease. This indicates that larger $p$-values emphasize regions with the highest neuron activations. However, this comes at a cost: the suppression of useful knowledge from benign neurons, as attention to many benign nodes diminishes. Horizontally, we observe that the backdoored model without purification tends to focus its attention predominantly on the trigger locations. In contrast, the purified model shifts its attention to benign nodes, effectively reducing its reliance on trigger-related features. Furthermore, as $p$-values increase, the brightness at trigger nodes in the purified model becomes progressively dimmer. This suggests that higher $p$-values enhance the model's ability to suppress attention on backdoor-related nodes, aiding in backdoor erasure. However, this improvement also leads to a trade-off: a higher $p$-value causes more benign nodes to dim, resulting in the loss of clean feature knowledge.

The analysis highlights the dual role of $p$-values. While larger $p$-values enhance backdoor mitigation by better redirecting attention away from trigger nodes, they also increase the risk of losing important benign features. This trade-off underscores the importance of selecting an appropriate $p$-value to strike a balance between effective backdoor removal and preserving clean feature knowledge.